\documentclass[isre,nonblindrev]{informs3}

\DoubleSpacedXI

\usepackage{natbib}
 \bibpunct[, ]{(}{)}{,}{a}{}{,}%
 \def\bibfont{\small}%

\usepackage{float}
\usepackage[para]{threeparttable}
\usepackage{makecell}
\usepackage{graphicx}
\usepackage{tikz}
\usetikzlibrary{arrows.meta,positioning,calc}
\usepackage{subfig}
\usepackage{algorithm}
\usepackage{algpseudocode}
\usepackage{longtable}
\usepackage{booktabs}
\usepackage{pdflscape}
\usepackage{algorithm}
\usepackage{algpseudocode}
\usepackage{alphalph}

\TheoremsNumberedThrough
\ECRepeatTheorems

\EquationsNumberedThrough

\MANUSCRIPTNO{ISRE-0000-0000.00}

\renewcommand{\theARTICLETOP}{}
\begin{document}

\RUNAUTHOR{Liu et al.}

\RUNTITLE{``Generate" the Future of Work}

\TITLE{``Generate" the Future of Work through AI: Empirical Evidence from Online Labor Markets}

\ARTICLEAUTHORS{%
\AUTHOR{Jin Liu\thanks{These authors contributed equally to the manuscript and are listed alphabetically.}}
\AFF{School of Management, University of Science and Technology of China, \EMAIL{liujin07@mail.ustc.edu.cn}}
\AUTHOR{Xingchen (Cedric) Xu\footnotemark[2]}
\AFF{Michael G. Foster School of Business, University of Washington, Seattle, WA 98195, \EMAIL{xcxu21@uw.edu}}
\AUTHOR{Xi Nan}
\AFF{Michael G. Foster School of Business, University of Washington, Seattle, WA 98195, \EMAIL{xinan02@uw.edu}}
\AUTHOR{Yongjun Li}
\AFF{School of Management, University of Science and Technology of China, \EMAIL{lionli@ustc.edu.cn}}
\AUTHOR{Yong Tan}
\AFF{Michael G. Foster School of Business, University of Washington, Seattle, WA 98195, \EMAIL{ytan@uw.edu}}
}

\ABSTRACT{Large Language Model (LLM)-based generative AI systems are general-purpose tools capable of augmenting or even automating a wide range of job functions, positioning them to reshape labor market dynamics. However, predicting their precise impact \textit{a priori} is challenging, given AI's simultaneous effects on both demand and supply, as well as the strategic responses of market participants. Leveraging an extensive dataset from a leading online labor platform, we document a pronounced displacement effect and an overall contraction in submarkets where required skills closely align with core LLM functionalities. Although demand and supply both decline, the reduction in supply is comparatively smaller, thereby intensifying competition among freelancers. Notably, further analysis shows that this heightened competition is especially pronounced in programming-intensive submarkets. This pattern is attributed to skill-transition effects: by lowering the human-capital barrier to programming, ChatGPT enables incumbent freelancers to enter programming tasks. Moreover, these transitions are not homogeneous, with high-skilled freelancers contributing disproportionately to the shift. Our findings illuminate the multifaceted impacts of general-purpose AI on labor markets, highlighting not only the displacement of certain occupations but also the inducement of skill transitions within the labor supply. These insights offer practical implications for policymakers, platform operators, and workers.
}%

\KEYWORDS{Generative AI, Large Language Models, General-Purpose Technology, ChatGPT, Online Labor Markets, Gig Economy, Labor Dynamics, Skill Transition}

\maketitle

\textit{``The species that survives is the one that is able best to adapt and adjust to the changing environment in which it finds itself." - \cite{megginson1963lessons}}

\section{Introduction}\label{sec:introduction}

Online labor markets, particularly gig work platforms, have emerged as a pivotal component of the global labor landscape. According to the World Bank, as of 2023, the gig economy constitutes up to 12\% of the global labor market and holds particular promise for vulnerable groups in developing nations\footnote{Please refer to \url{https://openknowledge.worldbank.org/handle/10986/40066} for additional details.}. The work-from-home and work-from-anywhere trends have continued to proliferate during and after the pandemic. As per Upwork's report, freelancers contributed \$1.35 trillion to the U.S. economy in annual earnings in 2022, representing a \$50 billion increase from the previous year, with 60 million American freelancers participating in this burgeoning sector\footnote{Please refer to \url{https://www.businesswire.com/news/home/20221213005068/en/} for additional details.}. Given their substantial impact on the economy, freelancer platforms have garnered significant attention from economic and information systems (IS) researchers \citep{stanton2025benefits, liang2025monitoring,liang2022screening}.

Nevertheless, the dynamics of the online labor market may have experienced a significant shift following the public release of ChatGPT on November 30, 2022. ChatGPT, along with other LLM-based generative AI\footnote{We hereafter collectively refer to LLM-based generative AI as ChatGPT in most instances to maintain brevity, except in formal research questions and hypotheses.}, exhibits distinct technical features that set it apart from previous technologies. In particular, these models exhibit notable zero-shot reasoning abilities across a broad range of tasks \citep{kojima2022large}, prompting some scholars to regard them as the first generation of artificial general intelligence (AGI) \citep{morris2024position}. Consequently, the advent of such tools holds the potential to directly automate certain tasks or to assist individuals in completing tasks more efficiently \citep{noy2023experimental}, thereby reducing the necessity to outsource work to others. This development is particularly consequential for online labor markets, where jobs are predominantly on-demand and short-term, rendering them especially vulnerable to fluctuations in external demand \citep{guo2025skill, sundararajan2017sharing}.

Given this potential demand shock, the resulting competitive landscape and dynamics in these markets are difficult to predict. On the one hand, prior research suggests that freelancers are highly responsive to changes in earning opportunities on online labor markets and may scale back their participation when conditions deteriorate \citep{chen2016research}, thereby stabilizing the level of competition among those who remain active. On the other hand, because ChatGPT can reduce the time and effort required to complete tasks \citep{eloundou2024gpts}, it raises the net return per contract and may encourage freelancers to continue bidding even in a contracting market, thereby intensifying competition. These divergent perspectives motivate our first research question: (1) \textit{How does LLM-based generative AI affect the competitive landscape of online labor markets?}

To address this question, we utilize a dataset comprising all job postings, bids, final transactions, and freelancer information from one of the leading freelancer platforms, covering the period from September 2021 to August 2023. We regard the launch of ChatGPT as an exogenous shock to the online labor market, given its sudden introduction and largely unexpected success. By applying natural language processing (NLP) techniques, we cluster jobs into submarkets and determine their treatment status based on the Language Modeling AI Occupational Exposure Index (LM-AIOE) \citep{eloundou2024gpts, felten2023occupational}. Using a standard difference-in-differences (DiD) framework, we identify a substantial demand displacement effect within the treated submarkets. Although labor supply in these submarkets also declines, the reduction is comparatively smaller, leading to increased competition as measured by the average number of bids per job.

Furthermore, the impact of ChatGPT may not be uniformly distributed across the treated submarkets. At the time of our study, ChatGPT primarily generated responses in natural language or programming language \citep{achiam2023gpt}. Consistent with these output formats, writing and programming continue to be the core use cases when people use generative AI for work \citep{handa2025economic}. However, the jobs requiring these two types of outputs differ substantially in their skill prerequisites, suggesting heterogeneous labor market effects. Specifically, natural language literacy can be acquired through everyday social interaction, such as conversations within families, without formal instruction. Programming, by contrast, is a skill that requires formal training and specialization \citep{robins2003learning}. This difference in skill acquisition is reflected in the population-level statistics: as of 2024, approximately 88\% of adults worldwide possess basic natural language literacy\footnote{See \url{https://data.worldbank.org/indicator/SE.ADT.LITR.ZS} for more details.}, whereas the proportion of individuals who can write in a programming language is considerably lower, even in developed countries\footnote{See \url{https://w3.unece.org/SDG/en/Indicator?id=115} for more details.}. However, as a general-purpose technology, ChatGPT enables individuals across occupations to generate code through natural language prompts \citep{eloundou2024gpts}, potentially lowering the entry barrier to programming-intensive submarkets and facilitating transition. In addition, specialization barriers can give rise to wage premiums \citep{rosen1983specialization}, which in turn create incentives for transition: On the platform we study, programming-intensive jobs have a higher average transaction value per job (\$344.75) than other jobs (\$239.05). To empirically examine this potential skill transition, we pose our second research question: (2a) \textit{Does LLM-based generative AI induce skill transition toward programming in online labor markets?}

Building on the previously constructed dataset, we conduct difference-in-differences-in-differences (DDD) analyses at the submarket level to examine the skill-transition effect. Our results show that programming-intensive submarkets experience a significantly smaller decrease in labor supply compared to other treated submarkets, a pattern that is not explained by heterogeneity in labor demand. We further investigate heterogeneity in new freelancer participation and rule out this alternative explanation. The remaining plausible explanation is therefore a skill transition toward programming among existing freelancers. To substantiate this interpretation, we perform freelancer-level analyses and document an overall tendency for freelancers to shift toward programming-intensive tasks. Moreover, freelancers with greater exposure to ChatGPT (measured by the proportion of treated jobs they apply for) exhibit a stronger transition, suggesting that the demand shock also incentivizes adaptive behavior. Finally, we find that freelancers who transition realize significant economic benefits, reinforcing the strategic value of such skill transitions.

Despite the observed skill-transition effect on average, a natural follow-up question is who participates in this transition and remains more resilient during the market downturn. The answer is difficult to predict a priori. On the one hand, ChatGPT may primarily facilitate high-skilled freelancers in making this shift, as high-skilled workers tend to be more complementary to new technologies \citep{acemoglu2011skills} and technology adoption requires absorptive capacity \citep{cohen1990absorptive}. On the other hand, ChatGPT's capabilities may lower the threshold for effective use, thereby compressing individual skill disparities and enabling less-skilled freelancers to benefit disproportionately \citep{brynjolfsson2025generative, noy2023experimental}. Given these competing perspectives, we propose to empirically examine heterogeneity in skill transition to complement the previous research question: (2b) \textit{How does the skill-transition effect vary across freelancers with different skill levels?}

To examine such heterogeneity, we construct moderators indicative of freelancers' skill levels, including pre-treatment earning efficiency and review ratings. Using DDD analyses at the freelancer level, we find that high-skilled freelancers predominantly drive the skill-transition process. Specifically, freelancers who achieve higher revenue per bid and possess superior market ratings are the primary participants in this transition.

\begin{figure}[t]
\begin{center}
\includegraphics[width=6in]{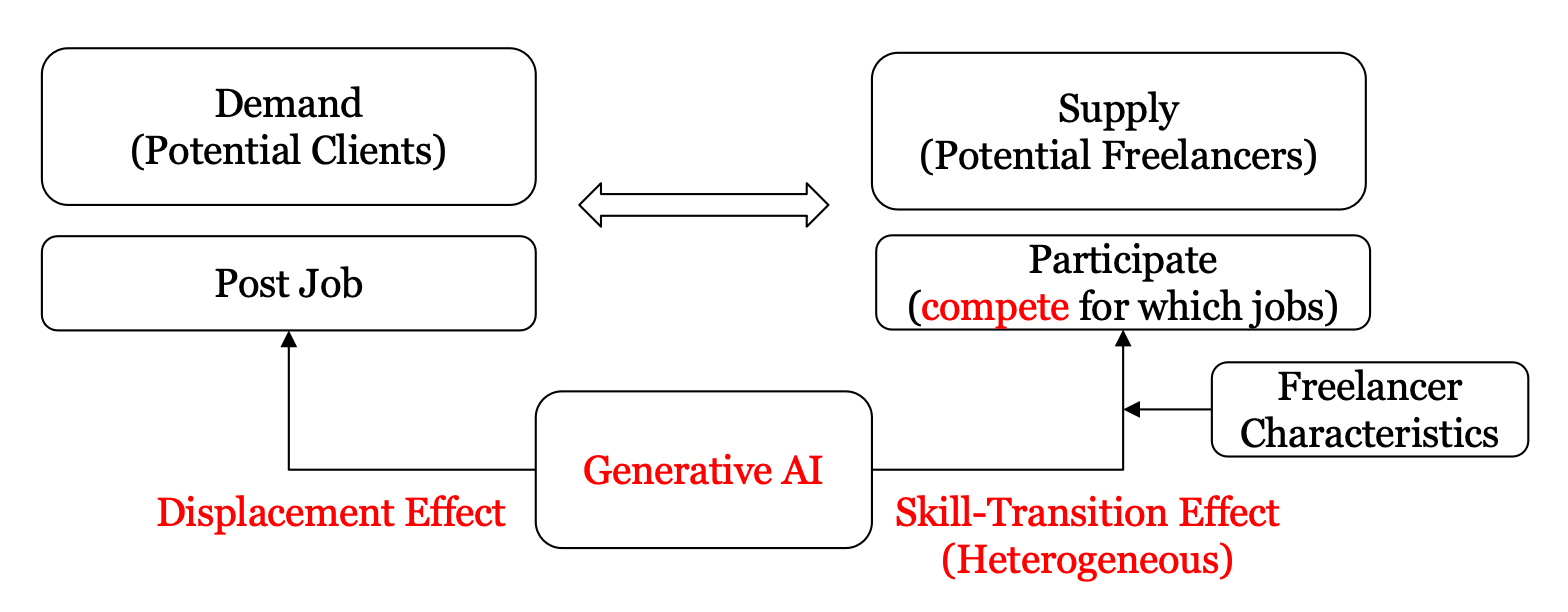}
\caption{\centering The Impact of Generative AI on Online Labor Market.}
\label{fig:framework}
\end{center}
\end{figure}

Figure \ref{fig:framework} summarizes the key empirical evidence underpinning our analysis. These results yield several contributions that distinguish our work from prior literature, the most relevant of which we summarize in Appendix \ref{appendix:recent literature}. First, we examine how a general-purpose AI affects online labor markets in which \textit{both} sides can leverage AI. We show that, beyond the familiar demand-displacement effect, AI can make the supply side more resilient on the platform, rendering the market more competitive overall. Second, when a general-purpose AI's capabilities span multiple job categories, exposed submarkets may experience \textit{symmetric} declines in demand but \textit{asymmetric} changes in supply. This asymmetry arises from a novel skill-transition effect across submarkets: general-purpose AI can simultaneously lower the barrier to acquiring occupation-specific skills for several occupations, and the strength of this enabling force varies across them. Prior literature on specialized AI has documented workers migrating to other fields because their original field is threatened by AI automation \citep{lysyakov2023threatened}. Our findings differ from these in that our skill transition is driven by AI's assistance in entering a new field (programming), even though demand in that new field is itself threatened by AI. Third, we extend skill-biased technical change (SBTC) theory by showing that skill transitions \textit{across} occupations can themselves be skill-biased, whereas prior work focuses primarily on within-occupation productivity effects that are either skill-biased or deskilling \citep{kanazawa2026ai, brynjolfsson2025generative, guo2025skill}.

The subsequent sections of this paper are organized as follows. Section 2 reviews the relevant literature and develops the theoretical hypotheses. Section 3 describes the empirical context, data collection, and variable construction methods. Section 4 outlines the identification challenges and details the selection of econometric specifications. Section 5 presents the empirical results in relation to the hypotheses. Section 6 reports robustness checks to validate our findings. Finally, Section 7 concludes by discussing detailed implications, as well as limitations and avenues for future research.

\section{Related Literature and Hypothesis Development}

\subsection{Online Labor Markets}

Contextually, our paper is situated within the realm of online labor markets, which have demonstrated a remarkable ability to effectively reconcile labor supply and demand across time and space with unparalleled flexibility \citep{stanton2025benefits, chen2019value}. A wealth of scholarly inquiries has been conducted to elucidate the complex dynamics of these digital labor marketplaces. These investigations encompass the interplay between online labor markets and external socioeconomic fluctuations, the relationship between user behavior patterns and market design, and the advancements in matchmaking algorithms.

The first line of inquiry examines how offline conditions, such as unemployment rates and stages of national development, drive individual participation in online labor markets \citep{laitenberger2023unemployment, huang2020unemployment, kanat2018surviving}. Concurrently, related research investigates the reciprocal effects of online labor markets on the external environment, such as local employment and entrepreneurship outcomes \citep{guo2025skill, burtch2018can}. The second line of inquiry focuses on modeling user behavior and developing improved market designs within online labor platforms. For example, several studies analyze how employers dynamically learn to optimize hiring strategies for suitable workers \citep{daviss2025reactions, kokkodis2023learning}, while others explore the influence of factors such as betrayal aversion and real-effort incentives on workers' labor supply decisions \citep{benndorf2024effects, hashim2024real}. Additional research examines the roles of reputation \citep{benson2020can}, monitoring systems \citep{liang2025monitoring}, capacity constraints \citep{horton2019buyer}, and wage policies \citep{chen2016research} in shaping participant dynamics, market equilibrium, and social welfare, thereby informing the design and implementation of more effective market mechanisms \citep{garg2021designing}. The third line of inquiry has refined optimization algorithms to enhance job recommendations and streamline matchmaking processes within online labor markets \citep{aouad2023online, kokkodis2023good}.

Nevertheless, the advent of LLM-based generative AI leaves open the question of how such disruptive general-purpose technology reshapes competitive dynamics in online labor markets and how freelancers adapt to it. By empirically examining ChatGPT's impact, we contribute to the first two lines of inquiry.

\subsection{Technology, Labor, and Human Capital}

This study contributes to the expanding literature on the technology-labor nexus and the role of human capital. A foundational concept is skill-biased technological change (SBTC), which posits that technological innovations disproportionately enhance the productivity and demand for skilled workers relative to their less-skilled counterparts \citep{acemoglu2011skills, autor2003skill}. Task-based frameworks further suggest that technology complements workers engaged in abstract, non-routine cognitive tasks, while substituting for routine activities \citep{autor2013growth}. Such dynamics can sometimes engender labor-market polarization: not only do high-skill roles benefit, but certain low-skill manual occupations (e.g., janitorial work) also exhibit lower susceptibility to automation, thereby expanding employment shares at both extremes of the skill distribution at the expense of middle-skill positions \citep{autor2013growth}.

Beyond technology's differential impact on the demand for jobs with varying skill requirements, the process of skill formation itself warrants attention, as indicated by human capital theory \citep{becker1962investment}. According to this theory, general human capital is transferable across firms, giving workers an incentive to finance their own training, whereas firm-specific skills are valuable only to the current employer, prompting firms to share training costs \citep{becker1962investment}. Some human capital, however, while general across firms, remains occupation-specific; this specificity can constrain workers' mobility \citep{kambourov2009occupational}. Empirical evidence confirms that workers tend to switch to jobs requiring similar task profiles in order to leverage existing skills \citep{gathmann2010general}. Although technology can enable workers to perform tasks beyond their original skill sets, successful adoption still (at both the organizational and individual levels) depends on absorptive capacity, defined as the ability to recognize the value of new knowledge, assimilate it, and apply it effectively \citep{zahra2002absorptive, cohen1990absorptive}.

Recent advances in AI have pushed the frontier in the technology-labor interplay, enabling systems to accomplish tasks once thought beyond machine capability \citep{paolillo2022compete, agrawal2019economics, brynjolfsson2014second}. Previous AI applications, however, have remained purpose-built and narrow by design \citep{morris2024position}. In contrast, LLM-based generative AI constitutes a new class of powerful general-purpose technology capable of assisting with (or automating) a broad spectrum of activities \citep{eloundou2024gpts, morris2024position}. Initial evidence documents sizable productivity gains in specific settings, such as consulting (field studies) \citep{dell2026navigating} and writing (laboratory experiments) \citep{noy2023experimental}. These studies, however, focus on workers with well-defined roles; it remains unclear how freelancers, who confront all kinds of tasks and freely choose which projects to pursue, will adapt in the post-GPT era. We address this gap by empirically examining the skill-transition effects induced by AGI. A summary of recent related literature on AI and labor markets is provided in Appendix \ref{appendix:recent literature}.

\subsection{Hypothesis Development}

\begin{figure}[t]
\begin{center}
\includegraphics[width=0.8\textwidth]{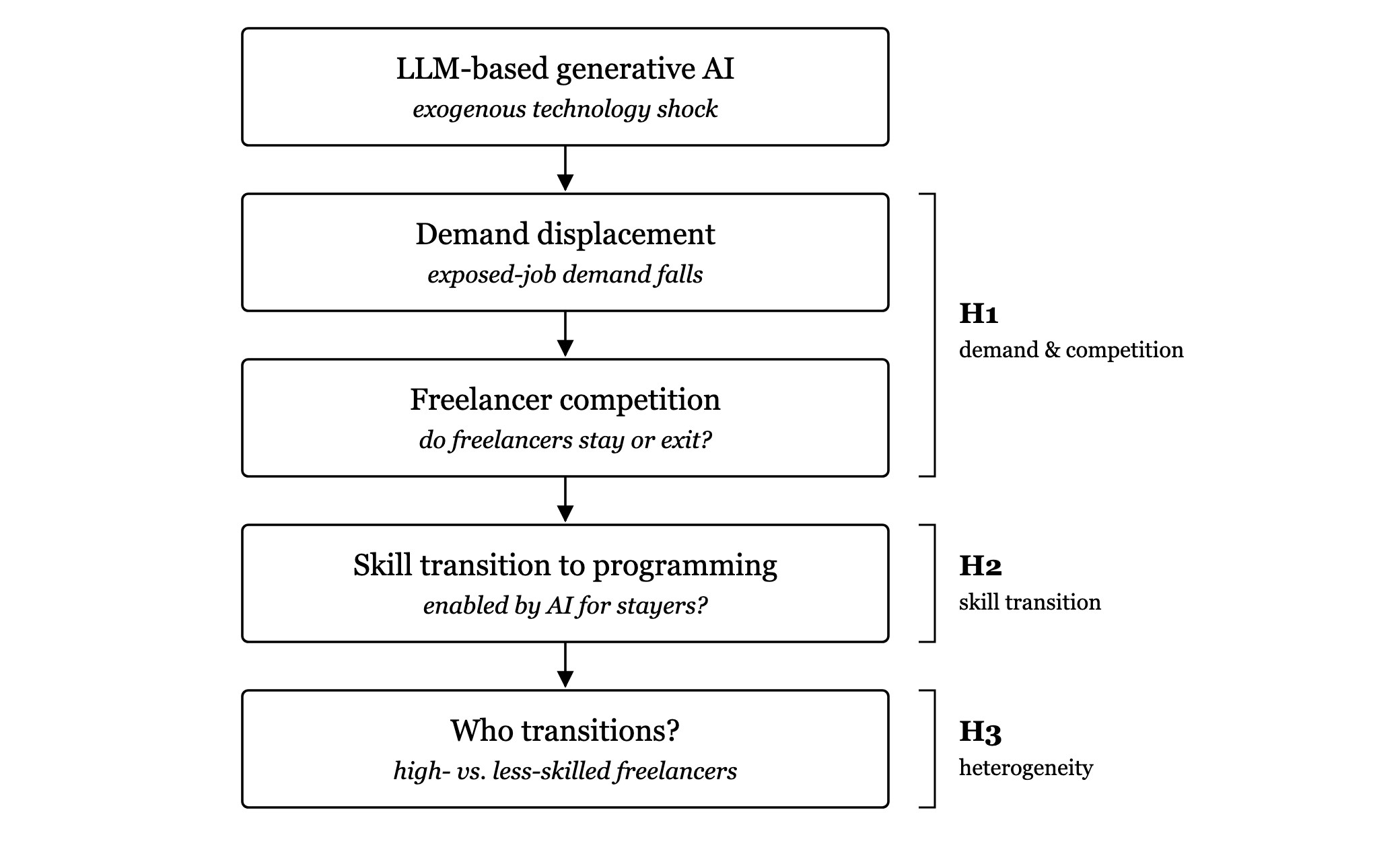}
\end{center}
\caption{\centering Conceptual Framework of the Hypotheses.}
\label{fig:hypotheses}
\end{figure}

In this section, we theorize the influence of LLM-based generative AI on market dynamics and freelancer behavior in a sequential manner that mirrors our research questions, as summarized by the conceptual framework in Figure~\ref{fig:hypotheses}. We begin by theorizing the demand displacement effect induced by generative AI and develop a pair of competing hypotheses concerning freelancers' decisions to stay in or exit the market, which determine the resulting level of competition (H1). Conditional on staying, we next theorize freelancers' potential skill-transition behavior, examining whether generative AI enables a strategic shift toward programming-intensive work (H2). Finally, we develop a pair of competing hypotheses regarding which freelancers are more likely to undertake such skill transitions (H3). The following subsections elaborate on the theoretical foundations underlying each of these hypotheses.

\subsubsection{Demand Displacement Effect and Competition Change} \

We begin our theoretical framework by formulating hypotheses regarding the overall impact of LLM-based generative AI on the competitive landscape of online labor markets. As a powerful general-purpose technology capable of fully automating or assisting with a wide range of tasks \citep{eloundou2024gpts}, generative AI has the potential to help satisfy employer demand, thereby generating a displacement effect on the online labor force \citep{acemoglu2019automation}. As a result, the advent of generative AI may lead to a reduction in demand and an overall contraction of online labor markets. Despite this potential contraction, the resulting competition dynamics remain ambiguous; accordingly, we present two opposing theoretical predictions.

On the one hand, competition among freelancers may remain stable. A fundamental advantage of online labor markets lies in their flexibility for both sides. Consequently, a substantial body of literature documents freelancers' rapid responses to changing economic conditions \citep{huang2020unemployment} and other strategic behavior in online labor markets \citep{filippas2024costly}. For instance, monitoring systems that mitigate the cold-start problem have been shown to significantly increase participation among less-experienced freelancers when expected benefits rise \citep{liang2025monitoring}. Additionally, freelancers exhibit prompt reactions to wage reductions by withdrawing their labor \citep{chen2016research}. Given freelancers' sensitivity to earning opportunities, it is plausible that they may quickly respond to decreases in demand by exiting, thereby maintaining a stable level of competition for these short-term tasks.

On the other hand, despite a potential decline in demand due to the displacement effect, freelancers may continue to participate in job bidding for several reasons, thereby intensifying competition. First, existing literature indicates that some freelancers are long-term participants in online labor markets, dedicating work hours comparable to full-time employment \citep{wood2019good}. Such freelancers tend to exhibit market stickiness \citep{gussek2024understanding}, allowing them to endure heightened competition and reduced earning opportunities. Second, generative AI can serve as a productivity-enhancing tool by assisting freelancers in task completion \citep{brynjolfsson2025generative, noy2023experimental}, thereby increasing the marginal benefit of securing jobs and making higher competition levels more sustainable. Together, these factors may result in a smaller contraction in labor supply and elevate the equilibrium competition level among remaining jobs. 

Based on the above reasoning, we propose a set of competing hypotheses for empirical testing.

\noindent \textbf{Hypothesis 1a.} LLM-based generative AI decreases demand for exposed jobs without increasing competition. 

\noindent \textbf{Hypothesis 1b.} LLM-based generative AI decreases demand for exposed jobs and heightens competition.

\subsubsection{Skill-Transition Effect} \

Next, we establish the theoretical foundations underlying the potential for freelancers to engage in skill-transition behavior within online labor markets following the introduction of LLM-based generative AI. As outlined in the development of Hypothesis 1, freelancers are strategic actors who respond adaptively to changes in the competitive landscape \citep{chen2016research}. For those who remain active in the market, it is plausible that they would change their participation across different job types as a strategic response. Specifically, human capital theory posits that jobs vary considerably in their human capital intensity, which influences wages and productivity \citep{becker1962investment}. In the digital era, programming-intensive occupations, characterized by higher cognitive complexity, typically command a wage premium \citep{brynjolfsson2014second, autor2013growth, autor2003skill}. However, because some skills tend to be occupation-specific, acquiring the competencies necessary for programming entails significant human capital investment and creates entry barriers \citep{acemoglu2011skills, kambourov2009occupational}. The introduction of ChatGPT, which demonstrates substantial capability in assisting with programming tasks \citep{eloundou2024gpts, poldrack2023ai}, may reduce these investment costs, thereby facilitating freelancers' skill transition.

While the above reasoning suggests that generative AI may promote skill transitions, additional frictions could weaken this effect. First, beyond the acquisition of technical skills, other forms of human capital, such as domain expertise and contextual knowledge, also entail substantial costs associated with learning through practice and accumulating work experience \citep{gathmann2010general, polanyi2009tacit, neal1995industry}. Consequently, ChatGPT may not sufficiently reduce the overall transition costs to induce widespread skill shifts. Second, the displacement effect (Hypothesis 1) may also impact programming jobs, thereby diminishing the relative benefits of transitioning, particularly for freelancers who are primarily engaged in tasks less susceptible to demand decreases. 

Given these competing forces, whether the reduction in programming entry barriers enabled by ChatGPT is sufficient to overcome the aforementioned frictions remains an open empirical question. To guide our investigation, we propose the following hypothesis for empirical testing.

\noindent \textbf{Hypothesis 2.} LLM-based generative AI introduces skill transitions towards programming.

\subsubsection{Heterogeneous Skill-Transition Effect} \

We further investigate the heterogeneity of skill-transition behavior, contingent upon the positive impact of ChatGPT on skill transition being sufficiently strong (Hypothesis 2). Theoretically, this heterogeneity may unfold along two competing and plausible dimensions.

On the one hand, existing literature generally suggests that high-skilled freelancers are more capable of adopting new technologies and enhancing their productivity \citep{autor2013growth, acemoglu2011skills}. Although our context involves cross-occupational skill transitions that require acquiring and leveraging new competencies, similar patterns may emerge due to the superior abilities of high-skilled freelancers in utilizing external tools \citep{hatch2004human, zahra2002absorptive, cohen1990absorptive}. Consequently, ChatGPT may provide greater incentives for skill transition among these high-skilled freelancers.

On the other hand, given the considerable capabilities of ChatGPT, the performance gap between high- and low-skilled freelancers may narrow, particularly within the same occupation \citep{brynjolfsson2025generative, noy2023experimental}. This dynamic may also apply to the skill-transition context, reducing the influence of ability differences on the transition cost \citep{hatch2004human, cohen1990absorptive}. Moreover, since less-skilled freelancers start from a comparatively disadvantaged position in their original occupations, the relative returns from making a skill transition could be greater for them \citep{becker1962investment}. Accordingly, this cost-benefit tradeoff may encourage less-skilled freelancers to engage in skill transitions more actively.

Which of these competing forces dominates is thus an empirical question, and we propose the following pair of competing hypotheses.

\noindent \textbf{Hypothesis 3a.} The proposed skill-transition effect is stronger among high-skilled freelancers.

\noindent \textbf{Hypothesis 3b.} The proposed skill-transition effect is stronger among less-skilled freelancers.

\section{Research Context, Data, and Variables}

To empirically address our research questions, we collect data from a leading freelancer platform and construct variables for both market-level and freelancer-level analyses. Section \ref{sec: research context} provides an overview of the platform, details the selection of the observation period, describes the raw dataset, outlines the job classification process based on clustering techniques, and explains the formation of submarkets. In Section \ref{sec: group}, we define the treatment and control groups at both levels of analysis. Section \ref{sec: variable} presents the development of the key variables used in our study.

\subsection{Research Context and Submarket Construction}\label{sec: research context}

The focal online labor platform enables interactions between freelancers and clients across a diverse range of job categories. Clients post job listings (i.e., projects) that outline the specific requirements and essential skills necessary for successful completion. Freelancers evaluate these postings and submit bids detailing their qualifications and proposed compensation. Clients then select from among the bidders, thereby finalizing transactions with their chosen freelancers. A detailed description of this process is provided in Appendix \ref{appendix: platform process}.

To investigate the impact of LLM-based generative AI, with a particular emphasis on ChatGPT, our observation window must encompass the launch date of ChatGPT, which occurred on November 30, 2022. To ensure sufficiently long observation periods before and after the launch time, while mitigating the confounding end-of-year effects, we establish a two-year observation window spanning from September 2021 to August 2023. Within this time period, we collect all job postings, associated job bids, and the data of freelancers who placed at least one bid on these projects. 

When a job listing is published on the platform, the client specifies relevant skill tags that denote the technical or domain-specific competencies required for the task. Across 1.6 million jobs, 2,719 unique skill tags are observed, forming 485,370 distinct combinations of skill tags (hereafter referred to as skill sets). These tags facilitate efficient search and filtering, as similar tags correspond to comparable skill requirements and attract overlapping pools of freelancers with relevant expertise. Conversely, divergent skill configurations reflect heterogeneous client demands, enabling differentiation among job types and appealing to diverse segments of the freelancer workforce. It is important to note that during this period, publicly accessible versions of text-to-image generative AI systems such as Dall-E and Midjourney were also released. We exclude the effects of these text-to-image AIs from the main analyses and address their potential influence separately in Appendix \ref{appendix: image}. After excluding skill sets containing any skill tags related to image output, 409,119 skill sets remain. For these skill sets, we define submarkets as clusters of job opportunities requiring similar skills, allowing us to assess the differential effects of ChatGPT on demand and competitive dynamics across various job categories to test our hypotheses.

Specifically, based on the skill sets available for each job, we employ a multi-step clustering procedure to uncover underlying job categories. Initially, each job's set of skill tags is transformed into a fixed-length vector representation using embedding techniques, specifically BERT embeddings, which capture the job's latent position within a multi-dimensional space \citep{wang2023scientific, devlin2019bert}. Leveraging these embeddings, we apply the Hierarchical Density-Based Spatial Clustering of Applications with Noise (HDBSCAN) algorithm to group similar skill combinations into distinct clusters \citep{mcinnes2017hdbscan}. HDBSCAN is a density-based clustering method well-suited for high-dimensional, sparse, and unevenly distributed data and has been widely adopted in management research \citep{mousavi2024resilience, nguyen2024unlimited}. Unlike centroid-based algorithms such as k-means, HDBSCAN accommodates clusters of varying densities and does not require pre-specification of the number of clusters, rendering it particularly appropriate for the data-driven identification of submarkets on the platform. Consequently, HDBSCAN facilitates the detection of cohesive job clusters characterized by similar skill requirements. Based on the clustering results, we identify 1,082 distinct submarkets. Each submarket corresponds to a cluster of similar skill sets. Since each job is associated with a specific skill set, which maps uniquely to a single cluster, we are able to assign every job to its corresponding submarket. A detailed description of this clustering procedure is provided in Appendix \ref{appendix: job classification}.

\subsection{Treatment Group, Control Group, and Treatment Timing} \label{sec: group}

In the preceding subsection, we cluster the skill sets into 1,082 distinct submarkets and subsequently assign each job to one of these submarkets based on its associated skill set. Building on this submarket classification, we then define the treatment and control groups at both the market and freelancer levels.

We operationalize treatment assignment by quantifying the degree to which job skills are related to the capabilities of language models. Specifically, we utilize LM-AIOE, a variant of the AIOE, which measures the extent to which various occupations are related to language modeling AI tools \citep{felten2023occupational}. Our approach begins by computing the semantic similarity between each skill tag and the 774 occupations for which LM-AIOE scores are available. This enables us to assign an LM-AIOE value to each skill tag based on its most semantically aligned occupation. Subsequently, we compute the weighted average LM-AIOE score for each submarket, where weights correspond to the relative prevalence of skill tags within the submarket. Submarkets exhibiting higher average LM-AIOE scores are thus considered to be more closely related to the broad functionalities of LLMs.

However, directly using the continuous LM-AIOE score as the treatment variable entails assuming a parametric linear relationship between the LM-AIOE value and outcomes in the online labor market. This assumption is strong, as the LM-AIOE score reflects the relatedness between a skill and AI capabilities \citep{felten2023occupational}, yet the precise functional form linking this relatedness to labor market outcomes is not known \textit{a priori}. Moreover, even in so-called ``sharp" designs where treatment unambiguously corresponds to a continuous measure (e.g., dosage of a medication), estimating treatment effects at varying levels remains challenging and carries a high risk of bias \citep{callaway2024difference}. Consequently, we adopt a more cautious approach, employing estimation methods with minimal parametric assumptions and conducting sensitivity analyses to evaluate robustness.

Specifically, we partition submarkets into treatment and control groups using a median split of associated occupations' LM-AIOE values: submarkets with LM-AIOE above the median are assigned to the treatment group ($Treat = 1$), while those below the median form the control group ($Treat = 0$). Consequently, our submarket-level DiD analyses compare labor market outcomes between submarkets whose required skills are more closely related to LLM core functionalities and those less related. Detailed procedures for LM-AIOE computation and treatment assignment are provided in Appendix \ref{appendix: job classification}. To verify robustness, we also experiment with alternative thresholds by varying the cutoff to the median LM-AIOE $\pm$ 0.05. The results, reported in Appendix \ref{appendix:Treatment/Control Group Definition Thresholds}, further confirm the stability of our findings.

Regarding treatment timing, it is important to note that several models and their applications were released during the observation window. We designate November 30, 2022 as the treatment date, because this date corresponds to the public release of ChatGPT, the first LLM-based text generative AI and a milestone towards artificial general intelligence \citep{morris2024position}. Because this date falls at the end of November, we define December 2022 as the first post-treatment month in our month-level analyses. Accordingly, we set $Post=1$ for all months from December 2022 onward and $Post=0$ for earlier months. For week-level analyses, we designate the week containing November 30, 2022, as the initial post-treatment period.

Finally, we construct treatment indicators at the freelancer level to examine heterogeneous responses and transitional behaviors, thereby empirically testing Hypotheses 2 and 3. We limit our sample to freelancers who placed at least one bid both before and after the treatment month, as the skill-transition process can only be measured for individuals active in both pre- and post-treatment periods \citep{wooldridge2010econometric, angrist2009mostly}. This restriction yields 132,260 freelancers for our analysis, approximately 6.7\% of the two million freelancers who bid during the observation window. Although this appears to be a small subset, it remains representative: most other freelancers bid only briefly after registration and then exit the market \citep{liang2025monitoring}. Importantly, this cohort accounts for roughly 12 million bids, 60\% of total bidding activity, making them the core contributors whose welfare and behavior underpin the platform's viability.

In the freelancer-level analysis, although freelancers may bid in multiple submarkets, most concentrate on a limited set of skills. As a result, the proportion of each freelancer's pre-treatment bids submitted in treated submarkets, which captures their initial skill relatedness to ChatGPT's capabilities, is typically close to either 0 or 1. Because this proportion resembles a binary indicator, using it as a continuous measure yields similar results to a binary treatment variable. We therefore adopt this proportion directly as the freelancer-level treatment variable ($Treat$).

\subsection{Variable Descriptions}
\label{sec: variable}

In this section, we define the primary variables used in our analysis and describe their construction. All market-level variables are aggregated at the submarket-month level. To assess labor demand and other facets of market dynamics, we compile several measures. Demand is proxied by the number of job postings in each submarket during a given month, denoted $Jobs$. Supply is captured by the total number of bids submitted to those postings, denoted $Bids$. To characterize competitive intensity and transaction value, we employ two additional variables: $Transaction$, the total monetary value of completed contracts, and $Avg\_Bids\_job$, the average number of bids per job. We also construct several alternative measures for robustness; the corresponding results are reported in Appendix \ref{appendix:Market-level Extended Analyses}.

Further, to evaluate potential transitions toward programming, we classify each submarket by its programming intensity. We first label every skill tag using the occupation mappings in \citet{felten2023occupational}; tags linked to occupations whose descriptions primarily involve programming are marked as programming-intensive. A submarket, which comprises multiple skill tags, is considered programming-intensive ($Programming = 1$) if more than 50\% of its tags are programming-intensive; otherwise, it is classified as non-programming-intensive ($Programming = 0$).

At the freelancer level, our primary interest is their transition toward programming. For each job a freelancer bids on, we compute its programming intensity, defined as the proportion of aforementioned programming-intensive tags among all tags. When constructing the panel data, we average this intensity across all jobs a freelancer bids on within a given month; this measure is denoted \(Programming\%\). To complement the panel, we calculate additional monthly variables for each freelancer that capture participation and earnings: the total number of bids submitted (\(Bids\)), the total value of completed transactions (\(Transaction\)), and the average revenue generated per bid (\(Avg\_Earn\_Bid\)). Several supplemental freelancer-level moderators are introduced in Section \ref{sec:freelancer-hte} and detailed in Appendix \ref{appendix: Variables Statistics}.

Due to the left-skewed nature of the data distribution, we apply a log transformation to all the outcome variables in the subsequent analysis, with the exception of $Programming\%$, which ranges between 0 and 1. The descriptive statistics of all these variables are presented in Appendix \ref{appendix: Variables Statistics} and Appendix \ref{appendix:Market-level Extended Analyses}.

\section{Identification Challenges and Econometric Specifications}
\label{sec: econometric}

To rigorously investigate the impact of LLM-based generative AI on online labor markets, we discuss several identification challenges prior to conducting formal analyses.

The primary challenge arises from the potential interactions among market participants. As is common in research involving markets, units can influence each other through various mechanisms, such as competition and selection \citep{jensen2018market}. In online labor markets, freelancers have the autonomy to select jobs and compete with one another \citep{liang2022screening}, rendering the comprehensive capture of market dynamics a methodological conundrum, even with randomized experiments \citep{barach2020steering}. Developing innovative statistical methods for analyzing matching markets is a promising avenue for future methodological research. However, given the currently available tools and the aspiration to expand the frontiers of our knowledge, we employ the following strategies: i) Rather than solely focusing on freelancer-level final transactional outcomes, we construct a series of submarkets to elucidate temporal dynamics in online labor markets, as detailed in Sections \ref{sec:main_results}. By utilizing market-level data, we can directly examine heterogeneous changes on the demand side, which is mainly on-demand and suffers less from cross-unit interactions \citep{hong2016comparing}. ii) However, the presence of freelancers who are active for relatively long periods and can transition across different submarkets raises concerns regarding the labor supply side analyses. Consequently, we exercise meticulous care in interpreting the effects. It is crucial to emphasize that our objective is not to capture the hypothetical labor change in the absence of freelancers' transition. Instead, at the market level, we capture the relative changes between submarkets whose required skills closely align with ChatGPT's capabilities and those less related, thereby allowing for freelancer transitions as a potential mechanism. (iii) To further elucidate the detailed transition processes on the supply side, we conduct freelancer-level analyses characterizing skill transition dynamics in Sections \ref{sec:skill_transition_effect} and \ref{sec:freelancer-hte}.

Another concern stems from the fact that all time-variant observables are potentially influenced by LLM-based generative AI, regardless of whether they exist within or outside the online labor markets. The inclusion of such post-treatment variables as controls can obstruct certain mechanisms and bias the overall effect estimation \citep{tafti2020beyond}, a phenomenon known as the bad control problem \citep{cinelli2024crash}. Therefore, we treat all observables as outcome variables and examine their changes, instead of presuming some to be exogenous control variables. In alignment with established practices \citep{braghieri2022social}, our primary specification incorporate two-way fixed effects without the introduction of control variables and presuming some to be exogenous:
\begin{align}\label{eq:main}
    &Outcome_{it} = \beta_0 + \beta_{1} * Treat_{i} * Post_t + u_i + T_t + \epsilon_{it} 
\end{align}

Here, $Outcome_{it}$ denotes the outcome of interest for unit $i$ (either a submarket or a freelancer) during month $t$. The coefficient $\beta_{1}$ captures the treatment effect, with the treatment variable at both levels defined in Section \ref{sec: group}. The term $u_i$ represents the unit-level fixed effect, while $T_t$ accounts for the year-month fixed effect. The error term $\epsilon_{it}$ encompasses unobserved factors influencing the outcome.

While the inclusion of certain post-treatment control variables can underestimate the treatment effects substantially, we perform robustness and sensitivity analyses to mitigate concerns regarding omitted variable bias and to provide more conservative interpretations of the coefficients. Further details are provided in Appendix \ref{appendix:Potential Confounders Control}.

\section{Main Analyses}
\label{sec:main_results}

In this section, we address our research questions and test the proposed hypotheses using a series of DiD analyses at both the market and freelancer levels. We first evaluate the overall impact of LLM-based generative AI on the competitive landscape of online labor markets (Hypothesis 1). We then examine freelancers' responses to this evolving landscape, assessing whether the introduction of ChatGPT has encouraged strategic skill transitions (Hypothesis 2). Finally, we analyze heterogeneity in these transitions, comparing the behavior of skilled and less-skilled freelancers (Hypothesis 3).

\subsection{Empirical Results of Demand Displacement Effect and Competition Change (Hypothesis 1)} \label{sec:demand_displacement}

To assess the displacement effect of LLM-based generative AI and the resulting shifts in the online labor market, we employ the DiD framework outlined in Section 4 and estimate ChatGPT's influence on labor demand, labor supply, transaction values, and competition levels. Column (1) of Table \ref{tab:competition(Hypothesis1)} shows a pronounced, statistically significant 22.1\% decline in labor demand (measured by job postings) in treated submarkets relative to control submarkets following ChatGPT's introduction ($\beta_{1} = -0.221$, $p < 0.01$). This contraction is consistent with a displacement mechanism: generative AI, as a powerful general-purpose technology, can automate or assist a wide array of tasks \citep{eloundou2024gpts}, thereby reducing people's reliance on online freelancers.

Column (2) of Table \ref{tab:competition(Hypothesis1)} shows that the sharp decrease in demand is accompanied by a significant reduction in labor supply, reflected in a 16.8\% decline in the total number of bids ($\beta_{1} = -0.168$, $p < 0.01$). Contraction on both sides of the market, in turn, translates into a decline in overall market value (Column (3)): total transaction volume falls by 31.5\% ($\beta_{1} = -0.315$, $p < 0.01$). Notably, the decline in total transaction volume is not only statistically significant but also economically meaningful. In the post-treatment period, the average monthly transaction value in treated submarkets is \$2,030. Since our estimates indicate a 31.5\% decrease attributable to generative AI, the counterfactual monthly transaction value in the absence of generative AI would be approximately $2,030 / (1 - 0.315) \approx \$2,964$. This implies a monthly reduction of roughly \$934 per treated submarket, or equivalently, an annual loss of approximately \$11,202. Aggregating across the 917 treated submarkets in our sample, this translates into a platform-wide reduction in transaction value of roughly \$10.27 million per year.

Beyond the aggregate displacement effect, the reductions in jobs and bids are not proportional: job postings decline by 22.1\%, whereas bids fall by 16.8\%. This asymmetry manifests in the average number of bids per job ($Avg\_Bids\_job$). Column (4) of Table \ref{tab:competition(Hypothesis1)} reports a significant increase in this metric for treated submarkets ($\beta_{1} = 0.138$, $p < 0.01$). Accordingly, competition within these submarkets intensifies after ChatGPT's introduction, lending support to Hypothesis 1b.

\begin{table}[H]
\centering
\caption{The Impact of Generative AI at the Market Level}
\label{tab:competition(Hypothesis1)}
\newcolumntype{L}[1]{>{\raggedright\arraybackslash}p{#1}}
\newcolumntype{C}[1]{>{\centering\arraybackslash}p{#1}}
\newcolumntype{R}[1]{>{\raggedleft\arraybackslash}p{#1}}
\renewcommand\arraystretch{0.55}
\begin{tabular}{L{2.7cm}C{2.5cm}C{2.5cm}C{2.5cm}C{2.5cm}}
\hline 
 \hline
 & \multicolumn{4}{c}{\(\ln(y+1)\)} \\
\cline{2-5}
 & $Jobs$ & $Bids$ & $Transaction$ & $Avg\_Bids\_job$ \\
 \cline{2-5}
 & (1) & (2) & (3) & (4) \\
 \hline
  $Treat * Post$ & -0.221*** & -0.168*** & -0.315*** & 0.138*** \\
 & (0.0318) & (0.0476) & (0.0896) & (0.0357) \\
 Constant & 3.019*** & 5.406*** & 5.625*** & 2.088*** \\
 & (0.0101) & (0.0152) & (0.0286) & (0.0114) \\
 \hline
 Submarket FE & Yes & Yes & Yes & Yes  \\
 Year-Month FE & Yes & Yes & Yes & Yes \\
 \hline
 Observations & 25,833 & 25,833 & 25,833 & 25,833 \\
 $R^2$ & 0.891 & 0.841 & 0.612 & 0.541 \\
\hline \hline
\end{tabular}
\begin{tablenotes}
\footnotesize
Note: Cluster-robust standard errors are reported at the submarket level. *** $p$ $<$ 0.01, ** $p$ $<$ 0.05, * $p$ $<$ 0.1. Unless otherwise noted, all numerical values are reported with three significant digits. When three significant digits require more than four decimal places, four decimal places are reported instead.
\end{tablenotes}
\end{table}

We complement our DiD analyses with an Interrupted Time-Series Analysis (ITSA), which offers two advantages in our setting. First, ITSA identifies the treatment effect from a shock within a single series rather than from cross-group comparisons, so time-varying differences across groups cannot confound the estimate. Second, ITSA separates the immediate effect around the shock from its effect on the post-treatment trend, a distinction that matters because the influence of a technological innovation such as LLM-based generative AI may diffuse through and accumulate in the labor market gradually. We report the model specification and full results in Appendix \ref{appendix:ITS}. As shown in Table \ref{tab: ITS}, the demand displacement effect is significant for both the immediate effect ($\beta_{2} = -0.101$, $p < 0.01$) and the trend effect ($\beta_{3} = -0.049$, $p < 0.01$), whereas the contraction in labor supply is significant only for the trend effect ($\beta_{3} = -0.0361$, $p < 0.01$; immediate effect $\beta_{2} = -0.0055$, $p > 0.1$). Moreover, the trend effect is smaller in absolute magnitude on the supply side ($-0.0361$) than on the demand side ($-0.049$). Taken together, these results suggest that some employers turn to generative AI as an alternative means of meeting their demand fairly quickly, whereas freelancers require more time to adjust their working strategies on the supply side.

These empirical results support Hypothesis 1b: LLM-based generative AI displaces demand immediately following its introduction, while labor supply declines only gradually and by a smaller magnitude, thereby intensifying competition. A freelancer's bidding decision can be modeled as a utility-maximization problem that weighs the expected probability of winning a contract against the expected benefit of completing it. Although heightened competition lowers the probability of winning, generative AI reduces the time and effort required to perform the work, which offsets part of the decline in expected utility. As a result, some freelancers remain willing to participate even as the market grows more competitive. In short, although the displacement effect reduces the number of available jobs, freelancer participation falls less sharply because AI-driven productivity gains raise expected utility and increase tolerance for stronger competition.

\subsection{Empirical Results of Skill-Transition Effect (Hypothesis 2)}\label{sec:skill_transition_effect}

As documented in Section \ref{sec:demand_displacement}, labor demand declines sharply while labor supply contracts more modestly, thereby intensifying competition among the remaining freelancers. But do these stayers continue to bid for the same type of jobs, or do they transition toward different ones? As stated in Hypothesis 2, skill transition is theoretically plausible for two reasons. First, to offset the lower probability of winning under heightened competition, remaining freelancers have an economic incentive to pursue jobs with a higher expected payoff conditional on winning. Second, LLM-based generative AI enables freelancers to upskill and take on jobs they were previously unable to complete.

Programming is a natural target for such transitions in our setting. On the platform we study, programming posts constitute a substantial share of all listings and pay, on average, roughly 44\% more than other jobs, offering the remaining freelancers both ample opportunities and a higher expected payoff conditional on winning. Historically, however, programming has posed a high entry barrier, since coding proficiency has typically demanded structured training and specialized human capital possessed by relatively few workers \citep{robins2003learning}. ChatGPT, as a general-purpose technology, is beginning to erode this barrier by translating plain natural-language instructions into working code \citep{eloundou2024gpts}, rendering the transition feasible even for freelancers without prior programming experience. We therefore empirically examine whether freelancers indeed transition toward programming.

We first analyze evidence at the market level. As described in Section \ref{sec: variable}, we construct a binary indicator, $Programming$, that classifies submarkets as programming-intensive or non-programming-intensive based on their skill requirements. We then estimate a triple-difference (DDD) model (Equation \ref{equ: ddd}), detailed in Appendix \ref{appendix: ddd}, to assess how market outcomes vary across these two submarket types. Using control submarkets as the baseline, the coefficient $\beta_{1}$ captures the post-ChatGPT change in non-programming-intensive submarkets, whereas $\beta_{2}$ measures the additional change observed in programming-intensive submarkets relative to $\beta_{1}$. The results are reported in Table \ref{tab: market-level HTE}.
\begin{equation}
    Outcome_{it} = \beta_{0} + \beta_{1} * Treat_{i} * Post_t + \beta_{2} * Treat_{i} * Post_t * Programming_{i} + u_i + T_t + \epsilon_{it}
    \label{equ: ddd}
\end{equation}

Interestingly, while we find no significant differential effect on labor demand between programming-intensive and non-programming-intensive submarkets ($\beta_{2} = 0.0313$, $p > 0.1$), programming-intensive submarkets experience a markedly smaller decline in labor supply ($\beta_{2} = 0.153$, $p < 0.01$). Consistent with this pattern, the average number of bids per job ($Avg\_Bids\_job$) increases more sharply in programming-related submarkets ($\beta_{2} = 0.116$, $p < 0.01$). Taken together, these results provide preliminary support for Hypothesis 2, suggesting that incumbent freelancers gravitate toward programming jobs and are willing to tolerate the heightened competition there.

\begin{table}[H]
\centering
\caption{The Heterogeneous Impact of Generative AI at the Market Level}
\label{tab: market-level HTE}
\newcolumntype{L}[1]{>{\raggedright\arraybackslash}p{#1}}
\newcolumntype{C}[1]{>{\centering\arraybackslash}p{#1}}
\newcolumntype{R}[1]{>{\raggedleft\arraybackslash}p{#1}}
\renewcommand\arraystretch{0.55}
\begin{tabular}{L{4.9cm}C{2.5cm}C{2.5cm}C{2.5cm}C{2.5cm}}
\hline 
 \hline
 & \multicolumn{4}{c}{\(\ln(y+1)\)} \\
 \cline{2-5}
 & $Jobs$ & $Bids$ & $Transaction$ & $Avg\_Bids\_job$ \\
 \cline{2-5}
 & (1) & (2) & (3) & (4) \\
 \hline
 $Treat * Post$ & -0.235*** & -0.239*** & -0.346*** & 0.0847**\\
 & (0.0344) & (0.0518) & (0.0964) & (0.0385) \\
 $Treat * Post * Programming$ & 0.0313 & 0.153*** & 0.0662 & 0.116*** \\
 & (0.0226) & (0.0348) & (0.0674) & (0.0270) \\
 Constant & 3.019*** & 5.406*** & 5.625*** & 2.088*** \\
 & (0.0101) & (0.0152) & (0.0286) & (0.0114) \\
 \hline
 Submarket FE & Yes & Yes & Yes & Yes  \\
 Year-Month FE & Yes & Yes & Yes & Yes \\
 \hline
 Observations & 25,833 & 25,833 & 25,833 & 25,833 \\
 $R^2$ & 0.891 & 0.841 & 0.612 & 0.541 \\
\hline \hline
\end{tabular}
\begin{tablenotes}
\footnotesize
\centerline{Note: Cluster-robust standard errors are reported at the submarket level. *** $p$ $<$ 0.01, ** $p$ $<$ 0.05, * $p$ $<$ 0.1.}
\end{tablenotes}
\end{table}

However, it is important to note that the observed heterogeneity in programming-intensive submarkets may not be entirely attributable to skill transition behavior within the platform. An alternative explanation is that generative AI reduces entry barriers and enhances efficiency for individuals outside the online labor markets to engage in programming work, potentially attracting new market entrants. To assess whether the observed supply-side heterogeneity in programming-intensive submarkets is driven by the entry of new freelancers, we introduce a new outcome variable, $New\_Freelancer$, defined as the monthly count of freelancers who newly register and submit at least one bid within a given submarket. Using the same econometric specifications as in Equations (1) and (2), we conduct both DiD and DDD analyses. The results, presented in Appendix \ref{appendix:new freelancer}, reveal no significant difference in the number of new freelancers between programming-intensive and non-programming-intensive submarkets ($\beta_2 = -0.043$, $p > 0.1$), effectively ruling out new entrants as the explanation for the observed heterogeneity. Consequently, the aggregate shift of incumbent freelancers toward programming stands out as the sole plausible explanation for the market-level heterogeneity documented in Table \ref{tab: market-level HTE}.

Next, we extend our inquiry to a more granular freelancer-level analysis of skill transition. As explained in Section \ref{sec: variable}, we construct the outcome variable \(Programming\%\), defined as the monthly average programming intensity of the jobs each freelancer bids on. Because this measure is observed only when a freelancer submits at least one bid in a given month, we restrict the sample to freelancers who are active at least once in both the pre- and post-treatment periods; otherwise their time series cannot capture ChatGPT's influence. Using this balanced panel, we estimate an Interrupted Time Series Analysis (ITSA) to trace changes in programming emphasis following ChatGPT's release. The results, shown in Appendix \ref{appendix: freelancer-level ITS}, reveal a significant post-treatment increase in \(Programming\%\), indicating that incumbent freelancers have, on average, redirected their bidding toward programming jobs. This evidence further supports Hypothesis 2.

Hypothesis 2 posits that a freelancer's decision to transition depends on the relative expected benefit of switching versus the cost incurred. Because ChatGPT induces a displacement effect, freelancers whose pre-treatment bids are less connected to ChatGPT's capabilities may face weaker incentives to transition. Consider an extreme case: if a freelancer bids exclusively on job types whose demand remains stable, when demand for programming jobs declines sharply, the relative benefit of shifting to programming diminishes, making continued bidding on original job types more attractive. By contrast, freelancers more heavily affected by displacement have stronger incentives to pivot toward programming to recoup lost earnings.

To test this mechanism, we estimate freelancer-level DiD models using Equation \ref{eq:main} with \(Programming\%\) as the dependent variable. The treatment variable, \(Treat\), equals the share of pre-treatment bids a freelancer placed in treated submarkets and thus proxies exposure to ChatGPT's displacement effect. Column (1) of Table \ref{tab:heterogeneous_transition} shows that freelancers with greater exposure are significantly more likely to shift their bidding activity toward programming work ($\beta_{1} = 0.0229$, $p < 0.01$).

Finally, we assess the economic consequences of skill transition to further corroborate Hypothesis 2. Using the same DiD specification as in Equation \ref{eq:main}, we compare ChatGPT's impact on transaction volume, bidding activity, and earning efficiency across different freelancer segments. Appendix \ref{appendix: freelancer_earnings} reports that freelancers who earned income in the pre-treatment period and relied exclusively on treated submarkets experienced an 11.6\% decline in monthly transaction value relative to those who bid only in control submarkets. More strikingly, freelancers who continued to depend on treated submarkets after ChatGPT's introduction without transitioning saw a 20.8\% reduction. Given that actively earning freelancers generate an average monthly transaction value of \$493, the average annual earnings for these freelancers would be roughly \$5,916, close to the average GDP per capita (\$6,800) in the developing economies from which 92\% of freelancers originate\footnote{Please refer to \url{https://www.imf.org/external/datamapper/NGDPDPC@WEO/OEMDC/ADVEC/WEOWORLD} for more details.}. The observed decline translates into a loss of approximately \$1,230 per year, or about 18\% of GDP per capita, posing a substantial threat to their livelihoods. By contrast, treated freelancers who undertake a skill transition realize an 11.75\% increase in earnings per bid and avoid the loss in total earnings (a modest 6.95\%, which is positive but not significant).

Taken together, these findings support Hypothesis 2 and illuminate the underlying mechanism: LLM-based generative AI lowers the human-capital barriers to programming, enabling freelancers to transition into higher-paying roles and preserve their earning efficiency despite intensified competition.

\begin{table}[h!]
\centering
\caption{Freelancer-Level Skill-Transition Effects}
\label{tab:heterogeneous_transition}
\newcolumntype{L}[1]{>{\raggedright\arraybackslash}p{#1}}
\newcolumntype{C}[1]{>{\centering\arraybackslash}p{#1}}
\renewcommand\arraystretch{0.55}
\begin{tabular}{L{3.2cm} C{1.9cm} C{1.9cm} C{1.9cm} C{1.9cm} C{1.9cm}}
\hline\hline
& \multicolumn{5}{c}{$Programming\%$} \\
\cline{2-6}
& (1) & (2) & (3) & (4) & (5) \\
\hline
$Treat \times Post$
  & 0.0229*** & 0.0697*** & 0.0138*** & 0.108*** & -0.0072*** \\
  & (0.0015) & (0.0036) & (0.0017) & (0.0025) & (0.0019) \\
Constant
  & 0.312*** & 0.360*** & 0.300*** & 0.380*** & 0.282*** \\
  & (0.0005) & (0.0012) & (0.0006) & (0.0008) & (0.0007) \\
 \hline
 $Rating$ & Both & High & Low & - & -\\
 $Earning\_Bid\_Pre$ & Both & - & - & High & Low \\
 Freelancer FE & Yes & Yes & Yes & Yes & Yes\\
 Year-Month FE & Yes & Yes & Yes & Yes & Yes\\
 \hline
Observations   & 711,738 & 140,079 & 571,659 & 218,791 & 492,947 \\
\(R^2\)        & 0.857 & 0.863 & 0.854 & 0.892 & 0.834 \\
\hline\hline
\end{tabular}
\begin{tablenotes}
\footnotesize
\centerline{Note: Cluster-robust standard errors are reported at the freelancer level. *** \(p<0.01\), ** \(p<0.05\), * \(p<0.1\).}
\end{tablenotes}
\end{table}

\subsection{Empirical Results of Heterogeneous Skill-Transition Effect (Hypothesis 3)}\label{sec:freelancer-hte}

Section \ref{sec:skill_transition_effect} demonstrates that LLM-based generative AI encourages freelancers to shift their skills toward programming. But absent any intervention, who is more likely to make this transition and seize the opportunity? Hypothesis 3 sets out a pair of competing predictions. On the one hand, high-skilled freelancers may be more inclined to switch occupations because their stronger absorptive capacity lets them adopt and exploit new technologies more effectively. On the other hand, ChatGPT's capabilities can lower the barriers to programming, narrowing the skill gap and offering larger relative gains to low-skilled freelancers, whose lower starting point leaves more room for improvement. Because these two mechanisms work in opposite directions, it is unclear \textit{ex ante} which group will show the stronger tendency to transition.

To adjudicate between these competing perspectives and test Hypothesis 3, we employ two proxies for freelancer skill level: (i) reputation, measured by the average rating received ($Rating$), and (ii) pre-treatment earning efficiency, calculated as total transaction value divided by total bids ($Avg\_Earn\_Bid\_Pre$). Both measures enter freelancer-level DDD regressions reported in Appendix \ref{appendix: freelancer-level DDD}. For interpretability, we report four subsample analyses (two proxies $\times$ two skill strata) here, which yield consistent results. 

Reputation is highly skewed. Because of the cold-start problem \citep{liang2025monitoring}, 75\% of freelancers receive no reviews, whereas those who do are almost always rated at the maximum (median = 5). This bimodal pattern (missing vs. 5) leads us to classify freelancers with $Rating = 5$ as high-skill and all others ($Rating < 5$ or unrated) as low-skill. A similar distribution characterizes earning efficiency: many freelancers earn nothing prior to treatment, leaving $Avg\_Earn\_Bid\_Pre = 0$ for a sizable share of the sample. A median split, therefore, assigns freelancers with positive pre-treatment earnings to the high-skill group and those with zero earnings to the low-skill group. For each subgroup, we replicate the DiD specification used in Section \ref{sec:demand_displacement} (Equation \ref{eq:main}) to estimate the treatment effect on freelancers' programming participation intensity, measured by \(Programming\%\).

Columns (2) and (3) of Table \ref{tab:heterogeneous_transition} report the estimates that use $Rating$ as a skill proxy. High-rated freelancers exhibit a significant post-treatment rise in the share of programming jobs ($\beta_{1}=0.0697$, $p<0.01$), an increase of roughly 19\% relative to their pre-treatment mean of 0.359. By contrast, low-rated freelancers show only a modest gain ($\beta_{1}=0.0138$, $p<0.01$), or about 4.9\% above their baseline mean of 0.284, indicating a far weaker propensity to transition. Columns (4) and (5) use pre-treatment earning efficiency ($Avg\_Earn\_Bid\_Pre$) as an alternative skill measure. Freelancers with higher earning efficiency experience a sizable increase in programming intensity ($\beta_{1}=0.108$, $p<0.01$), representing a 28\% uptick over their pre-treatment average of 0.386. In contrast, low earners even register a small decline ($\beta_{1}=-0.0072$, $p<0.01$), underscoring their difficulty in shifting toward programming work.

Taken together, these findings support Hypothesis 3a: freelancers with higher overall skill levels account for most of the observed transitions. Their superior prior skills signal greater individual absorptive capacity, which remains important for cross-occupational moves even though ChatGPT is easy to use and aids skill acquisition.

\section{Robustness Checks}

To further strengthen causal inference, we implement a series of robustness analyses addressing potential threats to identification. The key concerns and associated empirical tests are summarized below, and a summary table is provided in Appendix \ref{appendix:summary_Robust}.

First, to ensure our results are not driven by the specific classification method to define treatment and control groups, we test alternative cutoff values (Appendix \ref{appendix:Treatment/Control Group Definition Thresholds}), replace the binary indicator with the continuous LM-AIOE score (Appendix \ref{appendix:continues treatment}), adopt a tercile-based classification (Appendix \ref{appendix:multi_group}), and employ an alternative exposure measure (Appendix \ref{appendix:new_treat}), all yielding consistent results.
Second, to mitigate potential seasonal biases, we conduct a placebo test by assigning November 30, 2021, the same calendar day in the prior year, as a pseudo treatment date and reestimate the DiD specification. The resulting coefficients are statistically insignificant, thereby ruling out year end or seasonal confounders (Appendix \ref{appendix:Change Treat Time}). Third, to assess whether our results could simply reflect random fluctuations, we randomly reassign treatment status across observations and repeat the DiD estimation for each permutation. The distribution of placebo estimates is centered on zero, and the observed treatment effects lie in the tails of this distribution, indicating they are unlikely to be driven by random variation (Appendix \ref{appendix:Placebo Tests}). To address concerns about group comparability, we estimate a relative time model confirming parallel pre-treatment trends between treated and control submarkets (Appendix \ref{appendix:Pretrends and Temporal Dynamics}). We further enhance comparability using propensity score matching, which yields consistent results (Appendix \ref{appendix:PSM}). In addition, we apply ITSA to the treated submarkets, with results that align closely with our main estimates (Appendix \ref{appendix:ITS}). To mitigate omitted variable bias from unobserved external or internal factors, we add relevant controls and conduct a sensitivity analysis following \citet{altonji2005selection}. The similarity of coefficients with and without these controls indicates robustness to unobserved confounders (Appendix \ref{appendix:Potential Confounders Control}). We also test robustness through alternative modeling and data structures. Specifically, we use a Negative Binomial specification to account for count outcomes (Appendix \ref{appendix:Alternative model}), apply weekly data aggregation (Appendix \ref{appendix:Alternative Aggregation Level}), and include separate year and month fixed effects (Appendix \ref{appendix:Alternative Fixed Effect}). Moreover, we examine three additional outcome variables to assess market dynamics from different angles (Appendix \ref{appendix:Market-level Extended Analyses}). Lastly, excluding extreme submarkets at both ends of the distribution yields consistent results (Appendix \ref{appendix:Alternative Submarkets Quantity}). Taken together, all these comprehensive checks reinforce the robustness of our findings.

To further strengthen our conclusions, we conduct a series of extended analyses. First, given the rise of text-to-image generative AI tools during the observation period, we examine their impact on image-related submarkets. To capture both direct effects and potential publicity spillovers from ChatGPT's launch, we run two complementary analyses (Appendix \ref{appendix: image}). Neither yields significant results, suggesting that current text-to-image models lack the general-purpose capabilities needed to influence online labor markets. Further, while our main analysis shows a smaller decline in labor supply within programming-intensive occupations, heterogeneity may exist within programming jobs. To explore this, we categorize programming languages into scripting and compiled types. Results indicate that scripting-language-dominated submarkets do not experience significant decline relative to the control group, further illustrating the direction of skill transition (Appendix \ref{appendix:programming type}). Additionally, we examine other demand-side dynamics in treated submarkets over the observation period, focusing on budget, required skills, and the share of jobs related to programming. These results are presented in Appendices \ref{appendix: budget} and \ref{appendix: proportion and skill_num}.

\section{Implications and Conclusion}

\subsection{Summary of Key Findings}
In this study, we conduct a comprehensive analysis of the impact of LLM-based generative AI on the online labor market. Leveraging rich data from a leading freelancer platform and a DiD framework, we uncover key shifts in market dynamics. Specifically, we find a significant demand-side displacement effect in submarkets closely aligned with ChatGPT's core capabilities. While labor supply in these submarkets also declines, the reduction is smaller, intensifying competition among freelancers. More interestingly, we identify a notable skill-transition effect, with freelancers increasingly shifting their focus toward programming. Moreover, this transition exhibits substantial heterogeneity, revealing that it is primarily driven by high-skilled freelancers.

\subsection{Theoretical Contributions}

Our study contributes to the literature on online labor markets and the interplay among technology, labor markets, and human capital.

First, it expands the body of research examining how offline socioeconomic conditions influence online labor markets, such as the impact of local unemployment on freelancers' participation \citep{laitenberger2023unemployment, huang2020unemployment, kanat2018surviving}. While existing literature typically focuses on large-scale external socioeconomic shifts, our research highlights that technological innovations can also exert significant effects. Specifically, we empirically investigate how the advent of LLM-based generative AI substantially disrupts online labor demand and, notably, motivates supply-side dynamics and intensifies market competition.

Second, our research also contributes to the literature on the strategic behaviors of participants in online markets. For instance, clients may deliberately avoid high-quality yet overly popular freelancers due to capacity constraint concerns \citep{horton2019buyer}. Further, the implementation of monitoring systems influences freelancers to adjust their bidding strategies, particularly affecting inexperienced workers \citep{liang2025monitoring}. Our work extends this line of inquiry by identifying an additional type of strategic behavior, specifically freelancers' skill transition toward different job categories triggered by the introduction of ChatGPT.

Third, our study extends the SBTC theory by examining the effects of a general-purpose technology (GPT) on cross-occupation transitions. SBTC theory traditionally emphasizes how technological advances disproportionately enhance productivity and demand for high-skilled labor, potentially raising wages and attracting additional workers \citep{acemoglu2011skills, autor2003skill}. In contrast, we document a novel mechanism whereby even a uniformly negative demand impact from GPT across certain related job categories can induce skill-transition behavior among workers toward occupations requiring higher skills. This transition aligns with human capital theory, which models workers' skill acquisition decisions as a strategic trade-off between benefits and associated costs \citep{becker1962investment}. As GPT reduces the skill acquisition costs for higher-skilled and higher-paying jobs, workers become more likely to transition into these roles. Consequently, our research identifies a GPT-enabled cross-occupation transition towards more skill-intensive jobs on the supply side, a phenomenon distinct from the demand-driven effects commonly highlighted in SBTC literature \citep{autor2013growth}.

Finally, building on the observed skill-transition effects, we further demonstrate that this transition is itself skill-biased. Specifically, our findings indicate that freelancers with higher general abilities on the platform contribute disproportionately to the skill transition. This observation aligns with absorptive capacity theory, which underscores the critical role of an individual's ability to efficiently assimilate and utilize new technologies \citep{zahra2002absorptive, cohen1990absorptive}. While SBTC traditionally highlights heterogeneous productivity gains within occupations as the primary consequence of technological advancements, we extend this perspective by showing that cross-occupation transitions are also skill-biased, favoring higher-skilled workers. These theoretical insights hold significant implications for future advancements in general-purpose technologies, particularly in the trajectory towards AGI, where powerful AI could potentially both displace and augment human labor across diverse occupational contexts.

\subsection{Managerial Implications}

Our findings yield several practical implications for policymakers, platform operators, and freelancers. First, the centrality of individual absorptive capacity in enabling skill transitions implies that platforms should extend their training offerings beyond tool-specific and task-oriented competencies to include meta-learning strategies that help freelancers assimilate new knowledge amid rapid technological change. This emphasis parallels organizational absorptive capacity development: the long-term competitiveness of online labor platforms depends on the collective adaptability of their workforce to continuously attract demand. Second, in the AI era, fostering foundational AI literacy (i.e., the ability to understand, evaluate, and leverage diverse AI applications) can convert emerging technologies from potential risks into strategic opportunities. Realizing these outcomes requires a coordinated effort: platforms must design targeted programs to strengthen freelancers' absorptive capacity and AI literacy, while freelancers themselves must engage in ongoing, proactive skill development. Third, policymakers may also incorporate these insights into public upskilling and reskilling initiatives. Although the specific occupations affected may vary across broader labor markets, the underlying technology-enabled skill-transition dynamics remain relevant, making absorptive capacity and AI literacy critical considerations for effective workforce policy.

Further, AI's impact on freelancers is bidirectional: while powerful AI can enable workers to perform tasks previously beyond their capabilities, it also carries the risk of substantial demand displacement, as clients may deploy the same tools to meet their needs. Consequently, a lower strategic transition barrier does not necessarily constitute a favorable opportunity; freelancers must closely monitor shifts in demand to select appropriate skill-transition pathways, rather than blindly acquiring easily attainable skills. Furthermore, because peers are simultaneously adjusting their skill portfolios, the competitive landscape is in constant flux; freelancers should evaluate their relative positioning within the marketplace and plan transitions with strategic foresight. Platforms can also capitalize on this dynamism through targeted market and information designs. For example, by analyzing demand data, a platform might identify specific freelancer cohorts and nudge them toward alternative skills when reallocating labor would be welfare-enhancing and transitions are feasible with AI assistance.

\subsection{Limitations and Future Directions}

Several limitations of this study point to promising directions for future inquiry. First, the two-year observation window yields a relatively brief post-treatment period of only nine months, which may not fully capture longer-term market adjustments or the gradual adaptation of freelancers. Indeed, the temporal patterns surfaced in our robustness analyses (Appendices \ref{appendix:Pretrends and Temporal Dynamics} and \ref{appendix:new freelancer}) hint that the market's response to generative AI may be non-monotonic rather than a one-time, stable shift. A particularly intriguing possibility is a U-shaped adjustment, in which the immediate disruption represents only the first phase and the market subsequently recovers as the ecosystem adapts: incumbent freelancers complete their skill transitions, displaced participants exit or re-enter under revised strategies, clients recalibrate their sourcing between human and AI labor, and the two sides of the market settle into a new equilibrium. Under this view, the depth of the initial trough, the speed and extent of any rebound, and the level at which the market ultimately stabilizes each carry distinct welfare implications for platforms and freelancers, and they may also differ across submarkets depending on their exposure to generative AI. Because our short post-treatment window cannot adjudicate among these trajectories, future research should broaden the observation period and even apply network-analytic techniques alongside dynamic, game-theoretic models to trace how freelancer--client interactions unfold and adapt over the long run.

Second, this study focuses exclusively on a single online labor platform. Although our theoretical framework enhances the potential for generalization, alternative platform architectures and traditional labor markets exhibit unique institutional and operational characteristics that may produce context-specific outcomes. Accordingly, future research should conduct more analysis across a variety of settings, such as gig platforms with different governance models or offline labor exchanges, to develop a more holistic understanding of generative AI's impact on labor markets.

Third, our dataset does not include detailed measures of ChatGPT usage by individual freelancers or clients. Consequently, we analyze the overall changes following ChatGPT's introduction, rather than isolating the direct effects of adoption. Future research could obtain granular usage data to revisit these dynamics with deeper mechanism explorations. For example, it would be interesting to implement a field experiment that provisionally grants a random sample of freelancers access to specific tools and tracks subsequent shifts in their bidding strategies over time.

Furthermore, over the two-year observation window, macroeconomic fluctuations and other exogenous shocks may have influenced overall market dynamics. To mitigate these concerns, we exclude text-to-image generative AI from our primary analyses (see Appendix \ref{appendix: image}) and incorporate a set of internal and external control variables in robustness checks (see Appendix \ref{appendix:Potential Confounders Control}). We also conduct sensitivity analyses to assess the potential impact of unobserved confounders (Appendix \ref{appendix:Potential Confounders Control}). Nonetheless, fully isolating every external influence remains a limitation of this study.

In addition, client participation on the platform is highly sporadic, rendering the construction of a reliable client-side panel infeasible. Moreover, we lack data on clients' trade-offs between alternative sourcing options and platform engagement, which precludes a comprehensive understanding of demand-side dynamics. Future research could address this gap by examining demand-side behavior at a more granular level to complement our supply-focused analysis. For example, scholars might investigate how employers' hiring decisions evolve after they begin integrating generative AI tools into their working processes.

Finally, our outcome measures have room for refinement. We infer work quality solely from client ratings, which are both intermittently recorded and prone to inflation, a bias noted in prior research \citep{filippas2022reputation}. Moreover, our analysis of skill-transition effects focuses on movement into programming roles, constrained by the platform's predefined skill tags. Future studies should develop more nuanced measures of transition pathways and examine the corresponding quality outcomes to capture the full spectrum of skill-upgrading dynamics.

In summary, ongoing technological advances, especially in AI, are continually reshaping labor markets. These changes open a wealth of avenues for future research and practical innovation.

\OneAndAHalfSpacedXI
\bibliography{thebibliography}
\bibliographystyle{informs2014}

\newpage

\clearpage
\DoubleSpacedXI
\begin{APPENDICES}

\renewcommand{\thesection}{\Alph{section}}
\renewcommand{\thesection}{\AlphAlph{\value{section}}}
\counterwithin{table}{section}
\counterwithin{figure}{section}
\renewcommand{\thetable}{\thesection\arabic{table}}
\renewcommand{\thefigure}{\thesection\arabic{figure}}

\setcounter{equation}{0}
\renewcommand{\theequation}{\thesection\arabic{equation}}

\makeatletter
\@addtoreset{equation}{section}
\makeatother

\newpage

\begin{landscape}
\section{Summary of Challenges and Robustness Checks}
\label{appendix:summary_Robust}
\scriptsize
\begin{longtable}{
  >{\raggedright\arraybackslash}p{6cm}
  >{\raggedright\arraybackslash}p{8cm}
  >{\raggedright\arraybackslash}p{4cm}
  >{\raggedright\arraybackslash}p{3.5cm}
}
\caption{Summary of Challenges and Analyses} \label{table:summary_Robust} \\

\toprule
Challenges & Solutions: Reasoning and Analyses & Related Literature & Related Section  \\
\midrule
\endfirsthead

\multicolumn{4}{c}{Table \ref{table:summary_Robust} continued} \\
\toprule
Challenges & Solutions: Reasoning and Analyses & Related Literature & Related Section  \\
\midrule
\endhead

\midrule
\multicolumn{4}{r}{Continued on next page} \\
\endfoot

\bottomrule
\endlastfoot
    Treatment and Control Group Definition: Our findings may be sensitive to the treatment assignment. & 1. Alternative Thresholds: To avoid imposing a specific functional form for the relationship between LM-AIOE and labor-market outcomes, we partition submarkets at the median LM-AIOE value and then vary this cutoff in robustness checks to confirm that our results do not hinge on any particular threshold. We also report results under a continuous treatment specification (assuming a linear relationship) and a three-group partition of submarkets as robustness checks. & \citep{callaway2024difference, lu2015trade} & Section \ref{sec: group} and Appendices \ref{appendix:Treatment/Control Group Definition Thresholds}, \ref{appendix:continues treatment}, and \ref{appendix:multi_group} \\

    \multicolumn{1}{r}{} & 2. Alternative Exposure Measure: We also use an alternative LM-AIOE measure from a different paper to construct the treatment variable and rerun the estimation. & \citep{eloundou2024gpts} & Appendix \ref{appendix:new_treat} \\
    
    \midrule

    Potential Seasonal Effects: Since the launch of ChatGPT occurred at the end of the year, holidays and seasonal effects might influence the observed post-shock difference. & 3. Seasonal Placebo Test: We conduct a placebo test using November 30, 2021, the same day in the previous year, as a pseudo shock date. The resulting insignificant coefficients suggest that our findings are not driven by year-end effects or seasonal fluctuations, thereby reinforcing the robustness of the main results. & \citep{gong2023empirical} & Appendix \ref{appendix:Change Treat Time} \\

    \midrule
     
    Random Fluctuations: The observed effect could plausibly arise from random variation in the data rather than reflecting a genuine treatment impact. & 4. Placebo Test (Treatment Reassignment): We conduct a placebo test by randomly reassigning the treatment status ($Treat \times Post$) within the sample and estimate the distribution of placebo treatment effects. The resulting distribution approximates a normal distribution centered around zero, indicating that the observed treatment effects are unlikely to be driven by random chance. This further reinforces the robustness of our findings.& \citep{zhang2025omnificence,ye2025close} & Appendix \ref{appendix:Placebo Tests} \\
    
    \midrule

    Treatment and Control Group Comparability: Lack of comparability between treatment and control groups may undermine the validity of the analysis. & 5. Parallel Trends: We estimate a relative time model to verify that treated and control groups exhibit similar trends prior to treatment. The results show no significant pre-treatment differences between the groups, suggesting that the treatment and control groups are comparable. & \citep{ananthakrishnan2025political, angrist2009mostly} & Appendix \ref{appendix:Pretrends and Temporal Dynamics}\\

    \multicolumn{1}{r}{} & 6. Propensity Score Matching: We re-estimate the analysis using a propensity score-matched sample of treated and control submarkets to enhance group comparability and further validate our findings. & \citep{ran2025sometimes,ye2025close,faccio2021business} & Appendix \ref{appendix:PSM} \\
    
    \multicolumn{1}{r}{} & 7. Interrupted Time Series Analysis (ITSA): We employ an Interrupted Time Series Analysis to examine the impact exclusively within the treated submarkets, which yields consistent findings. & \citep{jiang2025reductions, al2023brand} & Appendix \ref{appendix:ITS} \\
    \midrule

    Omitted Variable Bias: Some unobservables might affect both the treatment and outcome variables. & 8. Fixed Effects: If unobservable factors are either unit-invariant or time-specific, our two-way fixed effects in the model account for these influences. & \citep{chan2019digital, angrist2009mostly} & Section \ref{sec: econometric} \\

    \multicolumn{1}{r}{} &  9. Sensitivity Analyses: We conduct sensitivity analyses by incorporating additional control variables that capture both external environmental conditions and internal market characteristics, demonstrating that our main estimates remain robust and reinforcing the validity of our findings. & \citep{sen2024does,petrova2021social,altonji2005selection} & Appendix \ref{appendix:Potential Confounders Control} \\

    \midrule

    Model Specification: All model specifications rest on particular assumptions, and relying on a single specification may inadvertently introduce bias arising from model selection. & 10. Alternative Model for Count Variables: Several of our outcome variables are count data, taking non-negative integer values. To account for their distributional characteristics, we re-estimate the models using a Negative Binomial specification. The consistent results further confirm the robustness of our findings. & \citep{gong2023empirical} & Appendix \ref{appendix:Alternative model} \\

    \multicolumn{1}{r}{} & 11. Alternative Aggregation Level: We aggregate all variables at the weekly level and re-estimate the models. The results remain qualitatively consistent with the main findings, confirming the robustness and reliability of our conclusions across different levels of data granularity.& \citep{ran2025sometimes} & Appendix \ref{appendix:Alternative Aggregation Level} \\

    \multicolumn{1}{r}{} & 12. Alternative Fixed Effect: We adopt an alternative model specification with separate fixed effects for year and month to better account for seasonal effects, enabling control over both annual fluctuations and recurring monthly patterns. & \citep{li2019product}  & Appendix \ref{appendix:Alternative Fixed Effect} \\
    
    \multicolumn{1}{r}{} & 13. Alternative Outcomes: To further validate our findings, we introduce three additional outcome variables to examine market dynamics from different perspectives, thereby demonstrating the robustness of our results.& \citep{wang2023political} & Appendix \ref{appendix:Market-level Extended Analyses} \\

    \midrule
    
    Influence of Outliers: Some submarkets with extremely large or small sizes may bias the estimates. & 14. Removing Outliers: We exclude the top 1\% of submarkets and those with an average job post count below one. The re-estimated results remain consistent with our main findings. & \citep{chen2025local} & Appendix \ref{appendix:Alternative Submarkets Quantity} \\
    
\end{longtable}

\end{landscape}

\newpage
\section{Platform Information and Matching Process}\label{appendix: platform process}

The focal platform, a publicly traded entity, stands as one of the preeminent online labor marketplaces, boasting millions of visits each month. As of the close of 2023, it had amassed over 70 million registered users and hosted upwards of 20 million job postings. The platform primarily utilizes a client-driven matching mechanism, which can be elucidated through the flowchart presented in Figure \ref{fig:Matching Process}:

\begin{enumerate}
\item \textbf{Project Posting}: A client (denoted as $u$) posts a job (project) on the platform.
\item \textbf{Bid Submission}: Freelancers can browse the available project list and potentially bid on this project. Should no bids be submitted, the process for this project terminates.
\item \textbf{Freelancer Selection}: In cases where one or more freelancers place bids, the client $u$ evaluates these bids and may select a freelancer (denoted as $v$) for the project.
\item \textbf{Project Execution}: The chosen freelancer $v$ undertakes the project. If $v$ does not complete the project, the procedure concludes prematurely.
\item \textbf{Payment and Conclusion}: Upon successful project completion, the client compensates the freelancer, effectively finalizing the transaction.
\end{enumerate}

\newpage

\begin{figure}[H]
\begin{center}
\includegraphics[width=4in]{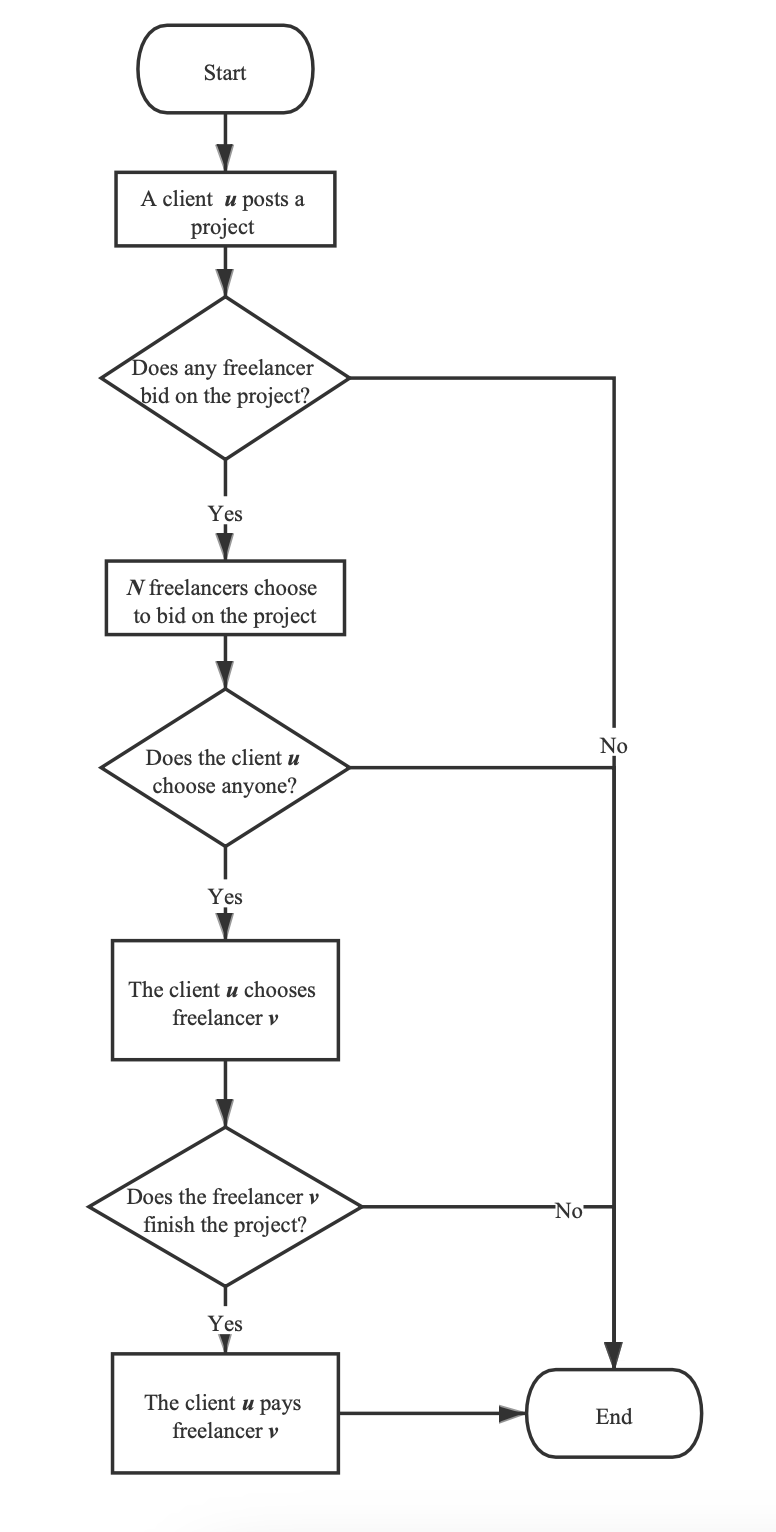}
\caption{\centering Matching Process}
\label{fig:Matching Process}
\end{center}
\end{figure}

\newpage
\section{Detailed Job Classification, Submarket Construction Process, and LM-AIOE-Based Treatment Definition}\label{appendix: job classification}

In this appendix, we provide a detailed description of the job classification procedure and the construction of submarkets. Our raw dataset includes all job postings along with their associated skill tags ($JT$), with many jobs sharing identical sets of skill tags. Jobs that share the same skill set are assumed to belong to a common demand pool and therefore compete for the same group of freelancers. Furthermore, many skill sets exhibit hierarchical or subset relationships. For example, Skill Set A may include [skill tag 1, skill tag 2, skill tag 3, \ldots, skill tag 7], while Skill Set B consists of only [skill tag 1, skill tag 2, skill tag 3]. Since Skill Set B is a subset of Skill Set A, both are considered part of the same demand pool and are presumed to compete for a similar pool of freelance labor. Building on this logic, we also treat skill tag sets that exhibit a high degree of overlap or semantic similarity as belonging to the same demand pool, assuming they compete for comparable freelancer resources. These skill sets, which share a common demand pool and freelancer pool, can be collectively regarded as a group that we define as a single submarket on the platform.

To classify all jobs based on their associated skill tag sets and subsequently construct submarkets, we first implement a multi-step clustering procedure to aggregate similar skill sets into distinct groups. Each job is then assigned to a specific group based on its associated skill set. Each resulting group represents a collection of jobs that require similar skill sets, thereby reflecting a common type of labor demand and forming the basis of a submarket. The relationships among jobs, skill sets, skill tag clusters, and submarkets are illustrated in Figure \ref{fig:process}.

After clustering, we identify 1,082 distinct submarkets from 409,119 unique skill sets, which serve as the units of our market-level analysis. Since each job is associated with a specific skill set, and each skill set is mapped to a single cluster, we are able to assign every job to a corresponding submarket. This mapping provides us with a complete classification of jobs into submarkets. Using these submarkets, we proceed to define treatment and control groups at both the market and freelancer levels. Specifically, we leverage the LM-AIOE, a variant of the AI Occupational Exposure (AIOE) Index developed by \cite{felten2023occupational, felten2021occupational}, which quantifies the degree to which various occupations are exposed to AI tools based on large language models. We begin by calculating the semantic similarity between each of the platform's 2,719 skill tags and the 774 occupations for which LM-AIOE scores are available. Each skill tag is then assigned an exposure score (LM-AIOE value) based on the most semantically aligned occupation. Next, we compute a weighted average LM-AIOE score for each submarket, where the weights reflect the distribution of skill tags within the submarket. We then adopt a conservative method by seting a threshold to define a binary treatment variable, classifying submarkets into treated and control groups. The details of this process are summarized in the pseudocode presented in Algorithm \ref{alg:Submarket_AIOE}, and the complete procedure for the multi-step clustering and the construction of treatment and control submarkets is outlined below:

\begin{enumerate}
\item \textbf{Skill Set Embedding}: We use the Sentence Transformer model (SBERT) \footnote{The Sentence Transformer model (SBERT) is a state-of-the-art neural architecture that generates semantically meaningful sentence or text embeddings. \cite{reimers2019sentence}} to generate vector representations for each skill tag set $ST$.
\item \textbf{Clustering of Skill Tag Sets Based on Embedding Representations}: We cluster the skill tag sets $ST$ based on their embedding representations using the Hierarchical Density-Based Spatial Clustering of Applications with Noise (HDBSCAN) algorithm.
\item \textbf{LM-AIOE Value Assignment for Skill Tags}: We leverage the 774 occupations $OT$ in the LM-AIOE data, each associated with a specific LM-AIOE score, to assign exposure values to skill tags. Specifically, we compute the cosine similarity between each of the platform's 2,719 skill tags $ST$ and each occupation.Then each skill tag is matched to the most semantically similar occupation, and its corresponding LM-AIOE value is assigned as the LM-AIOE value of this skill tag.
\item \textbf{Submarket LM-AIOE Values Calculation}: We use the LM-AIOE values of skill tags to compute the weighted LM-AIOE score for each submarket. Specifically, each submarket consists of multiple skill sets, and each skill set is composed of multiple skill tags. Within each submarket, skill tags occur with different frequencies. We define the frequency of a skill tag as $\frac{n_{\text{tag ST}}}{\sum n_{\text{all tags}}}$ and use this value as its weight. The submarket's LM-AIOE score is then calculated as the weighted average of the LM-AIOE values of its skill tags.
\item \textbf{Submarket Classification Based on a Defined LM-AIOE Threshold}: We define a binary treatment variable $Treat$ for each submarket based on a predefined LM-AIOE threshold $TH$. A submarket is assigned $Treat = 1$ if its LM-AIOE value exceeds the threshold $TH$, and $Treat = 0$ otherwise.
\end{enumerate}

\newpage

\begin{figure}[H]
\begin{center}
\includegraphics[width=6.5in]{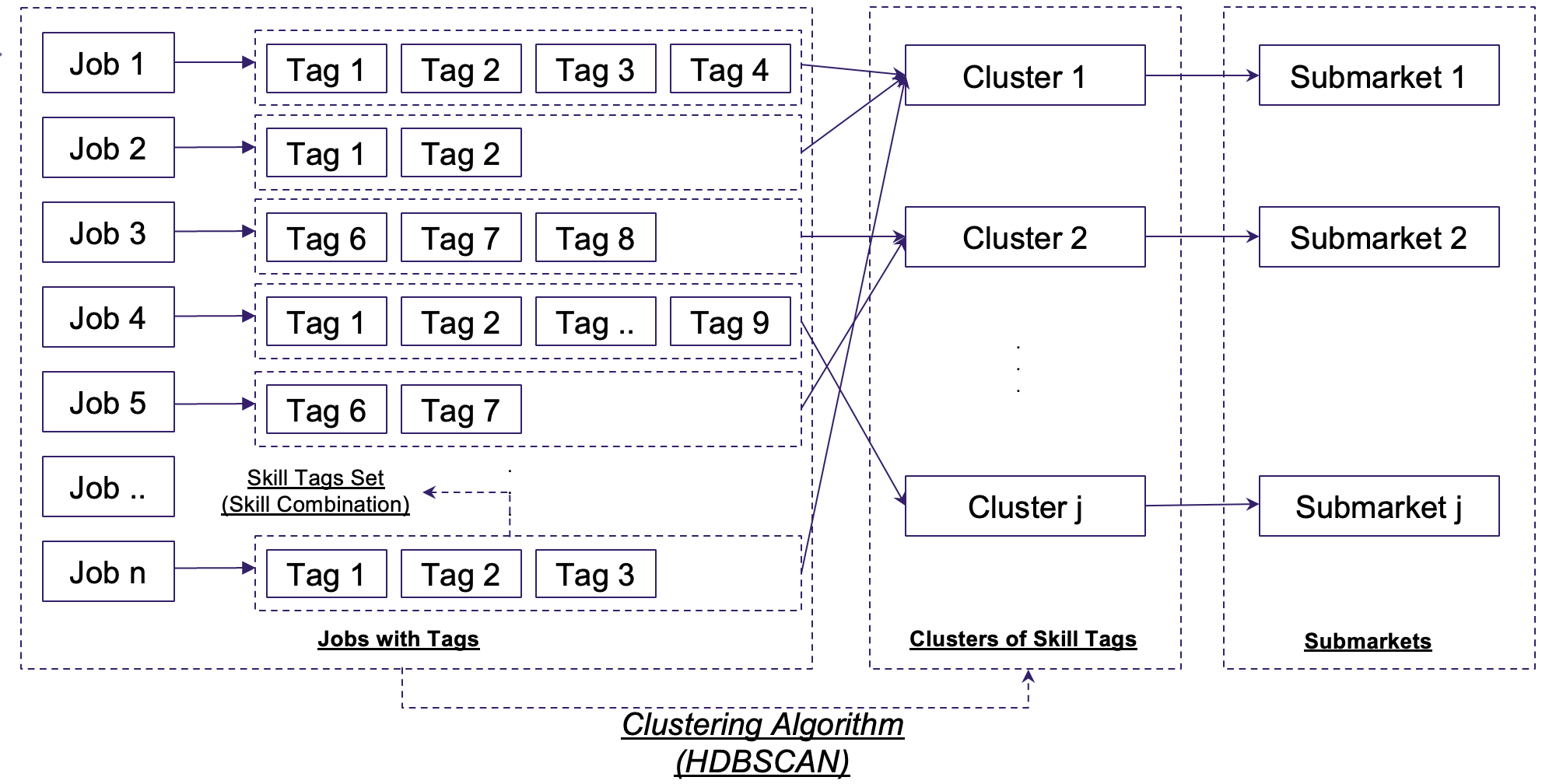}
\caption{\centering The Relationship between Jobs, Skill Sets, Skill Tags Clusters, and Submarkets.}
\label{fig:process}
\end{center}
\end{figure}

\newpage

\newpage

\newpage

\begin{algorithm}[H]
\caption{Submarket Construction and LM-AIOE-Based Treatment Definition}
\label{alg:Submarket_AIOE}
\begin{algorithmic}
\Require {Skill Tag Sets $ST$, LM-AIOE Dataset Occupations $OT$, Skill Tags List}

\State \textbf{1. Skill Set Embedding Generation}

\For{each skill tag set $st \in ST$}
    \State Compute embedding $E_{st}$ using Sentence Transformer (SBERT)\footnotemark.
\EndFor

\State \textbf{2. Skill Tag Set Clustering}

\State Cluster all embedding vectors $E_{st}$ using HDBSCAN algorithm.

\State Obtain clusters as distinct submarkets.

\State \textbf{3. Assign LM-AIOE Values to Skill Tags}

\For{each skill tag $t \in$ Skill Tags List}
    \For{each occupation $ot \in OT$}
        \State Compute cosine similarity $sim(t, ot)$
    \EndFor
    \State Assign the LM-AIOE value of the most similar occupation to tag $t$
\EndFor

\State \textbf{4. Calculate Submarket LM-AIOE Values}

\For{each submarket $s$}
    \State Initialize submarket LM-AIOE score $LM\text{-}AIOE_s$
    \State Count frequency of each skill tag $ST_s$ within submarket $s$
    \For{each skill tag $t \in ST_s$}
        \State Calculate tag frequency weight $w_t = \frac{n_{t}}{\sum_{t' \in ST_s} n_{t'}}$
        \State Update submarket LM-AIOE: $LM\text{-}AIOE_s \mathrel{+}= w_t \times LM\text{-}AIOE_t$
    \EndFor
\EndFor

\algstore{myalg}
\end{algorithmic}
\end{algorithm}

\begin{algorithm}[H]
\begin{algorithmic}
\algrestore{myalg}
\ContinuedFloat
\caption{Submarket Construction and AIOE-Based Treatment Definition (Continued)}
\State \textbf{5. Define Treatment Variable for Submarkets}

\For{each submarket $s$}
    \If{$LM\text{-}AIOE_s > TH$}
        \State Set $Treat_s = 1$
    \Else
        \State Set $Treat_s = 0$
    \EndIf
\EndFor

\end{algorithmic}
\end{algorithm}

\newpage

\section{Variable Summary Statistics}
\label{appendix: Variables Statistics}

This section reports summary statistics for the primary variables of interest in Table \ref{tab: Variables Statistics}. Because each submarket appears on the platform at different times, the number of observations varies by submarket. We define a submarket's inception as the month of its first job posting; for example, if the first posting occurs in January 2022, we set January 2022 as the submarket's start date. In each subsequent month, we record the count of job postings, assigning zero to months with no activity. Prior to January 2022, there are no observations for that submarket. As for freelancers, we treat their registration time as the moment they first enter the market.

It is notable that the variables exhibit substantial positive skewness. After applying the natural logarithm, the means and standard deviations become comparable in scale, rendering conventional count models unsuitable. The same pattern arises in the freelancer-level data: while some active freelancers submit bids to hundreds of jobs, many place only a few bids due to the cold-start problem. Accordingly, we apply log transformations to all variables except $Programming\%$, $Rating$, and $Developed$, whose limited ranges obviate the need for transformation. It is noteworthy that 301 occupations in our dataset are associated with at least one skill tag, and that the median LM-AIOE score for these occupations is 0.4. Using this value as a threshold, approximately 85\% of submarkets qualify as ``treated". This result accords with our theoretical framework and empirical observations: a substantial share of online labor-market tasks produce text or code as their primary outputs, and are therefore highly related to the capabilities of LLM.

Additionally, several data imbalances warrant attention. For the freelancer-month dataset, both $Programming\%$ and $ln(Avg\_Bids\_job+1)$ are defined as ratios with $Bids$ in the denominator, and $Bids$ may be zero. Moreover, $Rating$ values are missing for freelancers who have not yet received client reviews; in our DDD analyses, we impute these missing ratings as zero. Finally, $Tenure$ has some small number of missing data and these points are omitted from the DDD analyses. For those with a missing value in $Tenure$, we will regard their first bidding time as the beginning of the panel data.

\newpage
\begin{table}[H]
\centering
\caption{Variable Summary Statistics}
\label{tab: Variables Statistics}
\newcolumntype{L}[1]{>{\raggedright\arraybackslash}p{#1}}
\newcolumntype{C}[1]{>{\centering\arraybackslash}p{#1}}
\newcolumntype{R}[1]{>{\raggedleft\arraybackslash}p{#1}}
\renewcommand\arraystretch{0.55}
\begin{tabular}{L{5.3cm}C{1.8cm}C{1.8cm}C{1.8cm}C{1.8cm}C{1.8cm}}
\hline 
\hline
 Variable & Observation & Mean & Std & Min & Max
\\
 \hline
 \multicolumn{6}{l}{\textbf{Submarket-Month-Level}} \\
 \hline
 $ln(Jobs+1)$ & 25,833 & 2.949 & 1.244 & 0 & 6.926 \\
 $ln(Bids+1)$ & 25,833 & 5.353 & 1.858 & 0 & 10.636 \\
 $ln(Transaction+1)$ & 25,833 & 5.524 & 3.019 & 0 & 12.386 \\
 $ln(Avg\_Bids\_job+1)$ & 25,833 & 2.132 & 0.999 & 0 & 5.118 \\
 \hline
 \multicolumn{6}{l}{\textbf{Freelancer-Month-Level}} \\
 \hline
 $Programming\%$  &711,738 & 0.320 & 0.341 & 0 & 1  \\
 $ln(Bids+1)$ & 1,705,081 & 0.663 & 1.107 & 0 & 7.840\\
 $ln(Transaction+1)$ & 1,705,081 & 0.192 & 0.996 & 0 & 11.707\\
 $ln(Avg\_Earn\_Bid+1)$ & 711,738 & 0.213  & 0.828 & 0 & 11.707\\
 \hline
 \multicolumn{6}{l}{\textbf{Freelancer-Level}} \\
 \hline
$Treat$ & 132,260 & 0.879 & 0.275 & 0 &1 \\
 $Developed$  &132,260 & 0.085 & 0.279 & 0 & 1\\
 $ln(Avg\_Earn\_Bid\_Pre + 1)$ & 132,260 & 0.222 & 0.773 & 0 & 11.014\\
 $ln(Tenure + 1)$ & 130,686 & 5.954 & 1.516 & 0.003 & 8.499\\
 $Rating$ & 34,325 & 4.816 & 0.695 & 0 & 5\\
 \hline
 \multicolumn{6}{l}{\textbf{Submarket-Level}} \\
 \hline
$Treat$ & 1,082 & 0.848 & 0.360 & 0 &1 \\
 $Programming$  &1,082 & 0.390 & 0.488 & 0 & 1\\
\hline \hline
\end{tabular}
\end{table}

\newpage
\section{Notes on Difference-in-difference-in-differences Specification}
\label{appendix: ddd}

In Section \ref{sec:skill_transition_effect}, we employ a Triple Difference (DDD) specification that includes interactions among $Treat$, $Post$, and $Programming$. However, upon incorporating two-way fixed effects, we are left with only two coefficients to examine, as shown in Equation \ref{equ: ddd}. This phenomenon arises because, in our analysis, any submarket categorized under $Programming = 1$ is simultaneously considered a treated submarket ($Treat = 1$), due to its high relatedness to large language models and correspondingly elevated LM-AIOE. In other words, programming-intensive submarkets are a subset of the treated submarkets. Consequently, the interaction term $Treat * Programming$ is equivalent to $Programming$ and is also absorbed by the unit fixed effect. Similarly, $Programming * Post$ is indistinguishable from $Treat * Programming * Post$, leading to the omission of one variable. After omitting these two interaction terms, Equation \ref{equ: ddd} is formulated, wherein $Treat * Post$ captures the post-ChatGPT changes in non-programming-intensive treated submarkets relative to the control submarkets. Meanwhile, $Treat * Programming * Post$ captures additional differences in programming-intensive submarkets compared to non-programming-intensive submarkets, highlighting the heterogeneity of interest.

\newpage
\section{Interrupted Time Series Analysis of Freelancer-level Skill Transition}
\label{appendix: freelancer-level ITS}

As a supplement to Section~\ref{sec:skill_transition_effect}, we further examined skill-transition effects at the freelancer level using an Interrupted Time Series Analysis (ITSA). We estimate the ITSA model using the specification described below, with the results presented in Table~\ref{tab: ITS_freelancer}.
\begin{align}\label{eq:freelancer-its}
    Programming\%_{it} = \beta_0 + \beta_{1} * Time_t + \beta_2 * Post_t + \beta_3 * PostTime_t + u_i + \epsilon_{it} 
\end{align}
Here, $Programming\%_{it}$ denotes the average programming intensity of bids submitted by freelancer $i$ in month $t$. $Time_t$ represents the number of months elapsed since the start of the observation period. $Post_t$ is a binary indicator equal to 0 during the pre-treatment period and 1 during the post-treatment period. $PostTime_t$ equals 0 before the treatment and increases by one in each subsequent post-treatment month. We also include $u_i$ to control for freelancer-level fixed effects, while the error term $\epsilon_{it}$ captures unobserved factors that may influence the outcome. To ensure that the skill-transition process is observable, we restrict the sample to freelancers who submitted at least one bid in both the pre- and post-treatment periods. We find a statistically significant post-treatment increase in $Programming\%$ ($\beta_2 = 0.0500$, $p < 0.01$), indicating that freelancers indeed transitioned toward programming jobs following the introduction of ChatGPT.

\begin{table}[H]
\centering
\caption{Results for Interrupted Time Series Analysis}
\label{tab: ITS_freelancer}
\newcolumntype{L}[1]{>{\raggedright\arraybackslash}p{#1}}
\newcolumntype{C}[1]{>{\centering\arraybackslash}p{#1}}
\newcolumntype{R}[1]{>{\raggedleft\arraybackslash}p{#1}}
\renewcommand\arraystretch{0.55}
\begin{tabular}{L{2.5cm}C{3cm}}
\hline 
 \hline
 & $Programming\%$ \\
 \hline
 $Time$ & 0.0010***  \\
 & (0.0001)  \\
 $Post$ & 0.0500*** \\
 & (0.0009) \\
 $Post\_Time$  & -0.0001 \\
 & (0.0003)  \\
 \hline
 Freelancer FE & Yes \\
 \hline
 Observations & 432,558 \\
\hline \hline
\end{tabular}
\begin{tablenotes}
\footnotesize
\centerline{Note: Robust standard errors are reported at the freelancer level. *** $p$ $<$ 0.01, ** $p$ $<$ 0.05, * $p$ $<$ 0.1.}
\end{tablenotes}
\end{table}

\newpage
\section{Freelancer-level Heterogeneity in Skill Transition: DDD and Event-study Evidence}
\label{appendix: freelancer-level DDD}

To complement the analysis in Section~\ref{sec:freelancer-hte}, we investigate heterogeneity in freelancers' skill-transition behavior by estimating a difference-in-differences-in-differences (DDD) model. The specification is as follows, with results reported in Table~\ref{tab:F2}:
\begin{align}
Programming\%_{it} = \, & \beta_0 + \beta_1 * Treat_i * Post_t * Moderator_i + \beta_2 * Treat_i * Post_t \notag \\
& + \beta_3 * Post_t * Moderator_i + u_i + T_t + \epsilon_{it}
\label{eq:ddd}
\end{align}
In this specification, \(Treat_i\) is a continuous variable ranging from 0 to 1, representing the proportion of each freelancer's pre-treatment bids submitted in treated submarkets. This variable captures the degree of initial skill relatedness to ChatGPT's capabilities. \(Moderator_i\) reflects freelancer characteristics, operationalized in four ways across the columns: freelancer rating (Column~(1)), the logarithm of pre-treatment earning efficiency (Column~(2)), a binary indicator for whether the freelancer is located in a developed country (Column~(3)), and the logarithm of platform tenure (Column~(4)).

The estimates in Columns~(1) and~(2) are consistent with those from the main analysis: higher-skilled freelancers (proxied by higher ratings or by greater earning efficiency) are significantly more likely to shift into programming work (\(\beta_{1}=0.0256\) and \(0.0328\), respectively; \(p<0.01\)). Column~(3) reveals no statistically significant difference in transition behaviour between freelancers in developed and developing countries (\(\beta_{1}=0.0040\); \(p>0.10\)). Finally, Column~(4) indicates that freelancers with longer platform tenure are more inclined to move into programming-intensive jobs (\(\beta_{1}=0.027\); \(p<0.01\)).

Furthermore, we estimate an event-study (relative-time) specification to test the parallel-trends assumption. As shown in Figure \ref{fig:programming parallel trend}, the pre-treatment coefficients cluster around zero and are statistically insignificant, indicating no violation of the parallel-trends assumption.
\begin{align}
Outcome_{it} 
&= \sum_{k \neq -1} \beta_k \, D_{i,t=k} + u_i + T_t + \epsilon_{it} \nonumber \\
&= \sum_{k \neq -1} \beta_k \left( D_i \cdot \mathbf{1}\{t - \tau_i = k\} \right) + u_i + T_t + \epsilon_{it},
\label{equ:relative_time_freelancer}
\end{align}

\newpage
\begin{table}[H]
\centering
\caption{DDD Analyses about Skill-Transition Effects}
\label{tab:F2}
\newcolumntype{L}[1]{>{\raggedright\arraybackslash}p{#1}}
\newcolumntype{C}[1]{>{\centering\arraybackslash}p{#1}}
\newcolumntype{R}[1]{>{\raggedleft\arraybackslash}p{#1}}
\renewcommand\arraystretch{0.55}
\begin{tabular}{L{4.9cm} c c c c}
\hline\hline
& \multicolumn{4}{c}{Programming\%} \\
\cline{2-5}
 & (1) & (2) & (3) & (4) \\
\hline
\(Treat \times Post \times Moderator\)
  & 0.0256*** & 0.0328*** & 0.0040 & 0.0271***\\
& (0.0006) & (0.0013) & (0.0045)& (0.0010) \\
\(Treat \times Post\) 
  & -0.0366*** & 0.0084*** & 0.0226*** & -0.153***\\
& (0.0022) & (0.0017) & (0.0016) & (0.0068) \\
\(Post \times Moderator\) 
  & -0.0165*** & -0.0212*** & -0.0193*** & -0.0155***\\
& (0.0005) & (0.0012) & (0.0043)& (0.0009) \\
Constant 
  & 0.328*** & 0.316*** & 0.313*** & 0.348***\\
& (0.0008) & (0.0006) & (0.0006) & (0.0024) \\
\hline
Moderator    & $Rating$ & $ln(Earning\_Bid\_Pre + 1)$ & $Developed$ & $ln(Tenure + 1)$\\
Freelancer FE    & Yes & Yes & Yes & Yes\\
Year-Month FE    & Yes & Yes & Yes & Yes\\
\hline
Observations     &  711,738 &  711,738 & 711,738 & 697,908 \\
R-squared        & 0.8584   & 0.8578   & 0.8576 & 0.8572   \\
\hline\hline
\end{tabular}

\vspace{1mm}
\footnotesize
\begin{center}
Note: Cluster-robust standard errors are reported at the freelancer level. *** $p$ $<$ 0.01, ** $p$ $<$ 0.05, * $p$ $<$ 0.1.
\end{center}
\end{table}

\begin{figure}[H]
  \centering
  \includegraphics[width=5.5in,height=3in]{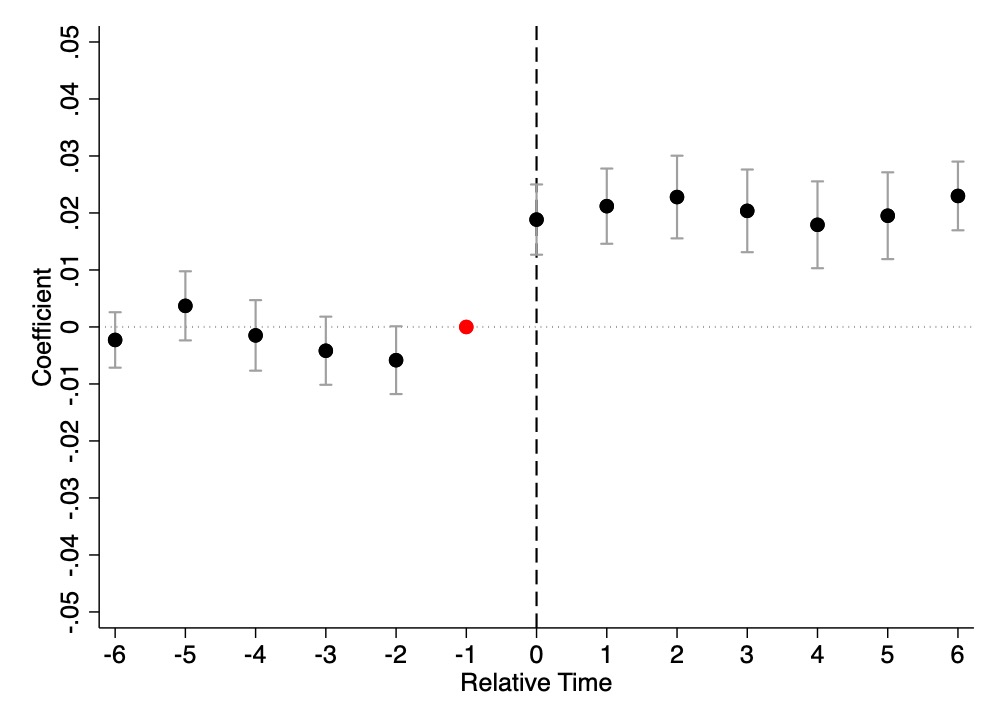}
  \captionsetup{singlelinecheck=false}
  \caption{\centering Pre-Trends and Temporal Dynamics of Freelancer's Programming Proportion}
  \label{fig:programming parallel trend}
\end{figure}

\newpage
\section{Additional Details on Freelancer-Level Earnings}
\label{appendix: freelancer_earnings}

Similar to our market-level analysis, we examine the effect of ChatGPT on freelancer-level earnings.  Using the specification in Equation \ref{eq:main}, we estimate the impact on freelancer's total transaction value and the mechanisms driving the changes, including changes in bidding behavior and earning efficiency.

Since freelancers who did not receive any payment during their pre-treatment period had no potential for income loss, we first focus on freelancers who were awarded at least one job and had prior earning activity, as shown in Table~\ref{tab:G21}. We examine how the magnitude of change in transactions is affected by ChatGPT, which varies across different freelancers. Here, $Treat$ is a continuous variable ranging from 0 to 1, representing the proportion of bids submitted to treated submarkets. A higher value of $Treat$ indicates that the freelancer is more likely to be affected by ChatGPT's demand displacement effect. We find a significant decline in transaction value (\( \beta_{1} = -0.116,\, p < 0.05 \)), which is primarily driven by a substantial reduction in the number of bids submitted (\( \beta_{1} = -0.0870,\, p < 0.01 \)) and a small, statistically insignificant decrease in earning efficiency (\( \beta_{1} = -0.0041,\, p > 0.1 \)).

In Tables \ref{tab:G22} and \ref{tab:G23}, we separately analyze freelancers who made a skill transition and those who did not. We define skill-transitioning freelancers as those whose $Programming\%$ increases after the introduction of ChatGPT. Freelancers who transitioned exhibit no significant decline in transaction value; their bidding activity remains stable, and their earning efficiency significantly improves. In contrast, those who did not transition experience substantial reductions in both transaction value and bid volume, along with a significant decline in earning efficiency.

For completeness, Tables \ref{tab:G11}, \ref{tab:G12} and \ref{tab:G13} replicate these analyses including all freelancers (even those with zero pre-treatment earnings). Unsurprisingly, the estimated effects are attenuated because most freelancers report zero earnings before treatment, which precludes substantial further reduction, but the qualitative patterns are consistent.

\newpage

\begin{table}[H]
\centering
\caption{Impact on Freelancers' Earnings}
\label{tab:G21}
\newcolumntype{L}[1]{>{\raggedright\arraybackslash}p{#1}}
\newcolumntype{C}[1]{>{\centering\arraybackslash}p{#1}}
\newcolumntype{R}[1]{>{\raggedleft\arraybackslash}p{#1}}
\renewcommand\arraystretch{0.55}
\begin{tabular}{L{3cm} c c c}
\hline\hline
& \multicolumn{3}{c}{\(\ln(y+1)\)} \\
\cline{2-4}
& \(Transaction\) 
& \(Bids\) 
& \(Avg\_Earn\_Bid\) \\ 
\cline{2-4}
 & (1) & (2) & (3) \\
\hline
\(Treat \times Post\) 
  & -0.116** & -0.0870*** & -0.0041 \\
& (0.0467) & (0.0265) & (0.0351) \\
\(Constant\) 
  & 1.005*** & 1.727*** & 0.629*** \\
& (0.0152) & (0.0086) & (0.011) \\
\hline
Freelancer FE    & Yes & Yes & Yes \\
Year-Month FE    & Yes & Yes & Yes \\
\hline
Observations     & 310,771 & 310,771 & 218,716 \\
\(R^2\)          & 0.279   & 0.670   & 0.244  \\
\hline\hline
\end{tabular}
\begin{tablenotes}
\footnotesize
\centerline{Note: Cluster-robust standard errors are reported at the freelancer level. *** $p$ $<$ 0.01, ** $p$ $<$ 0.05, * $p$ $<$ 0.1.}
\end{tablenotes}
\end{table}

\begin{table}[H]
\centering
\caption{Impact on Freelancers' Earnings (Transition)}
\label{tab:G22}
\newcolumntype{L}[1]{>{\raggedright\arraybackslash}p{#1}}
\newcolumntype{C}[1]{>{\centering\arraybackslash}p{#1}}
\newcolumntype{R}[1]{>{\raggedleft\arraybackslash}p{#1}}
\renewcommand\arraystretch{0.55}
\begin{tabular}{L{3cm} c c c}
\hline\hline
& \multicolumn{3}{c}{\(\ln(y+1)\)} \\
\cline{2-4}
& \(Transaction\) 
& \(Bids\) 
& \(Avg\_Earn\_Bid\) \\ 
\cline{2-4}
 & (1) & (2) & (3) \\
\hline
\(Treat \times Post\) 
  & 0.0695 & 0.0483 & 0.118*** \\
& (0.0523) & (0.0322) & (0.0452) \\
\(Constant\) 
  & 1.055*** & 1.888*** & 0.593*** \\
& (0.0175) & (0.0108) & (0.0149) \\
\hline
Freelancer FE    & Yes & Yes & Yes \\
Year-Month FE    & Yes & Yes & Yes \\
\hline
Observations     & 217,637 & 217,637 & 160,571 \\
\(R^2\)          & 0.290   & 0.670   & 0.246   \\
\hline\hline
\end{tabular}
\begin{tablenotes}
\footnotesize
\centerline{Note: Cluster-robust standard errors are reported at the freelancer level. *** $p$ $<$ 0.01, ** $p$ $<$ 0.05, * $p$ $<$ 0.1.}
\end{tablenotes}
\end{table}

\begin{table}[H]
\centering
\caption{Impact on Freelancers' Earnings (Non-Transition)}
\label{tab:G23}
\newcolumntype{L}[1]{>{\raggedright\arraybackslash}p{#1}}
\newcolumntype{C}[1]{>{\centering\arraybackslash}p{#1}}
\newcolumntype{R}[1]{>{\raggedleft\arraybackslash}p{#1}}
\renewcommand\arraystretch{0.55}
\begin{tabular}{L{3cm} c c c}
\hline\hline
& \multicolumn{3}{c}{\(\ln(y+1)\)} \\
\cline{2-4}
& \(Transaction\) 
& \(Bids\) 
& \(Avg\_Earn\_Bid\) \\ 
\cline{2-4}
 & (1) & (2) & (3) \\
\hline
\(Treat \times Post\) 
  & -0.208*** & -0.146*** & -0.156*** \\
& (0.0540) & (0.0340) & (0.0496) \\
\(Constant\) 
  & 0.762*** & 1.212*** & 0.683*** \\
& (0.0157) & (0.0099) & (0.0131) \\
\hline
Freelancer FE    & Yes & Yes & Yes \\
Year-Month FE    & Yes & Yes & Yes \\
\hline
Observations     & 98,504 & 98,504 & 60,573 \\
\(R^2\)          & 0.230 & 0.609 & 0.265 \\
\hline\hline
\end{tabular}
\begin{tablenotes}
\footnotesize
\centerline{Note: Cluster-robust standard errors are reported at the freelancer level. *** $p$ $<$ 0.01, ** $p$ $<$ 0.05, * $p$ $<$ 0.1.}
\end{tablenotes}
\end{table}

\begin{table}[H]
\centering
\caption{Impact on Freelancers' Earnings, Including Those with Zero Pre-Treatment Earnings}
\label{tab:G11}
\newcolumntype{L}[1]{>{\raggedright\arraybackslash}p{#1}}
\newcolumntype{C}[1]{>{\centering\arraybackslash}p{#1}}
\newcolumntype{R}[1]{>{\raggedleft\arraybackslash}p{#1}}
\renewcommand\arraystretch{0.55}
\begin{tabular}{L{3cm} c c c}
\hline\hline
& \multicolumn{3}{c}{\(\ln(y+1)\)} \\
\cline{2-4}
& \(Transaction\) 
& \(Bids\) 
& \(Avg\_Earn\_Bid\) \\ 
\cline{2-4}
 & (1) & (2) & (3) \\
\hline
\(Treat \times Post\)
  & -0.0228*** & -0.0425*** & -0.0140 \\
& (0.0072) & (0.0057) & (0.0094) \\
Constant 
  & 0.200*** & 0.678*** & 0.218*** \\
& (0.0025) & (0.0020) & (0.0033) \\

\hline
Freelancer FE    & Yes & Yes & Yes \\
Year--Month FE    & Yes & Yes & Yes \\
\hline
Observations     & 1,705,081 & 1,705,081 & 711,738 \\
\(R^2\)          & 0.361   & 0.671   & 0.322   \\
\hline\hline
\end{tabular}
\begin{tablenotes}
\footnotesize
\centerline{Note: Cluster-robust standard errors are reported at the freelancer level. *** $p$ $<$ 0.01, ** $p$ $<$ 0.05, * $p$ $<$ 0.1.}
\end{tablenotes}
\end{table}

\begin{table}[H]
\centering
\caption{Impact on Freelancers' Earnings, Including Those with Zero Pre-Treatment Earnings (Transition)}
\label{tab:G12}
\newcolumntype{L}[1]{>{\raggedright\arraybackslash}p{#1}}
\newcolumntype{C}[1]{>{\centering\arraybackslash}p{#1}}
\newcolumntype{R}[1]{>{\raggedleft\arraybackslash}p{#1}}
\renewcommand\arraystretch{0.55}
\begin{tabular}{L{3cm} c c c}
\hline\hline
& \multicolumn{3}{c}{\(\ln(y+1)\)} \\
\cline{2-4}
& \(Transaction\) 
& \(Bids\) 
& \(Avg\_Earn\_Bid\) \\ 
\cline{2-4}
 & (1) & (2) & (3) \\
\hline
\(Treat \times Post\) 
  & 0.0109 & 0.0229*** & 0.0084 \\
& (0.0087) & (0.0073) & (0.0121) \\
\(Constant\) 
  & 0.268*** & 0.823*** & 0.254*** \\
& (0.0030) & (0.0025) & (0.0043) \\

\hline
Freelancer FE    & Yes & Yes & Yes \\
Year-Month FE    & Yes & Yes & Yes \\
\hline
Observations     & 932,942 & 932,942 & 430,712 \\
\(R^2\)          & 0.376   & 0.699   & 0.318   \\
\hline\hline
\end{tabular}
\begin{tablenotes}
\footnotesize
\centerline{Note: Cluster-robust standard errors are reported at the freelancer level. *** $p$ $<$ 0.01, ** $p$ $<$ 0.05, * $p$ $<$ 0.1.}
\end{tablenotes}
\end{table}

\begin{table}[H]
\centering
\caption{Impact on Freelancers' Earnings, Including Those with Zero Pre-Treatment Earnings (Non-Transition)}
\label{tab:G13}
\newcolumntype{L}[1]{>{\raggedright\arraybackslash}p{#1}}
\newcolumntype{C}[1]{>{\centering\arraybackslash}p{#1}}
\newcolumntype{R}[1]{>{\raggedleft\arraybackslash}p{#1}}
\renewcommand\arraystretch{0.55}
\begin{tabular}{L{3cm} c c c}
\hline\hline
& \multicolumn{3}{c}{\(\ln(y+1)\)} \\
\cline{2-4}
& \(Transaction\) 
& \(Bids\) 
& \(Avg\_Earn\_Bid\) \\ 
\cline{2-4}
 & (1) & (2) & (3) \\
\hline
\(Treat \times Post\) 
  & -0.0117 & 0.0033 & -0.0179 \\
& (0.0077) & (0.0067) & (0.0118) \\
\(Constant\) 
  & 0.098*** & 0.449*** & 0.154*** \\
& (0.0026) & (0.0023) & (0.0039) \\

\hline
Freelancer FE    & Yes & Yes & Yes \\
Year--Month FE    & Yes & Yes & Yes \\
\hline
Observations     & 814,668 & 814,668 & 294,952 \\
\(R^2\)          & 0.291   & 0.539   & 0.341   \\
\hline\hline
\end{tabular}
\begin{tablenotes}
\footnotesize
\centerline{Note: Cluster-robust standard errors are reported at the freelancer level. *** $p$ $<$ 0.01, ** $p$ $<$ 0.05, * $p$ $<$ 0.1.}
\end{tablenotes}
\end{table}

\newpage
\section{The Heterogeneous Effects of Generative AI Across Programming Domains}\label{appendix:programming type}

In our market-level analysis, we find that although programming-intensive and non-programming-intensive submarkets exhibit no consistent differences in labor demand, the contraction in labor supply is significantly smaller in programming-intensive submarkets. We aim to take one step further in exploring whether any heterogeneity exists among programming jobs. Large language models support numerous programming languages since they have been trained on multi-language code corpora \citep{nguyen-etal-2023-vault, kocetkov2023the} and evaluated on benchmarks spanning a wide range of programming languages \citep{10.1109/TSE.2023.3267446}. Broadly, programming languages can be categorized into two distinct types: scripting languages and compiled languages \citep{prechelt2000empirical, ousterhout1998scripting}. These categories differ in syntactic complexity and learning cost: scripting languages typically feature simpler syntax and lower learning barriers, while compiled languages are generally more complex and demanding \citep{lokkila2023data}, potentially influencing how different languages mediate the impact of generative AI.

In light of this, we further divide the programming-intensive submarkets into two segments based on the predominant type of programming language, in order to examine the heterogeneous impact on scripting-language-dominated versus compiled-language-dominated submarkets. Specifically, we classify each submarket according to the relative proportions of scripting and compiled languages used. If the proportion of scripting languages exceeds that of compiled languages within a submarket, it is categorized as scripting-language-dominated; otherwise, it is considered compiled-language-dominated. A small number of submarkets, however, do not predominantly feature either language type and instead rely on alternative paradigms, such as the declarative language SQL. As there are only four such submarkets, we exclude them from the analysis.

Subsequently, following the model specifications in Section \ref{sec: econometric}, we separately compare the demand, supply, transaction, and competition in scripting-language-dominated and compiled-language-dominated submarkets with those in the control submarkets. The estimation results are reported in Table \ref{tab: market-level-HTE programming type}. Panel A compares scripting-language-dominated submarkets with the control group, while Panel B compares compiled-language-dominated submarkets with the control group. Consistent with the main results, both types of submarkets show a notable decline in labor demand ($\beta_1 = -0.188$ or $-0.234,$ $p < 0.01$) and total transaction volume ($\beta_1 = -0.283$, $p < 0.01$ or $\beta_1 = -0.268,$ $p < 0.05$) relative to the control group. However, it is worth noting that on the supply side, measured by the total number of bids, scripting-language-dominated submarkets do not exhibit a significant decline relative to the control group ($\beta_1 = -0.064$, $p > 0.1$). In contrast, compiled-language-dominated submarkets still experience a significant reduction in labor supply ($\beta_1 = -0.131$, $p < 0.05$). This result may reflect the lower barrier to entry associated with scripting languages compared to compiled languages, which generally require more formal training. Such suggestive evidence indicates that the skill-transition effect is inherently constrained: freelancers cannot move freely into every role without encountering substantive limitations.

\begin{table}[H]
\centering
\caption{The Heterogeneous Effects of Generative AI Across Different Types of Programming Submarkets}
\label{tab: market-level-HTE programming type}
\newcolumntype{L}[1]{>{\raggedright\arraybackslash}p{#1}}
\newcolumntype{C}[1]{>{\centering\arraybackslash}p{#1}}
\newcolumntype{R}[1]{>{\raggedleft\arraybackslash}p{#1}}
\renewcommand\arraystretch{0.55}
\begin{tabular}{L{2.7cm}C{2.5cm}C{2.5cm}C{2.5cm}C{2.5cm}}
\hline 
 \hline
 \multicolumn{5}{c}{Panel A} \\
 & \multicolumn{4}{c}{Programming Type: Scripting Language VS Control} \\
 \cline{2-5}
 & \multicolumn{4}{c}{\(\ln(y+1)\)} \\
 \cline{2-5}
  & $Jobs$ & $Bids$ & $Transaction$ & $Avg\_Bids\_job$ \\
  \cline{2-5}
 & (1) & (2) & (3) & (4) \\
 \hline
 $Treat * Post$ & -0.188*** & -0.0640 & -0.283*** & 0.213*** \\
 & (0.0348) & (0.0524) & (0.105) & (0.0408) \\
 Constant & 2.863*** & 5.199*** & 5.467*** & 2.0950***\\
 & (0.0082) & (0.0123) & (0.0246) & (0.0096)\\
 \hline
 Submarket FE & Yes & Yes & Yes & Yes  \\
 Year-Month FE & Yes & Yes & Yes & Yes \\
 \hline
 Observations & 10,268 & 10,268 & 10,268 & 10,268 \\
 $R^2$ & 0.895 & 0.848 & 0.636 & 0.544 \\
 \hline
 \multicolumn{5}{c}{Panel B} \\
 & \multicolumn{4}{c}{Programming Type: Compiled Language VS Control} \\
 \cline{2-5}
 & \multicolumn{4}{c}{\(\ln(y+1)\)} \\
 \cline{2-5}
  & $Jobs$ & $Bids$ & $Transaction$ & $Avg\_Bids\_job$ \\
  \cline{2-5}
 & (1) & (2) & (3) & (4) \\
 \hline
 $Treat * Post$ & -0.234*** & -0.131** & -0.268** & 0.181*** \\
 & (0.0371) & (0.0560) & (0.103) & (0.0421) \\
 Constant & 2.759*** & 4.887*** & 5.151*** & 1.933*** \\
 & (0.0067) & (0.0102) & (0.0187) & (0.0076) \\
 \hline
 Submarket FE & Yes & Yes & Yes & Yes  \\
 Year-Month FE & Yes & Yes & Yes & Yes \\
 \hline
 Observations & 7,496 & 7,496 & 7,496 & 7,496 \\
 $R^2$ & 0.898 & 0.847 & 0.645 & 0.524 \\
\hline \hline
\end{tabular}
\begin{tablenotes}
\footnotesize
\centerline{Note: Cluster-robust standard errors are reported at the submarket level. *** $p$ $<$ 0.01, ** $p$ $<$ 0.05, * $p$ $<$ 0.1.}
\end{tablenotes}
\end{table}

\newpage
\section{The Impact of Generative AI on Market Entry Decisions of New Freelancers}\label{appendix:new freelancer}

Given that some freelancers can be active in online labor markets for an extended period and work for diverse jobs, several potential mechanisms may explain the average and heterogeneous impacts on the labor supply side, as documented in Sections \ref{sec:demand_displacement} and \ref{sec:skill_transition_effect}. Utilizing the market-level data, we explore the influx of new freelancers as one such mechanism. To this end, we introduce a new outcome variable, $New\_Freelancer$, defined as the monthly count of freelancers who newly register and submit at least one bid within specific submarkets. We employ the same econometric specifications to perform DiD and DDD analyses, with the results presented in Table \ref{tab: Market Entry Decisions of New Freelancers}. The estimates suggest that the decline in new participants is one contributing mechanism to the observed decrease in average labor supply across treatment submarkets ($\beta_1 = -0.145$, $p < 0.01$). However, there is no significant heterogeneity in the number of new freelancers between programming-intensive and non-programming-intensive submarkets ($\beta_2 = -0.0430$, $p > 0.1$), effectively ruling out differences in new participation as the driving mechanism for heterogeneous effects on labor supply. We also test the parallel trend assumption and examine whether any differential trends existed prior to the introduction of ChatGPT. As shown in Figure \ref{fig:new freelancer parallel trend}, the confidence intervals for all pre-treatment periods encompass zero, suggesting an absence of systematic differences before the intervention.

\newpage

\begin{table}[H]
\centering
\caption{The Impact of Generative AI on Market Entry Decisions of New Freelancers}
\label{tab: Market Entry Decisions of New Freelancers}
\newcolumntype{L}[1]{>{\raggedright\arraybackslash}p{#1}}
\newcolumntype{C}[1]{>{\centering\arraybackslash}p{#1}}
\newcolumntype{R}[1]{>{\raggedleft\arraybackslash}p{#1}}
\renewcommand\arraystretch{0.55}
\begin{tabular}{L{4.9cm}C{3cm}C{3cm}}
\hline 
\hline
 & \multicolumn{2}{c}{$ln(New\_Freelancer+1)$}\\
 \cline{2-3} 
 & (1) & (2) \\
 \hline
 $Treat * Post$ & -0.145*** & -0.125*** \\
 & (0.0371) & (0.0411) \\
 $Treat * Post * Programming$ & & -0.0430 \\
 & & (0.0283) \\
 Constant & 3.591*** & 3.591*** \\
 & (0.0119) & (0.0119) \\
 \hline
 Submarket FE & Yes & Yes \\
 Year-Month FE & Yes & Yes \\
 \hline
 Observations & 25,833 & 25,833 \\
 $R^2$ & 0.900 & 0.901 \\
\hline \hline
\end{tabular}
\begin{tablenotes}
\footnotesize
\centerline{Note: Cluster-robust standard errors are reported at the submarket level. *** $p$ $<$ 0.01, ** $p$ $<$ 0.05, * $p$ $<$ 0.1.}
\end{tablenotes}
\end{table}

\begin{figure}[H]
  \centering
  \includegraphics[width=5.5in,height=3in]{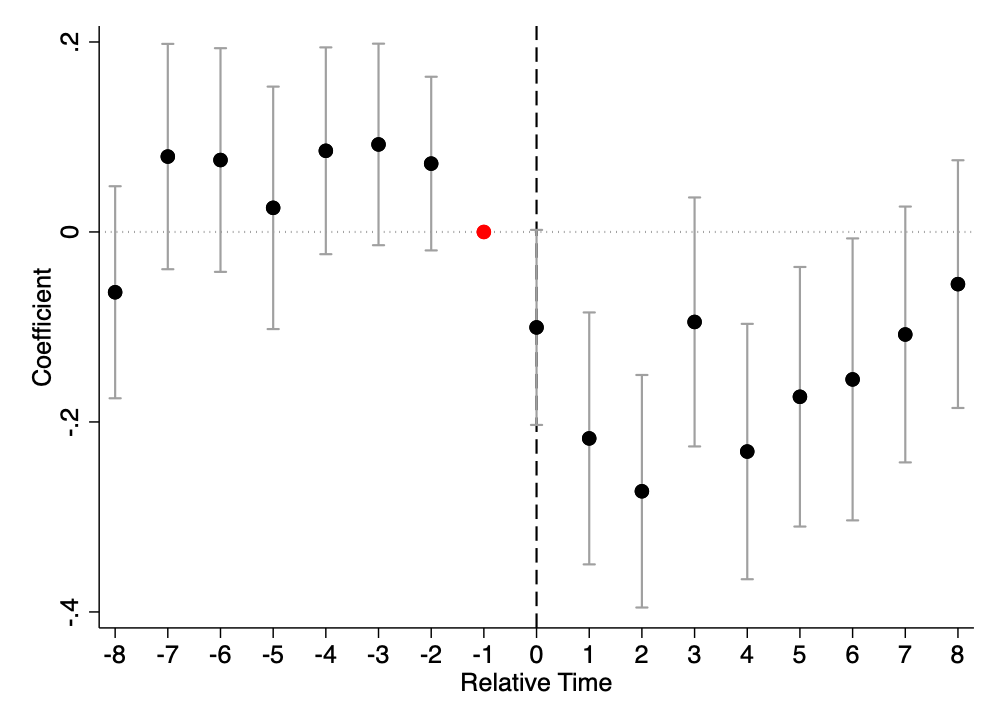}
  \captionsetup{singlelinecheck=false}
  \caption{\centering Pre-Trends and Temporal Dynamics of New Freelancers}
  \label{fig:new freelancer parallel trend}
\end{figure}

\newpage
\section{The Impacts on Project Budgets}
\label{appendix: budget}

To better understand the underlying motivations behind freelancers' transitions, we further examine how the introduction of ChatGPT affects job budgets. As demonstrated in Sections \ref{sec:skill_transition_effect}, freelancers shift toward bidding on programming-intensive jobs. While these jobs have traditionally commanded higher wages than text-related ones, ChatGPT may have further widened this wage gap. This widening gap likely strengthens freelancers' motivations to transit to programming-intensive submarkets as a way to mitigate Chatgpt's negative impact. Therefore, we examine the wages clients are willing to offer for completing a given task, as reflected in job budgets, and analyze how these budgets have changed differentially between programming-intensive and text-based submarkets following the introduction of ChatGPT.

Specifically, we define a new dependent variable, $Budget$, as the average project budget of all jobs posted within a submarket during a given month. We then follow the DDD model specification outlined in Section \ref{sec:skill_transition_effect} to investigate the heterogeneous treatment effects (HTE) on job budgets. The corresponding estimation results are presented in Table \ref{tab: budget}. The negative and statistically insignificant coefficients of the interaction term $Treat \times Post \times Programming$ suggest that the wage gap was not widened by the introduction of ChatGPT. Therefore, the absence of a significant widening in the wage gap allows us to rule out this explanation.

\begin{table}[H]
\centering
\caption{The Impact of Generative AI on Project Budgets}
\label{tab: budget}
\newcolumntype{L}[1]{>{\raggedright\arraybackslash}p{#1}}
\newcolumntype{C}[1]{>{\centering\arraybackslash}p{#1}}
\newcolumntype{R}[1]{>{\raggedleft\arraybackslash}p{#1}}
\renewcommand\arraystretch{0.55}
\begin{tabular}{L{4.9cm}C{3cm}C{3cm}}
\hline 
\hline
 & \multicolumn{2}{c}{$ln(Budget+1)$}\\
 \cline{2-3} 
 & (1) & (2) \\
 \hline
 $Treat * Post$ & 0.0575* & 0.0727** \\
 & (0.0327) & (0.0349) \\
 $Treat * Post * Programming$ & & -0.0327 \\
 & & (0.0223) \\
 Constant & 5.669*** & 5.669*** \\
 & (0.0105) & (0.0105) \\
 \hline
 Submarket FE & Yes & Yes \\
 Year-Month FE & Yes & Yes \\
 \hline
 Observations & 25,086 & 25,086 \\
 $R^2$ & 0.381 & 0.381 \\
\hline \hline
\end{tabular}
\begin{tablenotes}
\footnotesize
\centerline{Note: Cluster-robust standard errors are reported at the submarket level. *** $p$ $<$ 0.01, ** $p$ $<$ 0.05, * $p$ $<$ 0.1.}
\end{tablenotes}
\end{table}

\newpage
\section{Additional Evidence on Demand-Side Market Dynamics}
\label{appendix: proportion and skill_num}

To provide additional insight into market demand dynamics, we visualize the temporal evolution of three key indicators within the treated submarkets: the proportion of programming-intensive jobs, the average number of required skills per job, and the average job budget. Specifically, we calculate the monthly proportion of programming-intensive jobs in the overall market and plot its trend over time. The results are presented in Figure \ref{fig:programming proportion}, where the horizontal axis represents the number of months relative to the treatment point, and the vertical axis indicates the proportion of programming-intensive jobs. The blue dots show the observed monthly values, while the red line illustrates the fitted trend estimated through linear regression. Similarly, we calculate the monthly average number of required skills per job and average job budget across the entire market and plot its changes over time in Figure \ref{fig:skill num} and Figure \ref{fig:budget}. 

Based on the results presented in Figure \ref{fig:programming proportion}, the proportion of programming-intensive jobs remains largely stable over the observation period, exhibiting only a slight decline. The overall variation is minimal, and the fitted trend line is nearly flat. This suggests that, on the demand side, clients have not demonstrated any notable shift in preference or a growing tendency toward programming-intensive tasks. Figure \ref{fig:skill num} presents the temporal trend in the number of required skills per job. The fitted trend line appears relatively flat with a slight upward slope, indicating a gradual increase in the number of skills requested by clients. Figure \ref{fig:budget} illustrates the monthly evolution of the average job budget, revealing a modest upward trend over time, particularly following the introduction of ChatGPT. This pattern suggests a gradual rise in job budgets within the treated submarkets and aligns with the observed increase in skill requirements. We re-estimate the DiD model separately to compare high- and low-budget submarkets with the control group. As reported in Table \ref{tab:K1_new}, the results show a higher demand displacement effect in low-budget submarkets.

\newpage

\begin{figure}[H]
\begin{center}
\includegraphics[width=5.5in, height=3in]{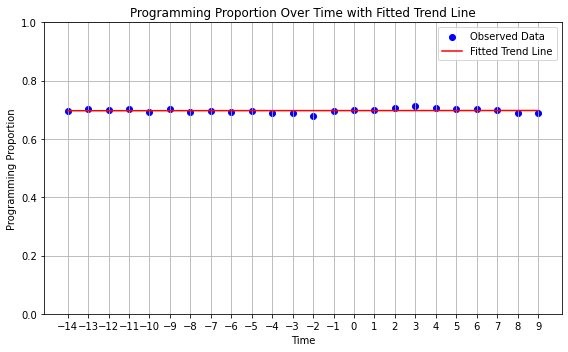}
\caption{\centering Temporal Trends in Proportion of Programming Jobs}
\label{fig:programming proportion}
\end{center}
\end{figure}

\begin{figure}[H]
\begin{center}
\includegraphics[width=5.5in, height=3in]{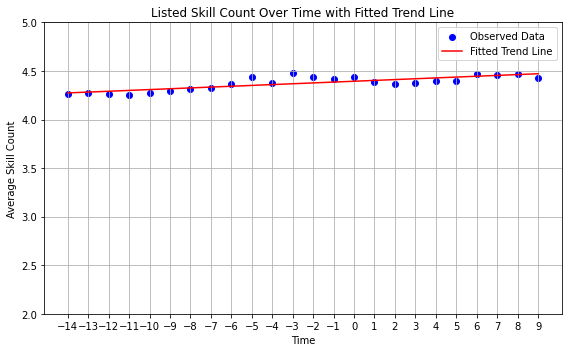}
\caption{\centering Temporal Trends in Average Skill Count}
\label{fig:skill num}
\end{center}
\end{figure}

\newpage

\begin{figure}[H]
\begin{center}
\includegraphics[width=5.5in, height=3in]{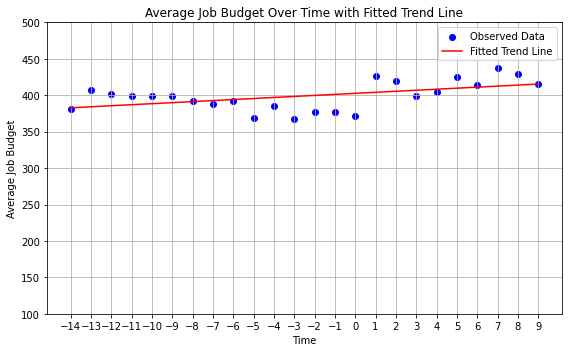}
\caption{\centering Temporal Trends in Average Budget}
\label{fig:budget}
\end{center}
\end{figure}

\begin{table}[H]
\centering
\caption{The Heterogeneous Effects of Generative AI Across Sub-markets with Different Budget Levels}
\label{tab:K1_new}
\newcolumntype{L}[1]{>{\raggedright\arraybackslash}p{#1}}
\newcolumntype{C}[1]{>{\centering\arraybackslash}p{#1}}
\renewcommand\arraystretch{0.55}
\begin{tabular}{L{2.7cm}C{2.5cm}C{2.5cm}C{2.5cm}C{2.5cm}}
\hline\hline
\multicolumn{5}{c}{\textbf{Panel A}}\\
&\multicolumn{4}{c}{High-Budget Sub-market}\\
\cline{2-5}
&\multicolumn{4}{c}{$ln(y+1)$}\\
\cline{2-5}
& $Jobs$ & $Bids$ & $Transaction$ & $Avg\_Bids\_job$\\
\cline{2-5}
& (1) & (2) & (3) & (4)\\
\hline
Treat $\times$ Post  & -0.185*** & -0.156*** & -0.233** & 0.0886**\\
                     & (0.0381)   & (0.0566)   & (0.110)  & (0.0400)\\
Constant             & 3.0780***  & 5.615***  & 5.713*** & 2.272***\\
                     & (0.0121)   & (0.0181)   & (0.0351)  & (0.0128)\\
\hline
Submarket FE & Yes & Yes & Yes & Yes\\
Year-Month FE & Yes & Yes & Yes & Yes\\
\hline
Observations  & 17{,}421 & 17{,}421 & 17{,}421 & 17{,}421\\
R$^{2}$       & 0.889 & 0.836 & 0.597 & 0.501\\
\hline
\multicolumn{5}{c}{\textbf{Panel B}}\\
&\multicolumn{4}{c}{Low-Budget Sub-market}\\
\cline{2-5}
&\multicolumn{4}{c}{$ln(y+1)$}\\
\cline{2-5}
& $Jobs$ & $Bids$ & $Transaction$ & $Avg\_Bids\_job$\\
\cline{2-5}
& (1) & (2) & (3) & (4)\\
\hline
Treat $\times$ Post  & -0.295*** & -0.192** & -0.489*** & 0.243***\\
                     & (0.0574)   & (0.0873)  & (0.154)   & (0.0713)\\
Constant             & 2.897***  & 4.972*** & 5.443***  & 1.707***\\
                     & (0.0183)   & (0.0279)  & (0.0493)   & (0.0228)\\
\hline
Submarket FE & Yes & Yes & Yes & Yes\\
Year-Month FE & Yes & Yes & Yes & Yes\\
\hline
Observations  & 8{,}412 & 8{,}412 & 8{,}412 & 8{,}412\\
R$^{2}$       & 0.895 & 0.838 & 0.643 & 0.532\\
\hline\hline
\end{tabular}
\begin{tablenotes}
\footnotesize
\centerline{Note: Cluster-robust standard errors are reported at the submarket level. *** $p$ $<$ 0.01, ** $p$ $<$ 0.05, * $p$ $<$ 0.1.}
\end{tablenotes}
\end{table}

\newpage
\section{The Impacts on Image-Related Jobs}
\label{appendix: image}

We further investigate how image-related work is affected by the emergence of generative AI. Specifically, we consider two key events: the release of the first publicly available text-to-image AI model, and the subsequent launch of ChatGPT. To analyze their respective impacts on the image-related submarket, we estimate two regression specifications. Consistent with our earlier submarket construction procedure, we apply the same clustering approach to the skill sets of image-related jobs to define the corresponding submarkets. In this setup, control submarkets serve as the baseline, while the treatment group comprises the image-related submarkets.

For the first specification, we use $Post_{image,t}$ to denote the first publicly available version as the treatment time, set at July 2022 on a monthly level.  For the second specification, we adopt the same $Post$ variable from our main analyses, using the ChatGPT launch date as the start of the treatment period. We define $Treat_{image,i}$ as an indicator variable equal to 1 if unit $i$ corresponds to an image-related submarket and 0 otherwise.
\begin{align}
    &Outcome_{it} = \beta_{0} + \beta_{1} * Treat_{image,i} * Post_{image, t} + u_i + T_t + \epsilon_{it} \label{eq:ddd_image1} \\
    &Outcome_{it} = \beta_{0} + \beta_{1} * Treat_{image,i} * Post_{t} + u_i + T_t + \epsilon_{it}
    \label{eq:ddd_image2}
\end{align}

Equation \ref{eq:ddd_image1} examines changes in demand, supply, and matching outcomes in image-related submarkets relative to control submarkets, in response to the introduction of text-to-image generative AI. This analysis includes data only up to November 2022, thereby excluding the post-ChatGPT period, with the corresponding results reported in Table \ref{tab: Image-related}. In the second analysis, we explore whether the widespread attention generated by ChatGPT led to increased interest in text-to-image AI applications, potentially encouraging broader adoption and influencing labor market outcomes. For this purpose, we extend the observation window to include the post-ChatGPT period and estimate relative changes in image-related submarkets. The results of this spillover analysis are presented in Table \ref{tab: Spillover Effect of Text-related}. 

Both sets of analyses show insignificant results across all measurements. This suggests that current text-to-image generative AI may not
be general-purpose or powerful enough to influence a wide range of jobs in online labor markets.

\newpage
\begin{table}[H]
\centering
\caption{The Effect of Generative AI on Image-related Labor Market}
\label{tab: Image-related}
\newcolumntype{L}[1]{>{\raggedright\arraybackslash}p{#1}}
\newcolumntype{C}[1]{>{\centering\arraybackslash}p{#1}}
\newcolumntype{R}[1]{>{\raggedleft\arraybackslash}p{#1}}
\renewcommand\arraystretch{0.55}
\begin{tabular}{L{3.7cm}C{3cm}C{3cm}C{3cm}C{3cm}}
\hline 
\hline
 & \multicolumn{4}{c}{$ln(y+1)$} \\
 \cline{2-5}
 & $Jobs$ & $Bids$ & $Transaction$ & $Avg\_Bids\_job$ \\
 \cline{2-5}
 & (1) & (2) & (3) & (4) \\
 \hline
 $Treat_{image} * Post_{image}$ & -0.0021 & 0.0854 & -0.150 & 0.0233 \\
 & (0.0402) & (0.0571) & (0.139) & (0.0600) \\
 Constant & 3.047*** & 5.460*** & 5.168*** & 2.192*** \\
 & (0.0075) & (0.0107) & (0.0261) & (0.0113) \\
 \hline
 Submarket FE & Yes & Yes & Yes & Yes \\
 Year-Month FE & Yes & Yes & Yes & Yes \\
 \hline
 Observations & 5,461 & 5,461 & 5,461 & 5,461 \\
 $R^2$ & 0.925 & 0.913 & 0.660 & 0.574 \\
\hline \hline
\end{tabular}
\begin{tablenotes}
\footnotesize
\centerline{Note: Cluster-robust standard errors are reported at the submarket level. *** $p$ $<$ 0.01, ** $p$ $<$ 0.05, * $p$ $<$ 0.1.}
\end{tablenotes}
\end{table}

\begin{table}[H]
\centering
\caption{The Spillover Effect of ChatGPT's Success on Image-related Labor Market}
\label{tab: Spillover Effect of Text-related}
\newcolumntype{L}[1]{>{\raggedright\arraybackslash}p{#1}}
\newcolumntype{C}[1]{>{\centering\arraybackslash}p{#1}}
\newcolumntype{R}[1]{>{\raggedleft\arraybackslash}p{#1}}
\renewcommand\arraystretch{0.55}
\begin{tabular}{L{3.7cm}C{3cm}C{3cm}C{3cm}C{3cm}}
\hline 
\hline
 & \multicolumn{4}{c}{$ln(y+1)$} \\
 \cline{2-5}
 & $Jobs$ & $Bids$ & $Transaction$ & $Avg\_Bids\_job$ \\
 \cline{2-5}
 & (1) & (2) & (3) & (4) \\
 \hline
 $Treat_{image} * Post$ & -0.0536 & -0.0619 & -0.0775 & 0.0683 \\
 & (0.0337) & (0.0470) & (0.0959) & (0.0421) \\
 Constant & 2.983*** & 5.510*** & 5.015*** & 2.303*** \\
 & (0.0071) & (0.0099) & (0.0201) & (0.0088) \\
 \hline
 Submarket FE & Yes & Yes & Yes & Yes \\
 Year-Month FE & Yes & Yes & Yes & Yes \\
 \hline
 Observations & 8,792 & 8,792 & 8,792 & 8,792 \\
 $R^2$ & 0.919 & 0.904 & 0.660 & 0.573 \\
\hline \hline
\end{tabular}
\begin{tablenotes}
\footnotesize
\centerline{Note: Cluster-robust standard errors are reported at the submarket level. *** $p$ $<$ 0.01, ** $p$ $<$ 0.05, * $p$ $<$ 0.1.}
\end{tablenotes}
\end{table}

\newpage
\section{Robustness Check - Alternative Treatment/Control Group Definition Thresholds}\label{appendix:Treatment/Control Group Definition Thresholds}

Given that our submarket classification is based on a conservative approach using the median value of LM-AIOE to distinguish between submarkets with relatively high and low LM-AIOE levels, we acknowledge that observations near the threshold may introduce sensitivity into the estimation. Varying the threshold can alter submarket classifications, particularly for those with LM-AIOE values close to the cutoff. For example, lowering the threshold may lead to some submarkets that were previously assigned to the control group being reclassified into the treatment group. 

To address this concern and enhance the robustness of our findings, we implement an alternative thresholding strategy to ensure that the results are not driven by the specific choice of cutoff. Specifically, we assess the sensitivity of our results by adjusting the original median-based threshold by $\pm$0.05 units. The lower threshold is defined as the median minus 0.05, while the higher threshold is defined as the median plus 0.05. This approach enables us to assess whether submarkets classified as high or low LM-AIOE continue to exhibit consistent patterns that are not influenced by the classification criteria. The estimation results based on these alternative thresholds are presented in Table \ref{tab: main results market-level (Low Threshold)} and Table \ref{tab: main results market-level (High Threshold)}. Table \ref{tab: main results market-level (Low Threshold)} reports the results using the lower threshold, while Table \ref{tab: main results market-level (High Threshold)} presents the results using the higher threshold.

Interestingly, we find that although the estimated coefficients exhibit slight fluctuations compared to those reported in Table \ref{tab: market-level HTE}, they remain statistically significant. Specifically, the outcomes for demand, supply, transaction value consistently show a moderate decline, indicating stable and coherent patterns of heterogeneous treatment effects. These results confirm the robustness and consistency of our findings under alternative threshold specifications.

\newpage

\begin{table}[H]
\centering
\caption{Main Effects Replication with Low Threshold}
\label{tab: main results market-level (Low Threshold)}
\newcolumntype{L}[1]{>{\raggedright\arraybackslash}p{#1}}
\newcolumntype{C}[1]{>{\centering\arraybackslash}p{#1}}
\newcolumntype{R}[1]{>{\raggedleft\arraybackslash}p{#1}}
\renewcommand\arraystretch{0.55}
\begin{tabular}{L{4.9cm}C{2.5cm}C{2.5cm}C{2.5cm}C{2.5cm}}
\hline 
 \hline
 & \multicolumn{4}{c}{$ln(y+1)$} \\
 \cline{2-5}
 & $Jobs$ & $Bids$ & $Transaction$ & $Avg\_Bids\_job$ \\
 \cline{2-5}
 & (1) & (2) & (3) & (4) \\
 \hline
  $Treat * Post$ & -0.220*** & -0.183*** & -0.332*** & 0.130***\\
 & (0.0359) & (0.0535) & (0.101) & (0.0397) \\
 $Treat * Post * Programming$ & 0.0207 & 0.133*** & 0.0523 & 0.109*** \\
 & (0.0224) & (0.0346) & (0.0667) & (0.0267) \\
 Constant & 3.0180*** & 5.393*** & 5.625*** & 2.074*** \\
 & (0.0109) & (0.0162) & (0.0311) & (0.0121) \\
 \hline
 Submarket FE & Yes & Yes & Yes & Yes  \\
 Year-Month FE & Yes & Yes & Yes & Yes \\
 \hline
 Observations & 25,833 & 25,833 & 25,833 & 25,833 \\
 $R^2$ & 0.890 & 0.841 & 0.612 & 0.541 \\
\hline \hline
\end{tabular}
\begin{tablenotes}
\footnotesize
\centerline{Note: Cluster-robust standard errors are reported at the submarket level. *** $p$ $<$ 0.01, ** $p$ $<$ 0.05, * $p$ $<$ 0.1.}
\end{tablenotes}
\end{table}

\begin{table}[H]
\centering
\caption{Main Effects Replication with High Threshold}
\label{tab: main results market-level (High Threshold)}
\newcolumntype{L}[1]{>{\raggedright\arraybackslash}p{#1}}
\newcolumntype{C}[1]{>{\centering\arraybackslash}p{#1}}
\newcolumntype{R}[1]{>{\raggedleft\arraybackslash}p{#1}}
\renewcommand\arraystretch{0.55}
\begin{tabular}{L{5cm}C{2.5cm}C{2.5cm}C{2.5cm}C{2.5cm}}
\hline 
 \hline
 & \multicolumn{4}{c}{$ln(y+1)$} \\
 \cline{2-5}
 & $Jobs$ & $Bids$ & $Transaction$ & $Avg\_Bids\_job$ \\
 \cline{2-5}
 & (1) & (2) & (3) & (4) \\
 \hline
 $Treat * Post$ & -0.198*** & -0.195*** & -0.299*** & 0.0862** \\
 & (0.0339) & (0.0513) & (0.0946) & (0.0372) \\
 $Treat * Post * Programming$ & 0.0294 & 0.149*** & 0.0657 & 0.112*** \\
 & (0.0229) & (0.0352) & (0.0680) & (0.0273) \\
 Constant & 3.0060*** & 5.391*** & 5.608*** & 2.0890*** \\
 & (0.0096) & (0.0145) & (0.0272) & (0.0106) \\
 \hline
 Submarket FE & Yes & Yes & Yes & Yes  \\
 Year-Month FE & Yes & Yes & Yes & Yes \\
 \hline
 Observations & 25,833 & 25,833 & 25,833 & 25,833 \\
 $R^2$ & 0.890 & 0.841 & 0.612 & 0.541 \\
\hline \hline
\end{tabular}
\begin{tablenotes}
\footnotesize
\centerline{Note: Cluster-robust standard errors are reported at the submarket level. *** $p$ $<$ 0.01, ** $p$ $<$ 0.05, * $p$ $<$ 0.1.}
\end{tablenotes}
\end{table}

\newpage
\section{Robustness Check - Fake Treatment Timing (Placebo Test)}\label{appendix:Change Treat Time}

Since the launch of ChatGPT occurred at the end of the year, holidays and seasonal effects might influence the observed post-shock differences. For instance, these differences could be attributed to an alternative mechanism where treated submarkets naturally experience a more pronounced downward trend around the end of the year compared to control submarkets. To address this concern, we conduct another placebo test using November 30, 2021, the same day in the previous year, as a fake shock date. We maintain the time period from September 2021 to August 2022 and rerun all regressions. As shown in Table \ref{tab: Alternative Shock}, our main findings remain robust, suggesting that they are not driven by end-of-year effects or seasonal factors.

\begin{table}[H]
\centering
\caption{Placebo Test with a Fake Shock Time}
\label{tab: Alternative Shock}
\newcolumntype{L}[1]{>{\raggedright\arraybackslash}p{#1}}
\newcolumntype{C}[1]{>{\centering\arraybackslash}p{#1}}
\newcolumntype{R}[1]{>{\raggedleft\arraybackslash}p{#1}}
\renewcommand\arraystretch{0.55}
\begin{tabular}{L{2.7cm}C{2.5cm}C{2.5cm}C{2.5cm}C{2.5cm}}
\hline 
 \hline
 & \multicolumn{4}{c}{$ln(y+1)$} \\
 \cline{2-5}
 & $Jobs$ & $Bids$ & $Transaction$ & $Avg\_Bids\_job$ \\
 \cline{2-5}
 & (1) & (2) & (3) & (4) \\
 \hline
  $Treat * Post$ & -0.0203 & 0.0229 & 0.202 & 0.0706 \\
 & (0.0315) & (0.0540) & (0.128) & (0.0476) \\
 Constant & 3.120*** & 5.345*** & 5.614*** & 1.906*** \\
 & (0.0196) & (0.0336) & (0.0796) & (0.0296) \\
 \hline
 Submarket FE & Yes & Yes & Yes & Yes  \\
 Year-Month FE & Yes & Yes & Yes & Yes \\
 \hline
 Observations & 11,769 & 11,769 & 11,769 & 11,769 \\
 $R^2$ & 0.928 & 0.873 & 0.666 & 0.606 \\
\hline \hline
\end{tabular}
\begin{tablenotes}
\footnotesize
\centerline{Note: Cluster-robust standard errors are reported at the submarket level. *** $p$ $<$ 0.01, ** $p$ $<$ 0.05, * $p$ $<$ 0.1.}
\end{tablenotes}
\end{table}

\newpage
\section{Robustness Check - Treatment Reassignment (Placebo Tests)}\label{appendix:Placebo Tests}

Statistically, there is a possibility that the observed effects are driven by random variations across observations. To address this, we conduct a placebo test by randomly reassigning the treatment ($Treat * Post$) within our sample. We simulate this permutation procedure 1,000 times and capture the distribution of the placebo treatment effects for all six main outcome variables based on random treatment assignment. The distribution of placebo treatment effects resembles a normal distribution, as shown in Figure \ref{fig:Placebo Tests1}, and our estimated coefficient lies far in the tail of this distribution. This placebo test indicates that the treatment effects are not driven by chance, further reinforcing the robustness of our findings.

\begin{figure}[H]
    \centering
    \subfloat[Jobs]{\includegraphics[width=0.4\textwidth]{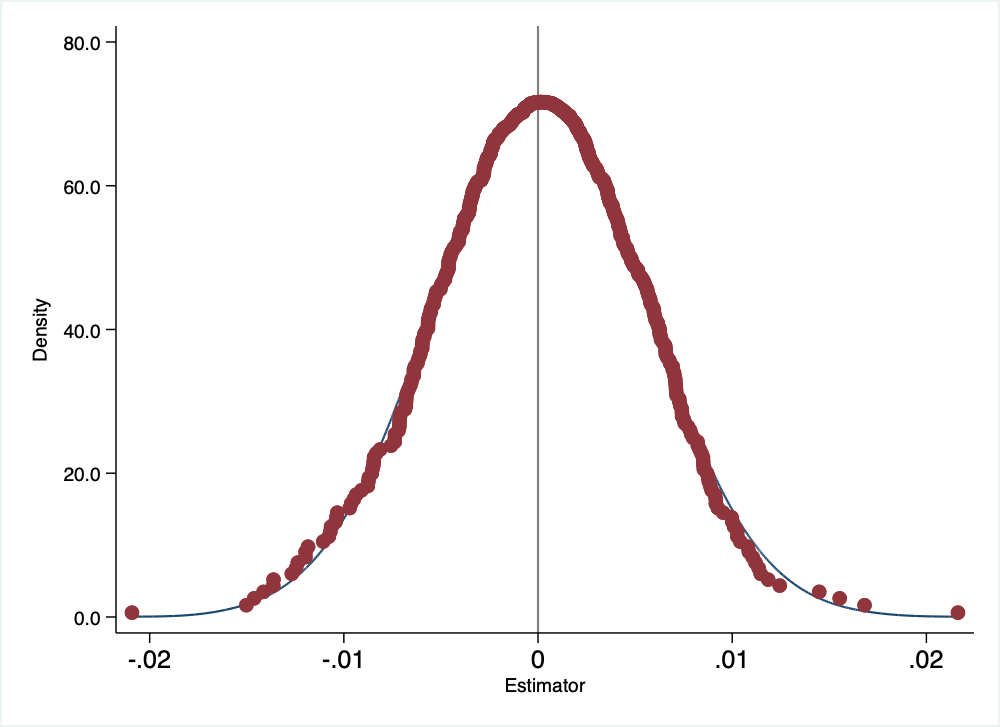}\label{subfig:jobs1}}\quad
    \subfloat[Bids]{\includegraphics[width=0.4\textwidth]{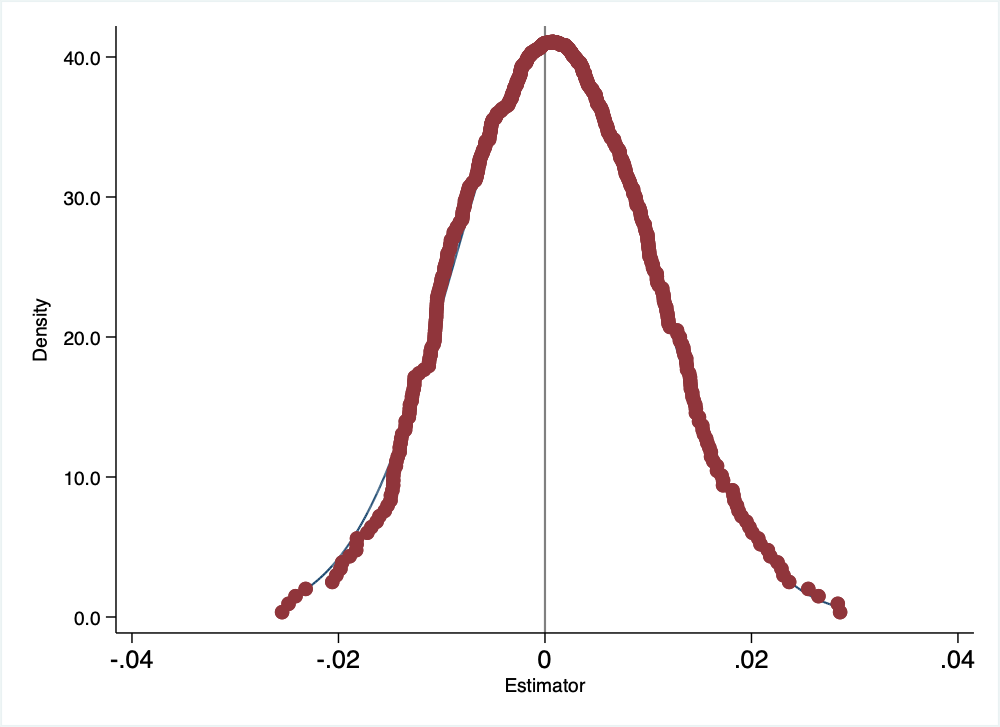}\label{subfig:bids1}}\quad
    \subfloat[Transaction]{\includegraphics[width=0.4\textwidth]{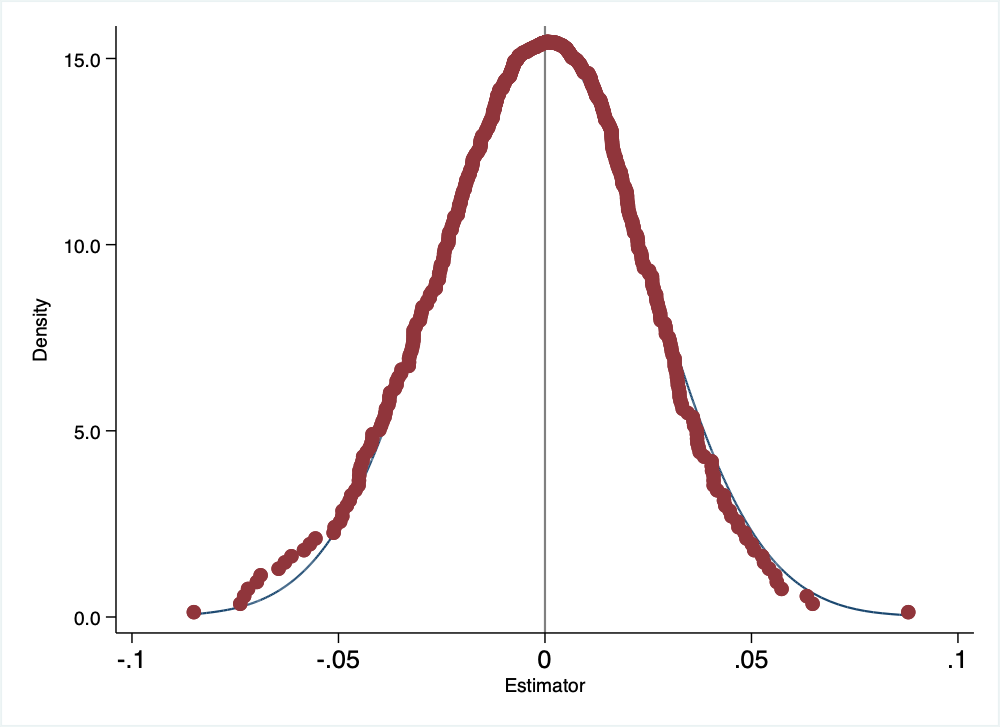}\label{subfig:transaction1}}\quad
    \subfloat[Average\_job\_bids]{\includegraphics[width=0.4\textwidth]{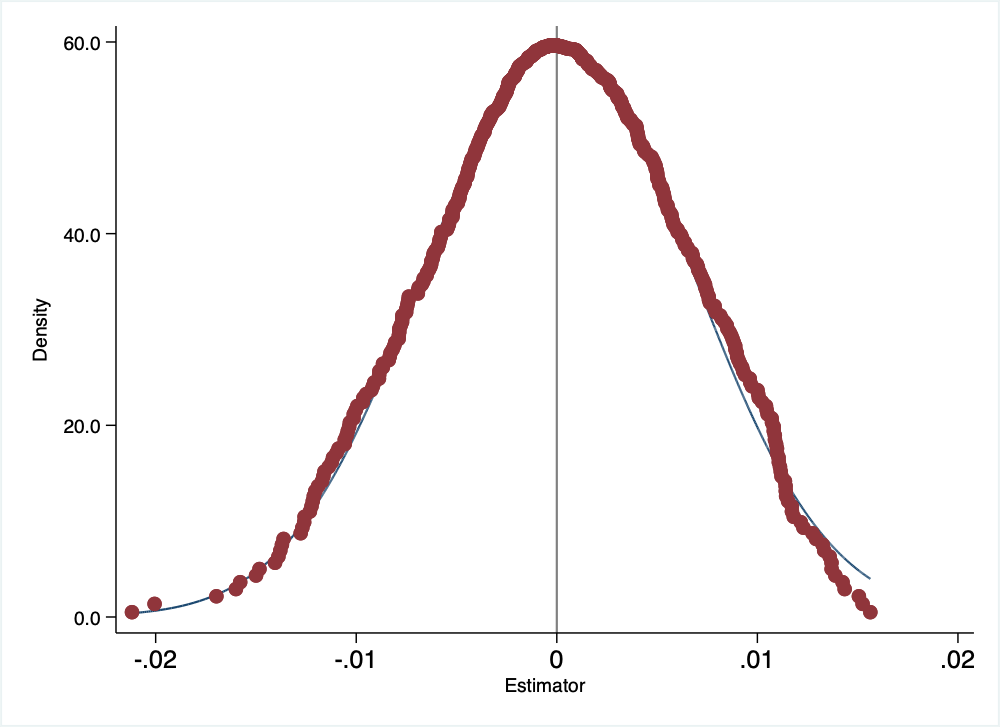}\label{subfig:success1}}
    \caption{\centering Placebo Tests} 
    \label{fig:Placebo Tests1}
\end{figure}

\newpage
\section{Robustness Check - Pretrends and Temporal Dynamics}\label{appendix:Pretrends and Temporal Dynamics}

We use a DiD specification for most analyses to uncover how generative AI introduces different trends for treated submarkets versus control submarkets. To attribute post-shock differences to the shock itself, we first need to examine whether these two sets of submarkets follow similar trends before the shock (parallel trend assumption). Additionally, we are interested in whether the observed effects appear primarily at the beginning of the shock (due to novelty effects, etc.) and attenuate over time, or whether these differences strengthen over time. With these objectives in mind, we employ the relative time model to obtain per-period effect estimates, following previous scholarly investigations \citep{braghieri2022social, perez2020effects}.
\begin{align}
Outcome_{it} 
&= \sum_{k \neq -1} \beta_k \, D_{i,t=k} + u_i + T_t + \epsilon_{it} \nonumber \\
&= \sum_{k \neq -1} \beta_k \left( D_i \cdot \mathbf{1}\{t - \tau_i = k\} \right) + u_i + T_t + \epsilon_{it},
\label{equ:relative_time}
\end{align}

where \(D_{i,t=k}\) equals 1 if unit \(i\) is treated and period \(t\) is \(k\) periods relative to unit \(i\)'s event (adoption) time \(\tau_i\), and 0 otherwise; the reference period \(k=-1\) is omitted. Equivalently, \(D_{i,t=k}\) can be written as \(D_i \cdot \mathbf{1}\{t - \tau_i = k\}\), where \(D_i\) indicates whether unit \(i\) is from the treatment group and \(\mathbf{1}\{\cdot\}\) is the indicator function. \(u_i\) and \(T_t\) denote unit and time fixed effects, respectively, and \(\epsilon_{it}\) is the idiosyncratic error term.

To estimate the relative time model, we utilize the full panel of data. Following common practice in event study analyses \citep{schmidheiny2023event}, we merge all periods prior to event time $k = -8$ into a single bin at $k = -8$, and here $k = +8$ is the final time period. This binning approach reduces noise and ensures a more balanced panel. Importantly, our main findings remain robust without this binning: even when all relative time periods are included without aggregation, we continue to find no evidence of differential pre-trends.

The results are shown in Figures \ref{fig:Pretrends and Temporal Dynamics1} and \ref{fig:Pretrends and Temporal Dynamics2}. First, the confidence intervals for the periods leading up to the launch of ChatGPT all include zero, indicating no significant pre-existing trend differences. This finding strengthens the credibility of attributing the observed post-shock differences to the launch of LLM-based generative AI. Second, the figure reveals an increasing impact over time in the post-treatment periods, suggesting that the influence of these technologies is not a one-time event but deepens with increasing adoption.

\newpage
\begin{figure}[H]
    \centering
    \subfloat[Jobs]{\includegraphics[width=5.5in, height=3in]{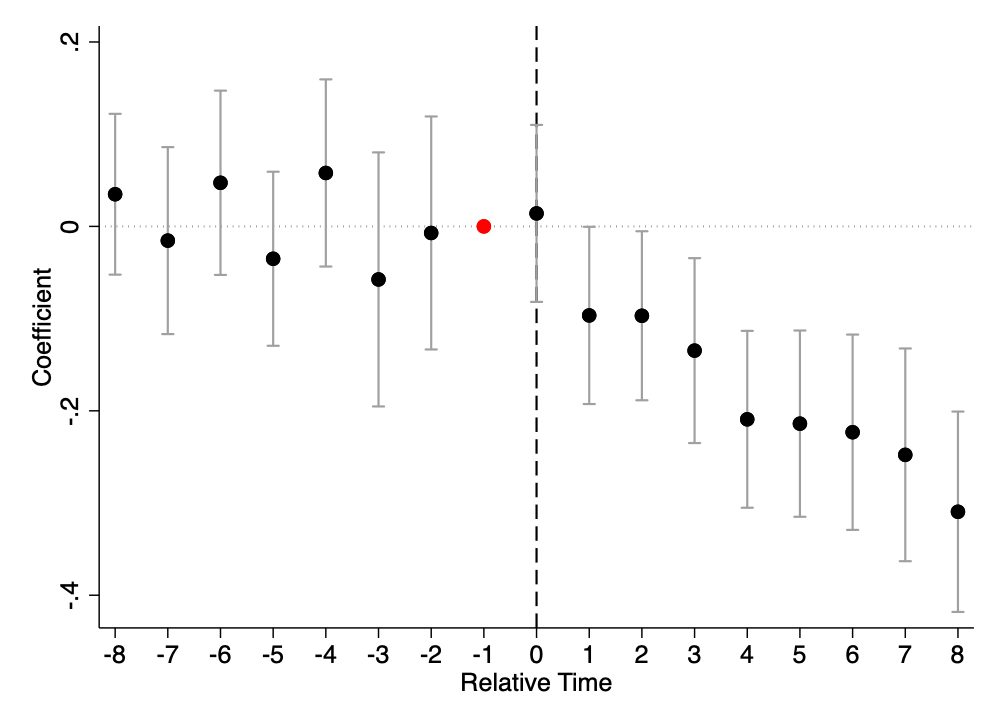}\label{subfig:jobs0}}\quad
    \subfloat[Bids]{\includegraphics[width=5.5in, height=3in]{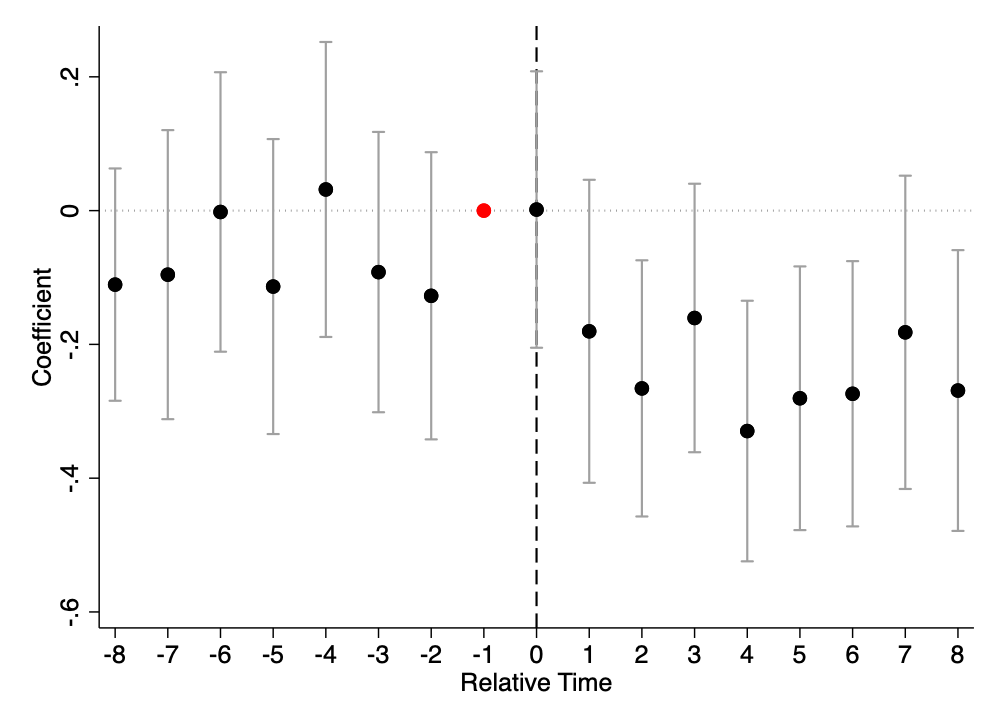}\label{subfig:bids0}}\quad
    \caption{\centering Pretrends and Temporal Dynamics in Demand and Supply} 
    \label{fig:Pretrends and Temporal Dynamics1}
\end{figure}

\newpage
\begin{figure}[H]
    \centering
    \subfloat[Transaction]{\includegraphics[width=5.5in, height=3inh]{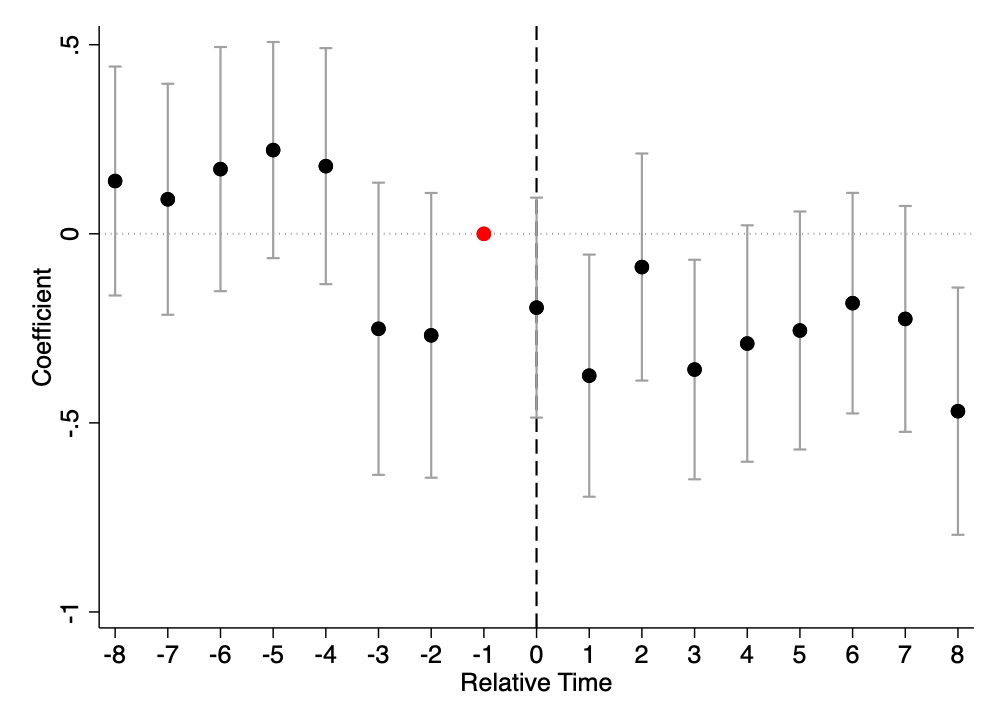}\label{subfig:transaction}}\quad
    \subfloat[Avg\_Bids\_job]{\includegraphics[width=5.5in, height=3in]{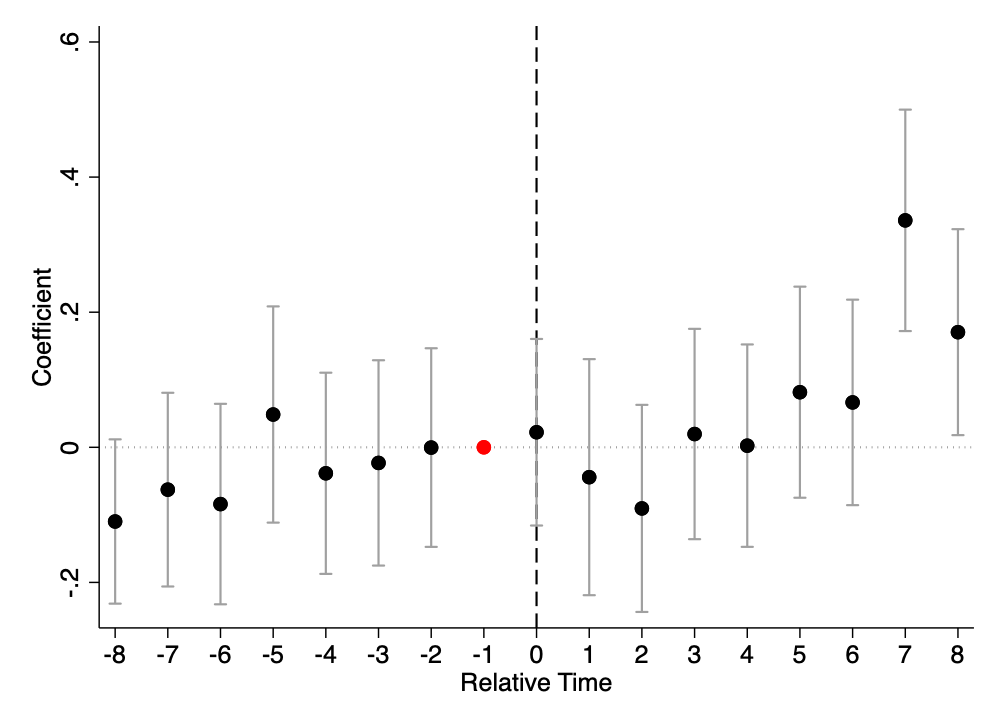}\label{subfig:avg_bid}}
    \caption{\centering Pretrends and Temporal Dynamics in Matching Outcomes} 
    \label{fig:Pretrends and Temporal Dynamics2}
\end{figure}

\newpage
\section{Robustness Check - Propensity Score Matching}\label{appendix:PSM}

Although the treated and control submarkets follow similar pre-shock trends, as demonstrated in Appendix \ref{appendix:Pretrends and Temporal Dynamics}, they may still differ across various dimensions such as market size prior to the treatment. To address this concern and enhance the comparability between the two sets of submarkets, we re-estimate our analysis using a propensity score matched sample of treated and control submarkets. Specifically, we estimate a probit selection model for each submarket to predict the probability of receiving treatment. To improve the comparability of the matched submarkets, we include all relevant pre-treatment variables ($Jobs$, $Clients$, $Bids$, $Bidders$, $Transaction$, $Matches$, and $Avg\_Bids\_job$) as covariates in the matching process. Based on the predicted propensity scores, we apply a nearest-neighbor matching procedure following standard practices in the literature \citep{ran2025sometimes,ye2025close,faccio2021business} to identify the closest control submarket for each treated submarket. The resulting matched sample consists of 952 submarkets.

Next, we evaluate the quality of the matches by comparing the treated and control submarkets in terms of the covariates used in the matching procedure. Table \ref{tab: Statistical Validation of Pre- and Post-Matching Balance} presents the mean values of the matching variables. In contrast to the full sample results shown in columns (2)--(4), columns (5)--(7) reveal no statistically significant differences between the treated and control submarkets. This indicates that the propensity score matching procedure effectively mitigates key pre-treatment differences between the two groups. Additionally, Figures \ref{fig:psm-1} and \ref{fig:psm-2} illustrates the distribution of propensity scores before and after matching, demonstrating that the matched sample is substantially more balanced.

In Table \ref{tab: main results market-level (PSM)}, we re-estimate our main regression model using the matched submarket sample. Following the specification outlined in Section \ref{sec: econometric}, the coefficient on the interaction term $Treat \times Post$ ($\beta_1$) remains negative and statistically significant. These results are consistent with our baseline findings. Overall, the robustness checks suggest that our main results are not driven by insufficient matching quality.

\newpage

\begin{table}[H]
\centering
\caption{Statistical Validation of Pre- and Post-Matching Balance}
\label{tab: Statistical Validation of Pre- and Post-Matching Balance}
\newcolumntype{L}[1]{>{\raggedright\arraybackslash}p{#1}}
\newcolumntype{C}[1]{>{\centering\arraybackslash}p{#1}}
\newcolumntype{R}[1]{>{\raggedleft\arraybackslash}p{#1}}
\renewcommand\arraystretch{0.55}
\resizebox{\textwidth}{!}{%
\begin{tabular}{L{2.8cm}C{1.9cm}C{1.9cm}C{1.9cm}C{1.9cm}C{1.9cm}C{1.9cm}}
\hline \hline 
 & \multicolumn{3}{c}{Before-Matching} &  \multicolumn{3}{c}{After-Matching} \\
 \hline
 Variables & Control & Treat & P-Value & Control & Treat & P-Value \\
 \hline
 $Jobs$ & 24.38 & 47.49 & 0.000 & 28.65 & 35.68 & 0.112 \\
 $Clients$ & 21.30 & 41.17 & 0.000 & 25.03 & 31.04 & 0.110 \\
 $Bids$ & 406.14 & 774.44 & 0.002 & 490.80 & 591.41 & 0.349 \\
 $Bidders$ & 234.65 & 443.35 & 0.001 & 281.71 & 363.47 & 0.189 \\
 $Matches$ & 3.28 & 6.35 & 0.000 & 3.85 & 4.71 & 0.199 \\
 $Transaction$ & 1677.13 & 3175.91 & 0.000 & 1995.10 & 2366.75 & 0.286 \\
 $Avg\_Bids\_job$ & 8.30 & 10.00 & 0.004 & 9.42 & 10.00 & 0.375 \\
\hline \hline
\end{tabular}%
}
\end{table}

\begin{table}[H]
\centering
\caption{Effects of Generative AI on Labor Market Outcomes (After PSM)}
\label{tab: main results market-level (PSM)}
\newcolumntype{L}[1]{>{\raggedright\arraybackslash}p{#1}}
\newcolumntype{C}[1]{>{\centering\arraybackslash}p{#1}}
\newcolumntype{R}[1]{>{\raggedleft\arraybackslash}p{#1}}
\renewcommand\arraystretch{0.55}
\begin{tabular}{L{2.7cm}C{2.5cm}C{2.5cm}C{2.5cm}C{2.5cm}}
\hline 
 \hline
 & \multicolumn{4}{c}{$ln(y+1)$} \\
 \cline{2-5}
 & $Jobs$ & $Bids$ & $Transaction$ & $Avg\_Bids\_job$ \\
 \cline{2-5}
 & (1) & (2) & (3) & (4) \\
 \hline
  $Treat * Post$ & -0.213*** & -0.159** & -0.362*** & 0.112** \\
 & (0.0471) & (0.0694) & (0.117) & (0.0443) \\
 Constant & 2.807*** & 5.162*** & 5.275*** & 2.129*** \\
 & (0.0089) & (0.0131) & (0.0221) & (0.0084) \\
 \hline
 Submarket FE & Yes & Yes & Yes & Yes  \\
 Year-Month FE & Yes & Yes & Yes & Yes \\
 \hline
 Observations & 38,962 & 38,962 & 38,962 & 38,962 \\
 $R^2$ & 0.871 & 0.799 & 0.584 & 0.474 \\
\hline \hline
\end{tabular}
\begin{tablenotes}
\footnotesize
\centerline{Note: Cluster-robust standard errors are reported at the submarket level. *** $p$ $<$ 0.01, ** $p$ $<$ 0.05, * $p$ $<$ 0.1.}
\end{tablenotes}
\end{table}

\begin{figure}[H]
\begin{center}
\includegraphics[width=5.5in, height=3in]{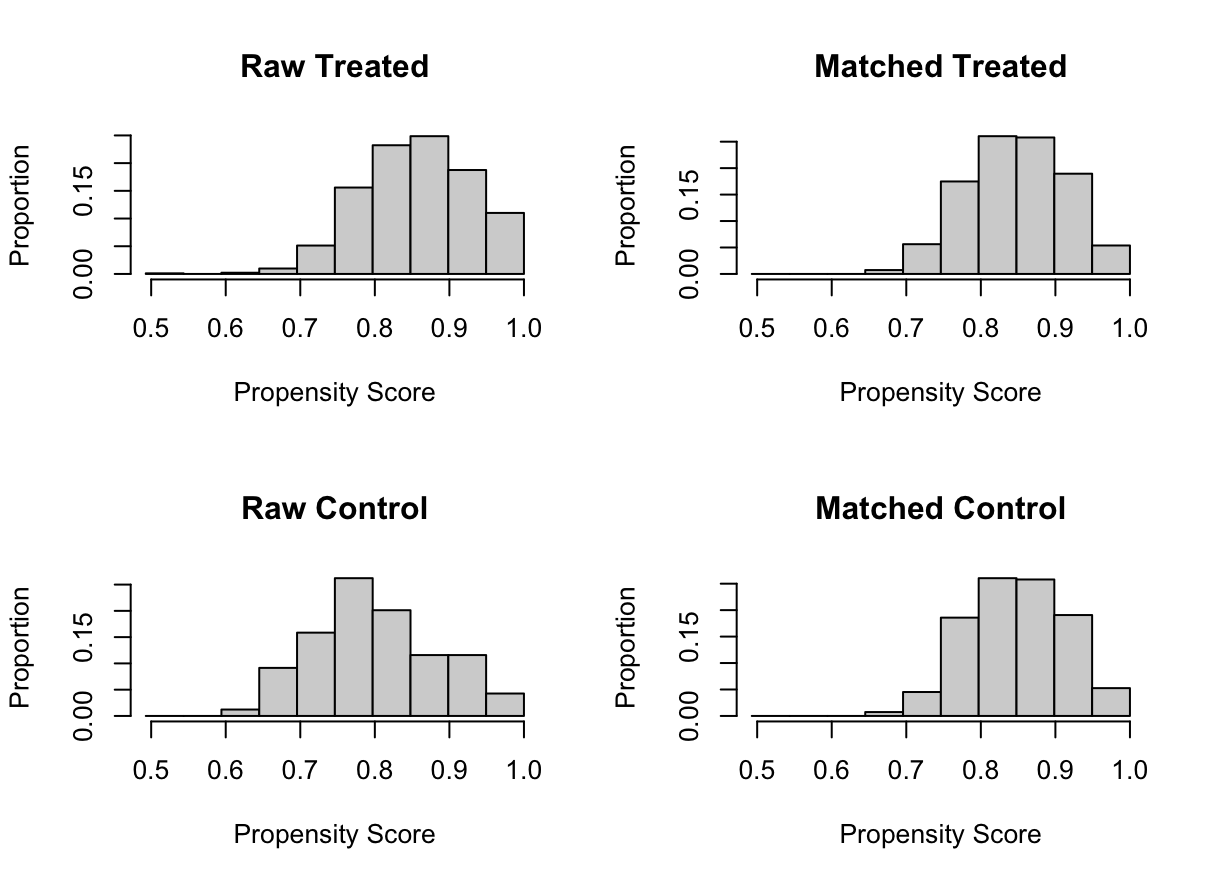}
\caption{\centering Propensity Score Distribution Pre- and Post-Matching}
\label{fig:psm-1}
\end{center}
\end{figure}

\begin{figure}[H]
\begin{center}
\includegraphics[width=5.5in, height=3in]{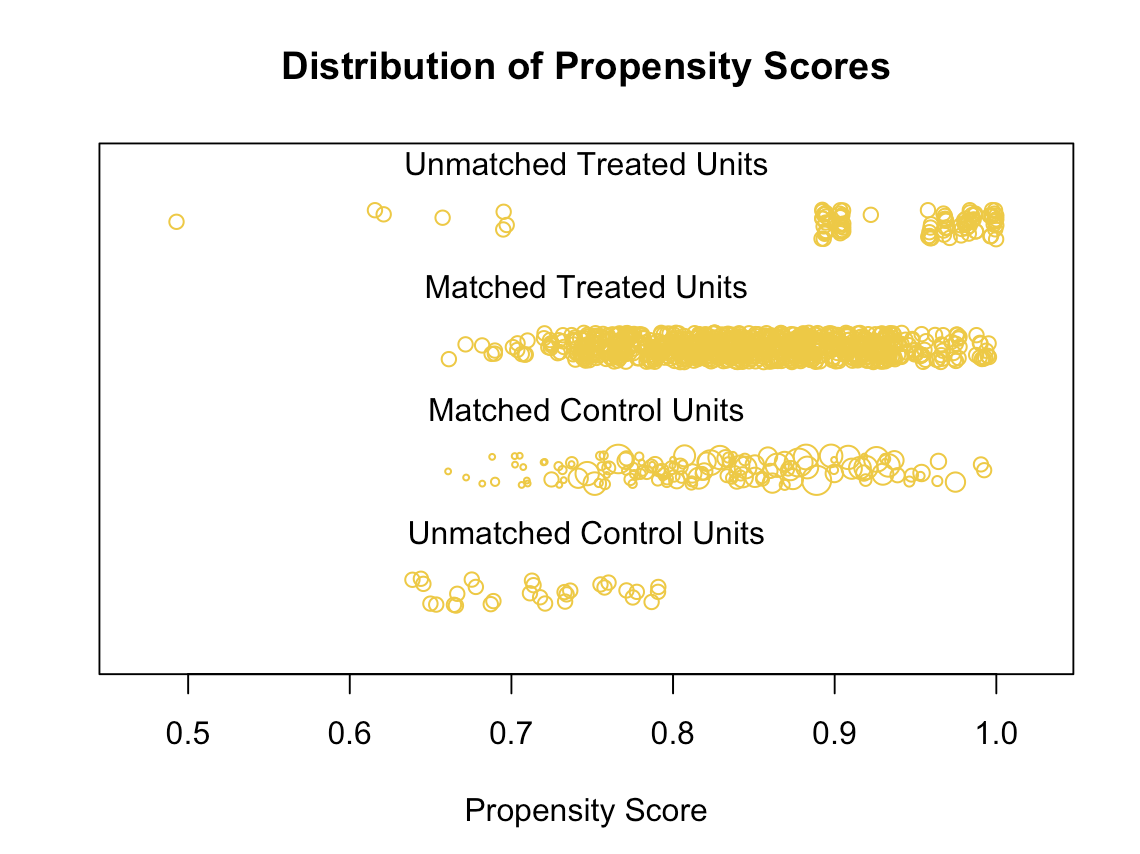}
\caption{\centering Comparison Between Matched and Unmatched Units in Propensity Score Matching}
\label{fig:psm-2}
\end{center}
\end{figure}

\newpage
\section{Robustness Check - Interrupted Time Series Analysis (ITSA)}\label{appendix:ITS}

ITSA is a quasi-experimental approach that projects the pre-intervention trend forward to form a counterfactual baseline, and then measures the intervention's effect as the deviation of actual post-intervention outcomes from that projected path \citep{kontopantelis2015regression,campbell2015experimental}. We estimate the regression model as follows:
\begin{align}\label{eq:its}
    &Outcome_{it} = \beta_0 + \beta_{1} * Time_t + \beta_2 * Post_t + \beta_3 * PostTime_t + u_i + \epsilon_{it} 
\end{align}

Here, $Outcome_{it}$ denotes the dependent variable ($Jobs$, $Bids$, $Transaction$, or $Avg\_Bids\_job$) for submarket $i$ in month $t$. $Time_t$ represents the time elapsed since the beginning of the observation period. $Post_t$ is a binary indicator equal to 0 during the pre-treatment period and 1 during the post-treatment period. $PostTime_t$ equals 0 prior to the treatment (i.e., the launch of ChatGPT) and increases by one for each subsequent time period following the treatment. Furthermore, we incorporate $u_i$ for the unit-level fixed effect. The error term $\epsilon_{it}$ encapsulates unobserved factors that influence the outcome. Our analysis centers on two coefficients of interest, $\beta_2$ and $\beta_3$. The coefficient on $Post_t$ ($\beta_2$) captures the immediate effect of the treatment: a statistically significant value indicates a discrete level change in the outcome attributable to the emergence of ChatGPT. The coefficient on $PostTime_t$ ($\beta_3$) captures the trend effect, reflecting the change in the slope of the outcome trajectory following the intervention. Distinguishing the two matters because the influence of LLM-based generative AI may manifest both as an abrupt shift at the shock and as a gradual change that accumulates over the post-treatment period.

We estimate the model specified in Equation \ref{eq:its} and report the results in Table \ref{tab: ITS}. The findings align closely with those from the DiD estimation. Specifically, labor demand ($Jobs$) and market transaction volume ($Transaction$) exhibit a significant immediate decline following the treatment ($\beta_2 = -0.101$ and $-0.287$, respectively; $p < 0.01$), and labor demand also continues along a significant downward trend thereafter ($\beta_3 = -0.0490$ for $Jobs$; $p < 0.01$). On the supply side, by contrast, labor supply ($Bids$) shows no significant immediate response ($\beta_2 = -0.0055$, $p > 0.1$) and instead declines only gradually, as captured by a significant negative trend ($\beta_3 = -0.0361$, $p < 0.01$) whose absolute magnitude is smaller than its demand-side counterpart. As a result, competition within the submarket intensifies, reflected in a significant rise in the average number of bids per job ($Avg\_Bids\_job$; $\beta_2 = 0.0726$, $p < 0.01$). Consistent with our main analysis, these dynamics point to an asymmetric adjustment across the two sides of the market: demand contracts almost at once as some clients substitute the technology for freelance labor, whereas the supply side responds with a lag, with freelancers recalibrating their participation only gradually. Figures \ref{fig: Interrupted Time Series (ITS) Analysis1} and \ref{fig: Interrupted Time Series (ITS) Analysis2} provide a more detailed and visually intuitive representation of these trends. To further validate our findings, we also implement a regression discontinuity design (RDD) to estimate the immediate effect at the shock point, and the results are consistent with those from the ITSA.

\begin{table}[H]
\centering
\caption{Results for Interrupted Time Series Analysis}
\label{tab: ITS}
\newcolumntype{L}[1]{>{\raggedright\arraybackslash}p{#1}}
\newcolumntype{C}[1]{>{\centering\arraybackslash}p{#1}}
\newcolumntype{R}[1]{>{\raggedleft\arraybackslash}p{#1}}
\renewcommand\arraystretch{0.55}
\begin{tabular}{L{2.7cm}C{2.5cm}C{2.5cm}C{2.5cm}C{2.5cm}}
\hline 
 \hline
 & \multicolumn{4}{c}{$ln(y+1)$} \\
 \cline{2-5}
 & $Jobs$ & $Bids$ & $Transaction$ & $Avg\_Bids\_job$ \\
 \cline{2-5}
 & (1) & (2) & (3) & (4) \\
 \hline
  $Time$ & 0.0049* & 0.0143*** & -0.0193 & 0.0132** \\
 & (0.0028) & (0.0055) & (0.0144) & (0.0053) \\
 $Post$ & -0.101*** & -0.0055 & -0.287*** & 0.0726*** \\
 & (0.0128) & (0.0248) & (0.0715) & (0.0240) \\
 $PostTime$ & -0.0490*** & -0.0361*** & -0.0177 & 0.0456*** \\
 & (0.0040) & (0.0075) & (0.0184) & (0.0076) \\
 Constant & 3.0950*** & 5.553*** & 5.997*** & 2.0840*** \\
 & (0.0356) & (0.0496) & (0.0795) & (0.0282) \\
 \hline
 Submarket FE & Yes & Yes & Yes & Yes  \\
 \hline
 Observations & 11,921 & 11,921 & 11,921 & 11,921 \\
\hline \hline
\end{tabular}
\begin{tablenotes}
\footnotesize
\centerline{Note: Cluster-robust standard errors are reported at the submarket level. *** $p$ $<$ 0.01, ** $p$ $<$ 0.05, * $p$ $<$ 0.1.}
\end{tablenotes}
\end{table}

\newpage

\begin{figure}[H]
    \centering
    \subfloat[Jobs]{\includegraphics[width=5.5in, height=3in]{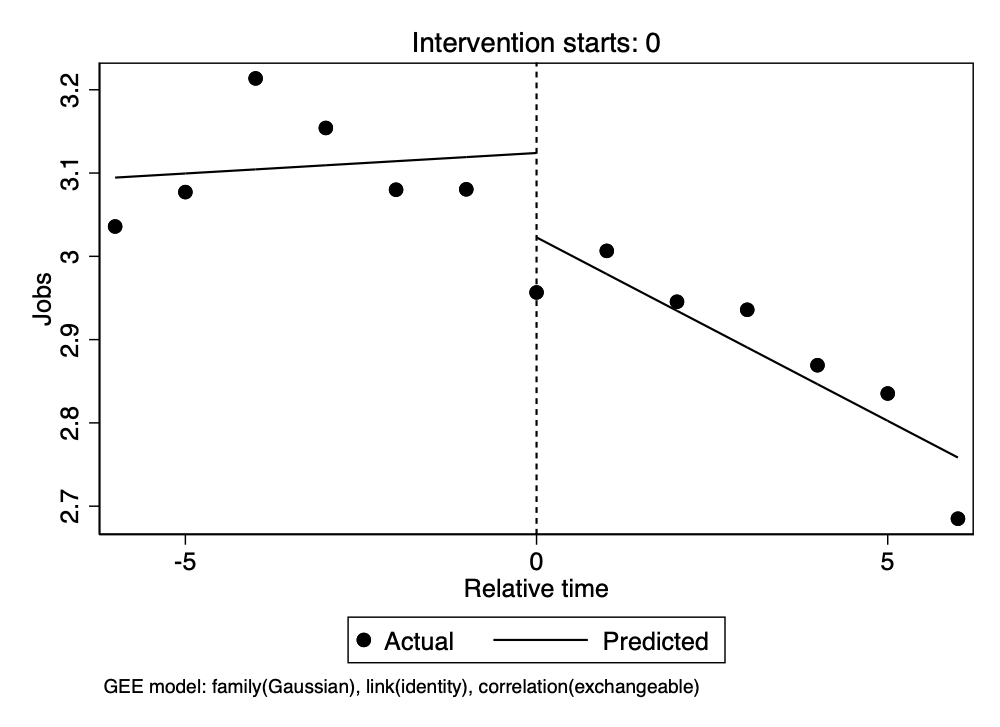}\label{subfig:jobs3}}\quad
    \subfloat[Bids]{\includegraphics[width=5.5in, height=3in]{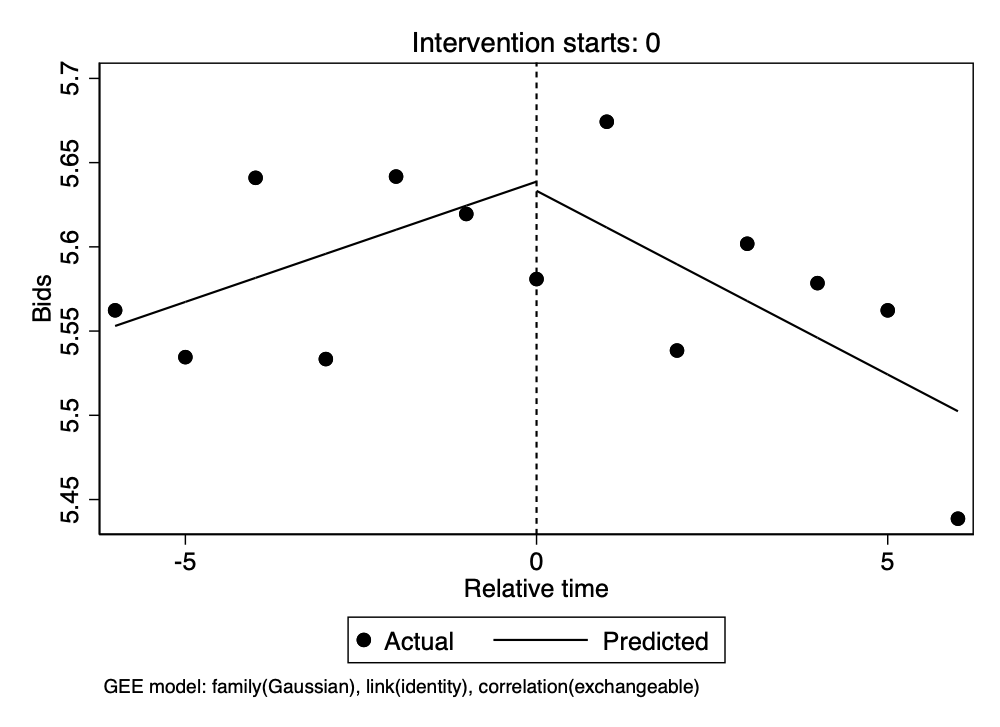}\label{subfig:bids3}}\quad
    \caption{\centering Interrupted Time Series Analysis (ITSA)}
    \label{fig: Interrupted Time Series (ITS) Analysis1}
\end{figure}

\begin{figure}[H]
    \centering
    \subfloat[Transaction]{\includegraphics[width=5.5in, height=3in]{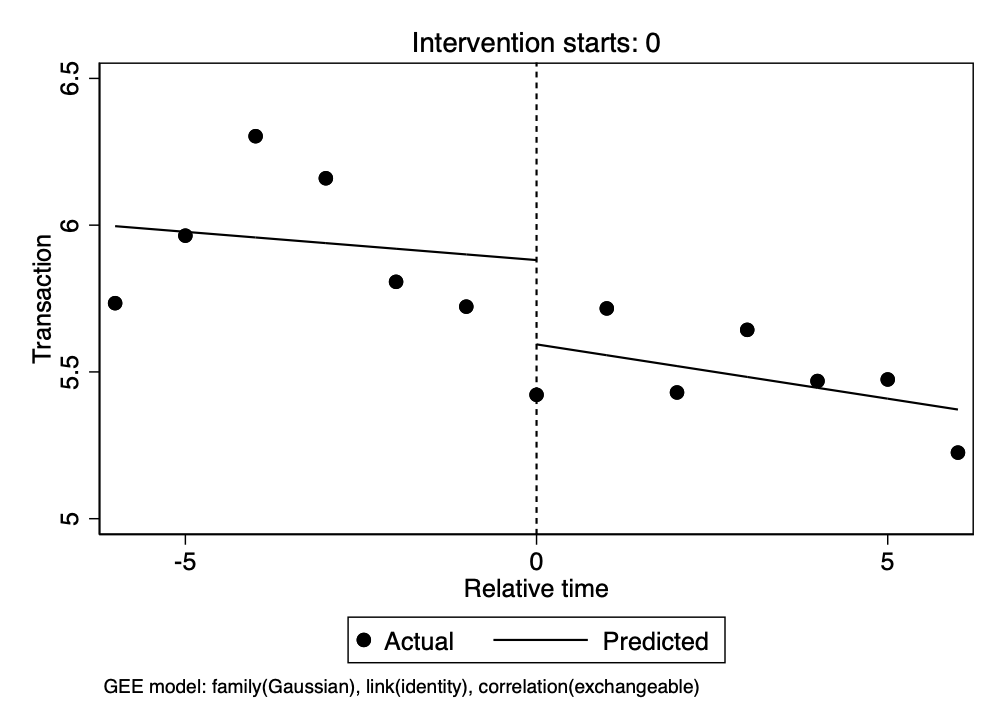}\label{subfig:transaction3}}\quad
    \subfloat[Average\_job\_bids]{\includegraphics[width=5.5in, height=3in]{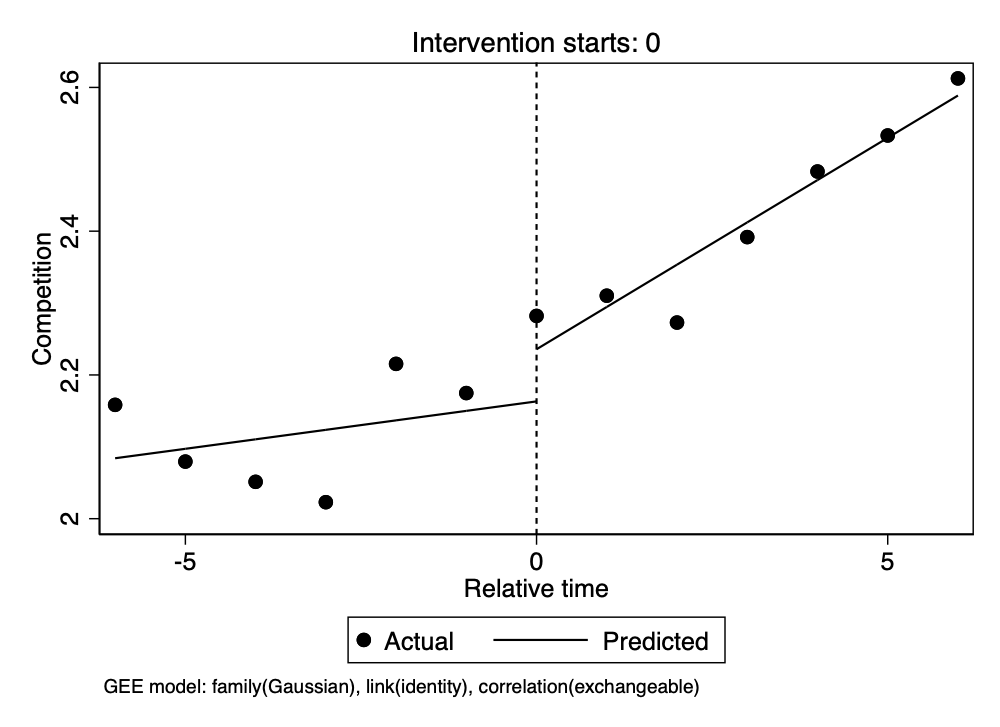}\label{subfig:success3}}
    \caption{\centering Interrupted Time Series Analysis (ITSA)}
    \label{fig: Interrupted Time Series (ITS) Analysis2}
\end{figure}

\newpage
\section{Robustness Check - More Details on Control Variable Inclusion and Omitted Variable Sensitivity Analyses}\label{appendix:Potential Confounders Control}

In this section, we incorporate control variables that capture both external conditions and internal market characteristics to obtain a conservative estimate of ChatGPT's impact and to conduct a sensitivity analysis.

Since online labor markets may be affected by external factors such as trends in the broader employment landscape, we account for these influences by collecting monthly Google Trends data for each skill, which reflects the relative popularity of each skill in a given month. Using the skill composition of each submarket, we construct a weighted monthly Google Trends index for each submarket to capture and control for fluctuations in the external environment. In addition, within the labor market, demand and supply interact dynamically. Demand in one period may influence supply in the subsequent period, while supply may, in turn, affect future demand. These two forces jointly determine the market transaction value and competition in the periods that follow. Therefore, when estimating the effect on demand ($Jobs$), we include lagged supply ($Lag\_Bids$) as a control variable. Similarly, when estimating the effect on supply ($Bids$), we control for lagged demand ($Lag\_Jobs$). In the estimation of transaction value and competition, both lagged demand and lagged supply ($Lag\_Jobs$ and $Lag\_Bids$) are included as control variables to account for their joint influence. The estimation results are presented in Table \ref{tab: Potential Confounders Control} and remain consistent with our main findings.

After incorporating these observable controls, the estimated coefficients $\beta_1$ change from -0.221 to -0.222 for $Jobs$, from -0.168 to -0.141 for $Bids$, from -0.315 to -0.267 for $Transaction$, and from 0.138 to 0.111 for $Avg\_Bids\_job$.

Including post-treatment covariates is likely to attenuate the estimated treatment effect. A consistent estimator should capture the total influence of ChatGPT on the outcome, that is, it should capture the cumulative effect of all underlying mechanisms. In practice, ChatGPT may first alter demand and subsequently affect labor supply and transaction volume. Competition, in particular, can rise both because the number of available jobs falls and because changes in job mix propagate through bidding behavior before they are reflected in competitive metrics.  Variables such as $Jobs$ or $Bids$ therefore lie on the causal pathway from the treatment to the final outcome, making them endogenous to ChatGPT adoption. Conditioning on such mediators severs part of the causal chain and introduces the classic ``bad-control" problem, which mechanically biases the treatment coefficient toward zero. Hence, our reported estimates should be interpreted as conservative lower bounds; even so, they remain both economically meaningful and statistically significant.

Furthermore, to evaluate the risk of omitted variable bias, we conduct a sensitivity analysis following the approach proposed by \cite{altonji2005selection}, which has been widely applied in prior empirical studies \citep{sen2024does, petrova2021social, galor2016agricultural}. Specifically, we assess the relative sensitivity of unobserved factors by incorporating these additional control variables, thereby generating an alternative set of estimates. We then compare the new coefficient to that obtained from the regression without the additional controls, using the ratio defined as $\dfrac{\beta_{original}}{\beta_{original}-\beta_{control}}$. The rationale behind this ratio is straightforward: a smaller difference between the uncontrolled and controlled estimates suggests that the parameter of interest is less affected by selection on observables, thus requiring a stronger degree of selection on unobservables relative to observables to fully explain the observed effect. In our estimated result, the similarity of the coefficients, suggests that selection on unobservables would need to be at least 5 times \citep{oster2019unobservable} stronger than selection on observables to fully account for the observed effect. This substantially alleviates concerns about omitted variable bias.

\begin{table}[H]
\centering
\caption{Effects of Generative AI on Demand, Supply, and Matching Outcomes}
\label{tab: Potential Confounders Control}
\newcolumntype{L}[1]{>{\raggedright\arraybackslash}p{#1}}
\newcolumntype{C}[1]{>{\centering\arraybackslash}p{#1}}
\newcolumntype{R}[1]{>{\raggedleft\arraybackslash}p{#1}}
\renewcommand\arraystretch{0.55}
\begin{tabular}{L{5.2cm}C{2.5cm}C{2.5cm}C{2.5cm}C{2.5cm}}
\hline 
 \hline
 & \multicolumn{4}{c}{$ln(y+1)$} \\
 \cline{2-5}
 & $Jobs$ & $Bids$ & $Transaction$ & $Avg\_Bids\_job$ \\
 \cline{2-5}
 & (1) & (2) & (3) & (4) \\
 \hline
  $Treat * Post$ & -0.222*** & -0.141*** & -0.267*** & 0.111*** \\
 & (0.0316) & (0.0469) & (0.0886) & (0.0355) \\
 Constant & 2.234*** & 3.958*** & 2.972*** & 1.904*** \\
 & (0.181) & (0.309) & (0.651) & (0.225) \\
 \hline
 External Confounder Controls & Yes & Yes & Yes & Yes  \\
 Internal Confounder Controls & Yes & Yes & Yes & Yes  \\
 \hline
 Submarket FE & Yes & Yes & Yes & Yes  \\
 Year-Month FE & Yes & Yes & Yes & Yes \\
 \hline
 Observations & 25,833 & 25,833 & 25,833 & 25,833 \\
 $R^2$ & 0.892 & 0.842 & 0.614 & 0.545 \\
\hline \hline
\end{tabular}
\begin{tablenotes}
\footnotesize
\centerline{Note: Cluster-robust standard errors are reported at the submarket level. *** $p$ $<$ 0.01, ** $p$ $<$ 0.05, * $p$ $<$ 0.1.}
\end{tablenotes}
\end{table}

\newpage
\section{Robustness Check - Alternative Model for Count Variables}\label{appendix:Alternative model}

It is noteworthy that several of our variables, including $Jobs$ and $Bids$, are count variables, taking on non-negative integer values. As shown in Appendix \ref{appendix: Variables Statistics}, the variables exhibit substantial skewness, rendering count models not perfectly appropriate. Since the Negative Binomial model is relatively better suited for fitting such data distributions compared to the Poisson model \citep{alma99203850030001451}, we employ it to re-estimate the impact on these count variables. Tables \ref{tab: Alternative Model} present the results of this analysis. The estimation results are qualitatively aligned with our main findings, corroborating the robustness of our conclusions. Specifically, we observe a significant decrease in labor demand in programming-intensive or non-programming-intensive submarkets, with no heterogeneity observed between these two types of treated submarkets. On the labor supply side, we still observe a significant decrease in non-programming-intensive submarkets, while the programming-intensive submarkets incur a significantly smaller supply loss, particularly in terms of the number of bids. The estimation results are still consistent with our main findings.

\begin{table}[H]
\centering
\caption{Alternative Model for Count Variables}
\label{tab: Alternative Model}
\newcolumntype{L}[1]{>{\raggedright\arraybackslash}p{#1}}
\newcolumntype{C}[1]{>{\centering\arraybackslash}p{#1}}
\newcolumntype{R}[1]{>{\raggedleft\arraybackslash}p{#1}}
\renewcommand\arraystretch{0.55}
\begin{tabular}{L{4.9cm}C{4cm}C{4cm}}
\hline 
\hline
 & \multicolumn{2}{c}{$Negative \ Binominal \ Regressions$} \\
 \cline{2-3}
 & $Jobs$ & $Bids$ \\
 \cline{2-3}
 & (1) & (2) \\
 \hline
 $Treat * Post$ & -0.189*** & -0.207*** \\
 & (0.0437) & (0.0500) \\
 $Treat * Post * Programming$ & 0.0070 & 0.143*** \\
 & (0.0332) & (0.0391) \\
 Constant & 3.752*** & 5.937*** \\
 & (0.0106) & (0.0173) \\
 \hline
 Submarket FE & Yes & Yes \\
 Year-Month FE & Yes & Yes \\
 \hline
 Observations & 25,833 & 25,833 \\
\hline \hline
\end{tabular}
\begin{tablenotes}
\footnotesize
\centerline{Note: Cluster-robust standard errors are reported at the submarket level. *** $p$ $<$ 0.01, ** $p$ $<$ 0.05, * $p$ $<$ 0.1.}
\end{tablenotes}
\end{table}

\newpage
\section{Robustness Check - Alternative Aggregation Level (Week)}\label{appendix:Alternative Aggregation Level}

Leveraging the two-year dataset, our primary analyses use variables aggregated at the monthly level to smooth trends, particularly for smaller submarkets. In this section, we aggregate all variables to the weekly level and rerun the estimations. The outcomes of this analysis are presented in Table \ref{tab: main results market-level (Week-level)}. The estimation results remain qualitatively consistent with the main findings, confirming the robustness and reliability of our conclusions across different levels of data granularity. However, it is important to note that at the weekly level, there will be a substantial number of zero observations, which makes it less appropriate to directly interpret the coefficients as proportional changes.

\begin{table}[H]
\centering
\caption{Effects of Generative AI on Labor Market Outcomes (Week-level)}
\label{tab: main results market-level (Week-level)}
\newcolumntype{L}[1]{>{\raggedright\arraybackslash}p{#1}}
\newcolumntype{C}[1]{>{\centering\arraybackslash}p{#1}}
\newcolumntype{R}[1]{>{\raggedleft\arraybackslash}p{#1}}
\renewcommand\arraystretch{0.55}
\begin{tabular}{L{2.7cm}C{2.5cm}C{2.5cm}C{2.5cm}C{2.5cm}}
\hline 
 \hline
 & \multicolumn{4}{c}{$ln(y+1)$} \\
 \cline{2-5}
 & $Jobs$ & $Bids$ & $Transaction$ & $Avg\_Bids\_job$ \\
 \cline{2-5}
 & (1) & (2) & (3) & (4) \\
 \hline
  $Treat * Post$ & -0.181*** & -0.127*** & -0.294*** & 0.106*** \\
 & (0.0225) & (0.0388) & (0.0520) & (0.0281) \\
 Constant & 1.730*** & 3.706*** & 3.177*** & 1.895*** \\
 & (0.0073) & (0.0126) & (0.0169) & (0.0091) \\
 \hline
 Submarket FE & Yes & Yes & Yes & Yes  \\
 Year-Month FE & Yes & Yes & Yes & Yes \\
 \hline
 Observations & 112,744 & 112,744 & 112,744 & 112,744 \\
 $R^2$ & 0.799 & 0.725 & 0.485 & 0.374 \\
\hline \hline
\end{tabular}
\begin{tablenotes}
\footnotesize
\centerline{Note: Cluster-robust standard errors are reported at the submarket level. *** $p$ $<$ 0.01, ** $p$ $<$ 0.05, * $p$ $<$ 0.1.}
\end{tablenotes}
\end{table}

\newpage
\section{Robustness Check - Alternative  Fixed Effect}\label{appendix:Alternative Fixed Effect}

To more effectively account for seasonal effects, we also adopt an alternative specification that includes separate fixed effects for year ($Y_t$) and month ($M_t$), allowing us to control for both annual variations and recurring monthly patterns. This approach enhances interpretability by decomposing temporal variation into distinct components. The definitions and constructions of the remaining variables remain consistent with those outlined in Section \ref{sec: econometric}. The model specification is presented as follows:
\begin{align}\label{eq:main1}
    &Outcome_{it} = \beta_{0} + \beta_{1} * Treat_{i} * Post_t + \beta_{2} * Post_t + u_i + Y_t + M_t + \epsilon_{it}
\end{align}

Using this model specification, we re-estimate the market-level outcomes, with the results reported in Table \ref{tab: main results market-level (different FE)}. The findings suggest that the alternative fixed effects structure yields results that are broadly consistent with those of the main specification. 
\begin{table}[H]
\centering
\caption{Effects of Generative AI on Demand, Supply, and Matching Outcomes (Alternative Fixed Effects)}
\label{tab: main results market-level (different FE)}
\newcolumntype{L}[1]{>{\raggedright\arraybackslash}p{#1}}
\newcolumntype{C}[1]{>{\centering\arraybackslash}p{#1}}
\newcolumntype{R}[1]{>{\raggedleft\arraybackslash}p{#1}}
\renewcommand\arraystretch{0.55}
\begin{tabular}{L{2.7cm}C{2.5cm}C{2.5cm}C{2.5cm}C{2.5cm}}
\hline 
 \hline
 & \multicolumn{4}{c}{$ln(y+1)$} \\
 \cline{2-5}
 & $Jobs$ & $Bids$ & $Transaction$ & $Avg\_Bids\_job$ \\
 \cline{2-5}
 & (1) & (2) & (3) & (4) \\
 \hline
 $Treat * Post$ & -0.220*** & -0.168*** & -0.314*** & 0.137*** \\
 & (0.0318) & (0.0476) & (0.0896) & (0.0356) \\
 $Post$ & 0.147*** & 0.211*** & -0.108 & 0.0343 \\
 & (0.0327) & (0.0557) & (0.121) & (0.0467) \\
 Constant & 2.964*** & 5.327*** & 5.665*** & 2.075*** \\
 & (0.0065) & (0.0129) & (0.0358) & (0.0120) \\
 \hline
 Submarket FE & Yes & Yes & Yes & Yes  \\
 Year FE & Yes & Yes & Yes & Yes \\
 Month FE & Yes & Yes & Yes & Yes \\
 \hline
 Observations & 25,833 & 25,833 & 25,833 & 25,833 \\
 $R^2$ & 0.890 & 0.840 & 0.610 & 0.538 \\
\hline \hline
\end{tabular}
\begin{tablenotes}
\footnotesize
\centerline{Note: Cluster-robust standard errors are reported at the freelancer level. *** $p$ $<$ 0.01, ** $p$ $<$ 0.05, * $p$ $<$ 0.1.}
\end{tablenotes}
\end{table}

\newpage
\section{Robustness Check - Alternative Outcomes}\label{appendix:Market-level Extended Analyses}

In Section \ref{sec:main_results}, we have analyzed the average post-ChatGPT changes in labor demand, supply, and matching outcomes for treated submarkets relative to control submarkets, with a primary focus on changes in the overall volume of demand, supply, and transactions. To further substantiate these findings, we introduce three additional count variables corresponding to the demand side, supply side, and matching outcomes. These variables serve to both complement and validate the aggregate analysis by capturing changes in the number of platform participants and successfully matched jobs. Similar to other market-level variables used in our analysis, these are constructed at the submarket-month level to ensure consistency in temporal and unit resolution across all outcome measures. Specifically, on the demand side, we introduce the variable $Clients$, denoting the number of unique clients who post jobs within the submarket in a given month. On the supply side, we adopt the variable $Freelancers$, representing the number of distinct freelancers who have placed at least one bid within a particular submarket during a month. To complement the analysis of market matching outcomes, we construct the variable $Matches$, epitomizing the number of successful matches made (i.e., successfully completed jobs) within a submarket in a given month.

Following the same procedure as in the main analysis presented in Section \ref{sec:main_results}, we estimate the results using Equation \ref{eq:main}, as specified in Section \ref{sec: econometric}. The findings further confirm the observed contractions in demand, supply, and matching outcomes in the online labor market. Specifically, The results indicate a marginal decrease in the number of participating clients ($\beta_1 = -0.199$, $p < 0.01$), active freelancers ($\beta_1 = -0.141$, $p < 0.01$), and successful matches ($\beta_1 = -0.102$, $p < 0.01$) in the treatment submarkets relative to the control submarkets following the launch of ChatGPT. These additional analyses provide further support for the validity of our findings. 

Summary statistics for these additional variables are reported in Table \ref{tab: Variables Statistics other variables}. The corresponding estimation results, based on our empirical specification, are presented in Tables \ref{tab: main results market-level (Other Variables)} and \ref{tab: hte (Other Variables)}. These results provide additional support for the robustness of our findings. Furthermore, the parallel trends assumption is supported for these variables, as illustrated in Figure \ref{fig:pt}.

\newpage

\begin{table}[H]
\centering
\caption{Variable Summary Statistics }
\label{tab: Variables Statistics other variables}
\newcolumntype{L}[1]{>{\raggedright\arraybackslash}p{#1}}
\newcolumntype{C}[1]{>{\centering\arraybackslash}p{#1}}
\newcolumntype{R}[1]{>{\raggedleft\arraybackslash}p{#1}}
\renewcommand\arraystretch{0.55}
\begin{tabular}{L{3.5cm}C{1.8cm}C{1.8cm}C{1.8cm}C{1.8cm}C{1.8cm}}
\hline 
\hline
 Variable & Observation & Mean & Std & Min & Max \\
\hline
 $ln(Clients+1)$ & 25,833 & 2.849 & 1.211 & 0 & 6.898 \\
 $ln(Freelancers+1)$ & 25,833 & 5.0859 & 1.681 & 0 & 10.348 \\
 $ln(Matches+1)$ & 25,833 & 1.287 & 1.0445 & 0 & 5.278 \\  
\hline \hline
\end{tabular}
\end{table}

\begin{table}[H]
\centering
\caption{Effects of Generative AI on Other Market Outcomes}
\label{tab: main results market-level (Other Variables)}
\newcolumntype{L}[1]{>{\raggedright\arraybackslash}p{#1}}
\newcolumntype{C}[1]{>{\centering\arraybackslash}p{#1}}
\newcolumntype{R}[1]{>{\raggedleft\arraybackslash}p{#1}}
\renewcommand\arraystretch{0.55}
\begin{tabular}{L{2.7cm}C{2.5cm}C{2.5cm}C{2.5cm}}
\hline 
 \hline
 & $Clients$ & $Freelancers$ & $Matches$  \\
 \cline{2-4}
 & (1) & (2) & (3) \\
 \hline
 $Treat * Post$ & -0.199*** & -0.141*** & -0.102*** \\
 & (0.0292) & (0.0456) & (0.0252) \\
 Constant & 2.912*** & 5.131*** & 1.319*** \\
 & (0.0093) & (0.0146) & (0.0081) \\
 \hline
 Submarket FE & Yes & Yes & Yes \\
 Year-Month FE & Yes & Yes & Yes \\
 \hline
 Observations & 25,833 & 25,833 & 25,833 \\
 $R^2$ & 0.899 & 0.820 & 0.786 \\
\hline \hline
\end{tabular}
\begin{tablenotes}
\footnotesize
\centerline{Robust standard errors in parentheses *** $p$ $<$ 0.01, ** $p$ $<$ 0.05, * $p$ $<$ 0.1}
\end{tablenotes}
\end{table}

\begin{table}[H]
\centering
\caption{Effects of Generative AI on Other Market Outcomes}
\label{tab: hte (Other Variables)}
\newcolumntype{L}[1]{>{\raggedright\arraybackslash}p{#1}}
\newcolumntype{C}[1]{>{\centering\arraybackslash}p{#1}}
\newcolumntype{R}[1]{>{\raggedleft\arraybackslash}p{#1}}
\renewcommand\arraystretch{0.55}
\begin{tabular}{L{4.9cm}C{2.5cm}C{2.5cm}C{2.5cm}}
\hline 
 \hline
 & $Clients$ & $Freelancers$ & $Matches$  \\
\cline{2-4}
 & (1) & (2) & (3)  \\
 \hline
 $Treat * Post$ & -0.207*** & -0.194*** & -0.101*** \\
 & (0.0317) & (0.0495) & (0.0278) \\
 $Treat * Post * Programming$ & 0.0183 & 0.117*** & -0.0007 \\
 & (0.0210) & (0.0325) & (0.0216) \\
 Constant & 2.912*** & 5.131*** & 1.319*** \\
 & (0.0093) & (0.0146) & (0.0081) \\
 \hline
 Submarket FE & Yes & Yes & Yes \\
 Year-Month FE & Yes & Yes & Yes \\
 \hline
 Observations & 25,833 & 25,833 & 25,833 \\
 $R^2$ & 0.899 & 0.820 & 0.786 \\
\hline \hline
\end{tabular}
\begin{tablenotes}
\footnotesize
\centerline{Robust standard errors in parentheses *** $p$ $<$ 0.01, ** $p$ $<$ 0.05, * $p$ $<$ 0.1}
\end{tablenotes}
\end{table}

\newpage

\begin{figure}[H]
    \centering
    \subfloat[Clients]{%
      \includegraphics[width=0.55\textwidth]{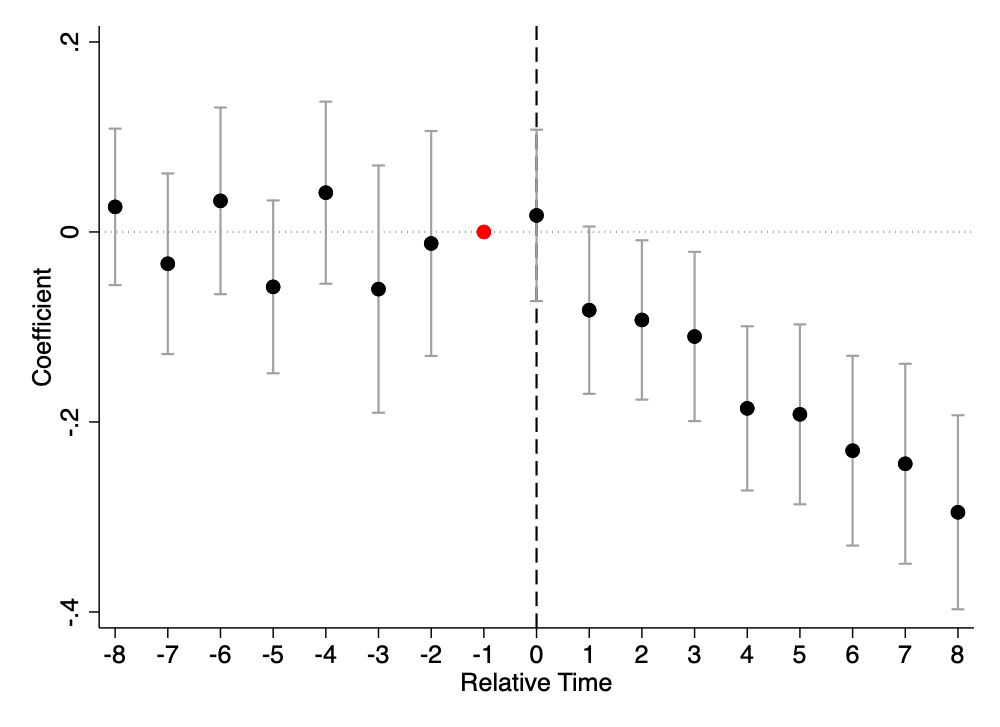}%
      \label{subfig:clients}%
    }\\[1em] 
    \subfloat[Freelancers]{%
      \includegraphics[width=0.55\textwidth]{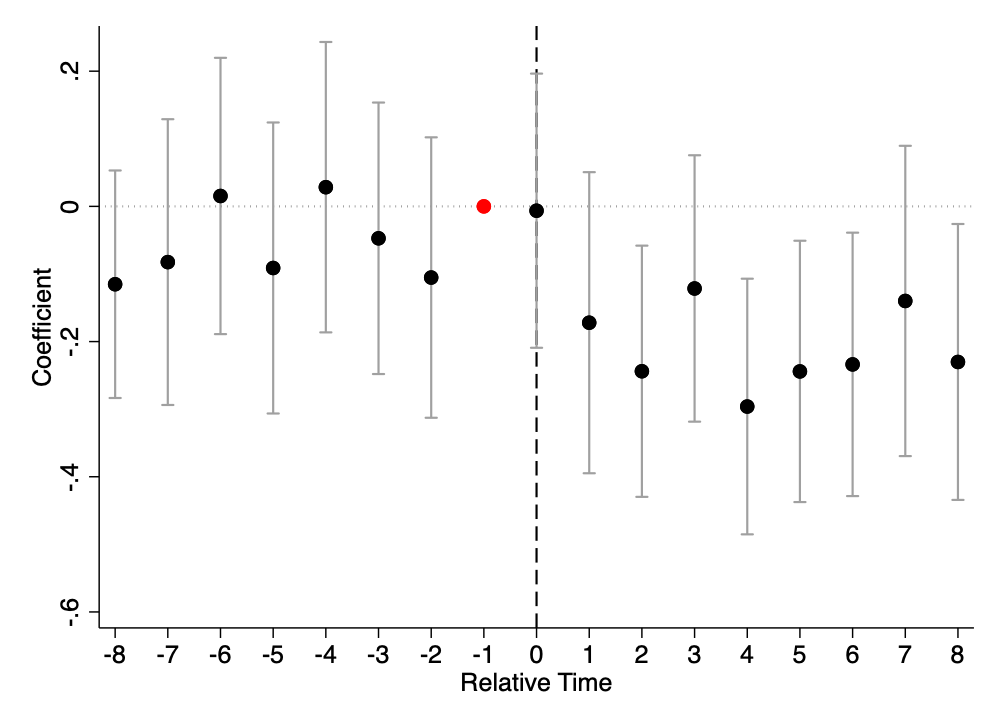}%
      \label{subfig:freelancers}%
    }\\[1em]
    \subfloat[Matches]{%
      \includegraphics[width=0.55\textwidth]{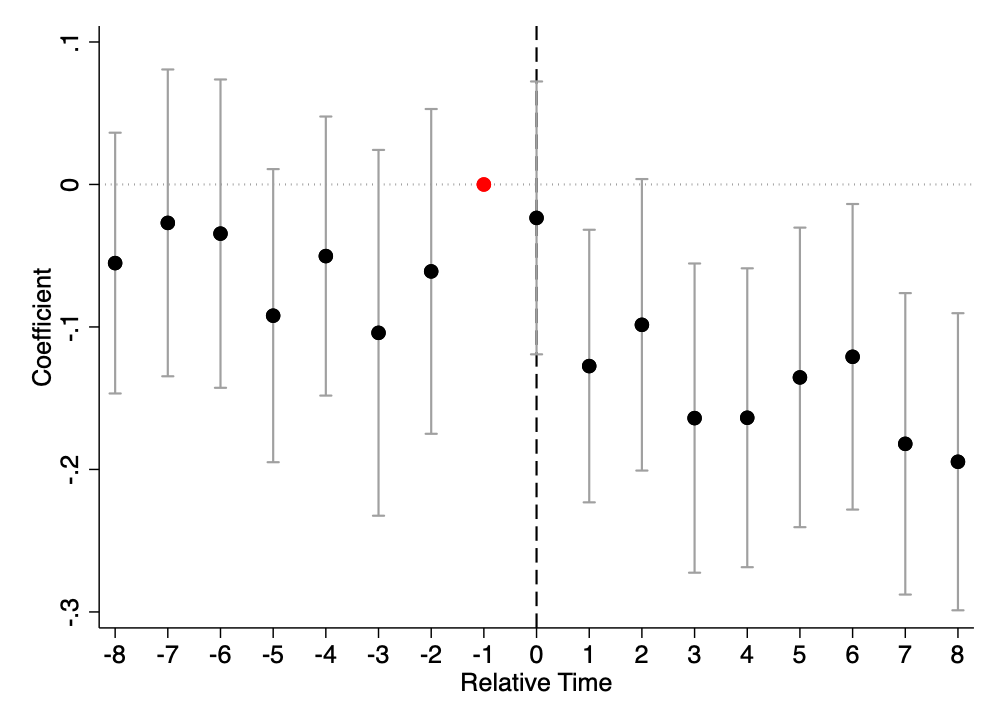}%
      \label{subfig:matches}%
    }

    \caption{\centering Pretrends and Temporal Dynamics}
    \label{fig:pt} 
\end{figure}

\newpage

\section{Robustness Check - Removing Outliers}\label{appendix:Alternative Submarkets Quantity}

Given that submarkets can vary significantly in size, one might question whether the estimates are disproportionately affected by the largest and the smallest submarkets. To address this concern, we first exclude the top 1\% of submarkets, followed by the removal of submarkets with an average job post count of less than one. The re-estimated model results, as presented in Table \ref{tab: Alternative Submarkets Quantity large} and \ref{tab: Alternative Submarkets Quantity small}, remain consistent with our main findings. This indicates that our conclusions are not driven by a few large submarkets.

\begin{table}[H]
\centering
\caption{The Impact of Generative AI after Excluding Outliers (Drop Top Submarkets)}
\label{tab: Alternative Submarkets Quantity large}
\newcolumntype{L}[1]{>{\raggedright\arraybackslash}p{#1}}
\newcolumntype{C}[1]{>{\centering\arraybackslash}p{#1}}
\newcolumntype{R}[1]{>{\raggedleft\arraybackslash}p{#1}}
\renewcommand\arraystretch{0.55}
\begin{tabular}{L{2.7cm}C{2.5cm}C{2.5cm}C{2.5cm}C{2.5cm}}
\hline 
 \hline
 & \multicolumn{4}{c}{$ln(y+1)$} \\
 \cline{2-5}
 & $Jobs$ & $Bids$ & $Transaction$ & $Avg\_Bids\_job$ \\
 \cline{2-5}
 & (1) & (2) & (3) & (4) \\
 \hline
  $Treat * Post$ & -0.222*** & -0.173*** & -0.318*** & 0.135*** \\
 & (0.0319) & (0.0479) & (0.0902) & (0.0358) \\
 Constant & 2.989*** & 5.371*** & 5.580*** & 2.0860*** \\
 & (0.0102) & (0.0153) & (0.0288) & (0.0114) \\
 \hline
 Submarket FE & Yes & Yes & Yes & Yes  \\
 Year-Month FE & Yes & Yes & Yes & Yes \\
 \hline
 Observations & 25,569 & 25,569 & 25,569 & 25,569 \\
 $R^2$ & 0.884 & 0.835 & 0.603 & 0.540 \\
\hline \hline
\end{tabular}
\begin{tablenotes}
\footnotesize
\centerline{Note: Cluster-robust standard errors are reported at the submarket level. *** $p$ $<$ 0.01, ** $p$ $<$ 0.05, * $p$ $<$ 0.1.}
\end{tablenotes}
\end{table}

\begin{table}[H]
\centering
\caption{The Impact of Generative AI after Excluding Outliers (Drop Bottom Submarkets)}
\label{tab: Alternative Submarkets Quantity small}
\newcolumntype{L}[1]{>{\raggedright\arraybackslash}p{#1}}
\newcolumntype{C}[1]{>{\centering\arraybackslash}p{#1}}
\newcolumntype{R}[1]{>{\raggedleft\arraybackslash}p{#1}}
\renewcommand\arraystretch{0.55}
\begin{tabular}{L{2.7cm}C{2.5cm}C{2.5cm}C{2.5cm}C{2.5cm}}
\hline 
 \hline
 & \multicolumn{4}{c}{$ln(y+1)$} \\
 \cline{2-5}
 & $Jobs$ & $Bids$ & $Transaction$ & $Avg\_Bids\_job$ \\
 \cline{2-5}
 & (1) & (2) & (3) & (4) \\
 \hline
  $Treat * Post$ & -0.219*** & -0.160*** & -0.299*** & 0.137*** \\
 & (0.0339) & (0.0480) & (0.0960) & (0.0348) \\
 Constant & 3.0740*** & 5.495*** & 5.727*** & 2.115*** \\
 & (0.0110) & (0.0155) & (0.0310) & (0.0112) \\
 \hline
 Submarket FE & Yes & Yes & Yes & Yes  \\
 Year-Month FE & Yes & Yes & Yes & Yes \\
 \hline
 Observations & 25,306 & 25,306 & 25,306 & 25,306 \\
 $R^2$ & 0.882 & 0.829 & 0.589 & 0.540 \\
\hline \hline
\end{tabular}
\begin{tablenotes}
\footnotesize
\centerline{Note: Cluster-robust standard errors are reported at the submarket level. *** $p$ $<$ 0.01, ** $p$ $<$ 0.05, * $p$ $<$ 0.1.}
\end{tablenotes}
\end{table}

\newpage
\section{Robustness Checks - Continues Treatment}\label{appendix:continues treatment}

To further verify the robustness of our main results, we re-estimate the main models using each sub-market's LM-AIOE score as a continuous treatment measure, replacing the median-split binary classification. Specifically, we estimate the following specification:
\begin{align}\label{eq:main-Continues}
    &Outcome_{it} = \beta_0 + \beta_{1} * LM\_AIOE_{i} * Post_t + u_i + T_t + \epsilon_{it} 
\end{align}
where $LM\_AIOE_{i}$ denotes the continuous LM-AIOE score of submarket $i$, capturing its degree of exposure to LLM-based technologies, and all other variables are defined the same as before. The coefficient $\beta_1$ captures the differential effect of ChatGPT across submarkets with varying levels of LM-AIOE exposure. The estimation results are reported in Tables \ref{tab:continues_treat} and \ref{tab: continues treatment (HTE)}. Across both outcomes, the estimates remain consistent with our baseline findings in terms of direction and statistical significance, further supporting the robustness of our main conclusions.

\begin{table}[H]
\centering
\caption{The Impact of Generative AI at the Market Level (Continuous Treatment)}
\label{tab:continues_treat}
\newcolumntype{L}[1]{>{\raggedright\arraybackslash}p{#1}}
\newcolumntype{C}[1]{>{\centering\arraybackslash}p{#1}}
\newcolumntype{R}[1]{>{\raggedleft\arraybackslash}p{#1}}
\renewcommand\arraystretch{0.55}
\begin{tabular}{L{3cm}C{2.5cm}C{2.5cm}C{2.5cm}C{2.5cm}}
\hline 
 \hline
 & \multicolumn{4}{c}{\(\ln(y+1)\)} \\
\cline{2-5}
 & $Jobs$ & $Bids$ & $Transaction$ & $Avg\_Bids\_job$ \\
 \cline{2-5}
 & (1) & (2) & (3) & (4) \\
 \hline
  $LM\_AIOE * Post$ & -0.292*** & -0.221*** & -0.525*** & 0.187*** \\
 & (0.0357) & (0.0535) & (0.114) & (0.0410) \\
 Constant & 3.0300*** & 5.414*** & 5.671*** & 2.0800*** \\
 & (0.0100) & (0.0150) & (0.0318) & (0.0114) \\
 \hline
 Submarket FE & Yes & Yes & Yes & Yes  \\
 Year-Month FE & Yes & Yes & Yes & Yes \\
 \hline
 Observations & 25,833 & 25,833 & 25,833 & 25,833 \\
 $R^2$ & 0.891 & 0.841 & 0.612 & 0.541 \\
\hline \hline
\end{tabular}
\begin{tablenotes}
\footnotesize
\centerline{Note: Cluster-robust standard errors are reported at the submarket level. *** $p$ $<$ 0.01, ** $p$ $<$ 0.05, * $p$ $<$ 0.1.}
\end{tablenotes}
\end{table}

\newpage

\begin{table}[H]
\centering
\caption{The Heterogeneous Impact of Generative AI at the Market Level (Continuous Treatment)}
\label{tab: continues treatment (HTE)}
\newcolumntype{L}[1]{>{\raggedright\arraybackslash}p{#1}}
\newcolumntype{C}[1]{>{\centering\arraybackslash}p{#1}}
\newcolumntype{R}[1]{>{\raggedleft\arraybackslash}p{#1}}
\renewcommand\arraystretch{0.55}
\resizebox{\textwidth}{!}{
\begin{tabular}{L{5.9cm}C{2.3cm}C{2.3cm}C{2.5cm}C{2.5cm}}
\hline 
 \hline
 & \multicolumn{4}{c}{\(\ln(y+1)\)} \\
 \cline{2-5}
 & $Jobs$ & $Bids$ & $Transaction$ & $Avg\_Bids\_job$ \\
 \cline{2-5}
 & (1) & (2) & (3) & (4) \\
 \hline
 $LM\_AIOE * Post$ & -0.296*** & -0.264*** & -0.536*** & 0.146*** \\
 & (0.0363) & (0.0540) & (0.116) & (0.0419) \\
 $LM\_AIOE * Post * Programming$ & 0.0144 & 0.159*** & 0.0413 & 0.153*** \\
 & (0.0266) & (0.0413) & (0.0798) & (0.0316) \\
 Constant & 3.0300*** & 5.407*** & 5.669*** & 2.0730*** \\
 & (0.0100) & (0.0151) & (0.0320) & (0.0114) \\
 \hline
 Submarket FE & Yes & Yes & Yes & Yes  \\
 Year-Month FE & Yes & Yes & Yes & Yes \\
 \hline
 Observations & 25,833 & 25,833 & 25,833 & 25,833 \\
 $R^2$ & 0.891 & 0.841 & 0.612 & 0.542 \\
\hline \hline
\end{tabular}
}
\begin{tablenotes}
\footnotesize
\centerline{Note: Cluster-robust standard errors are reported at the submarket level. *** $p$ $<$ 0.01, ** $p$ $<$ 0.05, * $p$ $<$ 0.1.}
\end{tablenotes}
\end{table}

\newpage
\section{Robustness Checks - Multi-group Analysis}\label{appendix:multi_group}

To further assess the robustness of our findings to the choice of treatment categorization, we replace the median-split binary classification with a tercile-based approach, dividing submarkets into three groups (high, medium, and low exposure) based on their LM-AIOE values. We use the low-exposure group as the baseline and estimate separate treatment effects for the high- and medium-exposure groups. Tables \ref{tab:multi-group} and \ref{tab:multi-group hte} report the results for the main analysis and heterogeneity analyses, respectively.

As shown in Table \ref{tab:multi-group}, the results remain consistent with our baseline findings: relative to the low-exposure group, both the high- and medium-exposure groups experience significant declines in jobs, bids, and transactions following ChatGPT's introduction, with stronger effects for the high-exposure group. Results in Table \ref{tab:multi-group hte} also show a broadly similar heterogeneity pattern with main analysis. Together, these results further support the robustness of our main findings.

\newpage
\begin{table}[H]
\centering
\caption{The Impact of Generative AI at the Market Level (Three Groups)}
\label{tab:multi-group}
\newcolumntype{L}[1]{>{\raggedright\arraybackslash}p{#1}}
\newcolumntype{C}[1]{>{\centering\arraybackslash}p{#1}}
\renewcommand\arraystretch{0.55}
\begin{tabular}{L{2.7cm}C{2.5cm}C{2.5cm}C{2.5cm}C{2.5cm}}
\hline\hline
\multicolumn{5}{c}{\textbf{Panel A}}\\
\hline
&\multicolumn{4}{c}{High vs Low}\\
\cline{2-5}
&\multicolumn{4}{c}{$ln(y+1)$}\\
\cline{2-5}
& $Jobs$ & $Bids$ & $Transaction$ & $Avg\_Bids\_job$\\
\cline{2-5}
& (1) & (2) & (3) & (4)\\
\hline
$Treat * Post$  & -0.258*** & -0.201*** & -0.421*** & 0.167*** \\
& (0.0345) & (0.0518) & (0.0982) & (0.0387)\\
Constant & 3.0070*** & 5.382*** & 5.534*** & 2.0460*** \\
& (0.0096) & (0.0144) & (0.0272) & (0.0107) \\
\hline
Submarket FE & Yes & Yes & Yes & Yes\\
Year-Month FE & Yes & Yes & Yes & Yes\\
\hline
Observations  & 14,855 & 14,855 & 14,855 & 14,855 \\
R$^{2}$ & 0.886 & 0.837 & 0.614 & 0.526 \\
\hline
\hline
\multicolumn{5}{c}{\textbf{Panel B}}\\
\hline
&\multicolumn{4}{c}{Medium vs Low}\\
\cline{2-5}
&\multicolumn{4}{c}{$ln(y+1)$}\\
\cline{2-5}
& $Jobs$ & $Bids$ & $Transaction$ & $Avg\_Bids\_job$\\
\cline{2-5}
& (1) & (2) & (3) & (4)\\
\hline
$Treat * Post$  & -0.183*** & -0.135*** & -0.209** & 0.109*** \\
& (0.0331) & (0.0499) & (0.0934) & (0.0378) \\
Constant & 2.825*** & 5.0880*** & 5.299*** & 2.0550*** \\
& (0.0092) & (0.0139) & (0.0260) & (0.0105) \\
\hline
Submarket FE & Yes & Yes & Yes & Yes\\
Year-Month FE & Yes & Yes & Yes & Yes\\
\hline
Observations  & 14,850 & 14,850 & 14,850 & 14,850 \\
R$^{2}$   & 0.894 & 0.845 & 0.631 & 0.543 \\
\hline\hline
\end{tabular}
\begin{tablenotes}
\footnotesize
\centerline{Note: Cluster-robust standard errors are reported at the submarket level. *** $p$ $<$ 0.01, ** $p$ $<$ 0.05, * $p$ $<$ 0.1.}
\end{tablenotes}
\end{table}

\begin{table}[H]
\centering
\caption{The Heterogeneous Impact of Generative AI at the Market Level (Three Groups)}
\label{tab:multi-group hte}
\newcolumntype{L}[1]{>{\raggedright\arraybackslash}p{#1}}
\newcolumntype{C}[1]{>{\centering\arraybackslash}p{#1}}
\renewcommand\arraystretch{0.55}
\begin{tabular}{L{4.9cm}C{2.5cm}C{2.5cm}C{2.5cm}C{2.5cm}}
\hline\hline
\multicolumn{5}{c}{\textbf{Panel A}}\\
\hline
&\multicolumn{4}{c}{High vs Low}\\
\cline{2-5}
&\multicolumn{4}{c}{$ln(y+1)$}\\
\cline{2-5}
& $Jobs$ & $Bids$ & $Transaction$ & $Avg\_Bids\_job$\\
\cline{2-5}
& (1) & (2) & (3) & (4)\\
\hline
$Treat * Post$  & -0.282*** & -0.289*** & -0.461*** & 0.0982** \\
& (0.0378) & (0.0567) & (0.107) & (0.0422) \\
$Treat * Post * Programming$ & 0.0649* & 0.234*** & 0.107 & 0.185***\\
& (0.0348) & (0.0530) & (0.107) & (0.0395) \\
Constant & 3.0070*** & 5.382*** & 5.534*** & 2.0460*** \\
& (0.0096) & (0.0143) & (0.0272) & (0.0107) \\
\hline
Submarket FE & Yes & Yes & Yes & Yes\\
Year-Month FE & Yes & Yes & Yes & Yes\\
\hline
Observations  & 14,855 & 14,855 & 14,855 & 14,855 \\
R$^{2}$       & 0.886 & 0.837 & 0.614 & 0.527 \\
\hline
\multicolumn{5}{c}{\textbf{Panel B}}\\
\hline
&\multicolumn{4}{c}{Medium vs Low}\\
\cline{2-5}
&\multicolumn{4}{c}{$ln(y+1)$}\\
\cline{2-5}
& $Jobs$ & $Bids$ & $Transaction$ & $Avg\_Bids\_job$ \\
\cline{2-5}
& (1) & (2) & (3) & (4)\\
\hline
$Treat * Post$  & -0.169*** & -0.169*** & -0.186* & 0.0667 \\
& (0.0393) & (0.0604) & (0.108) & (0.0450) \\
$Treat * Post * Programming$ & -0.0251 & 0.0624 & -0.0426 & 0.0776** \\
 & (0.0305) & (0.0483) & (0.0871) & (0.0379) \\
Constant & 2.825*** & 5.0880*** & 5.299*** & 2.0550*** \\
& (0.0092) & (0.0139) & (0.0260) & (0.0105) \\
\hline
Submarket FE & Yes & Yes & Yes & Yes\\
Year-Month FE & Yes & Yes & Yes & Yes\\
\hline
Observations  & 14,850 & 14,850 & 14,850 & 14,850 \\
R$^{2}$       & 0.894 & 0.845 & 0.631 & 0.543 \\
\hline\hline
\end{tabular}
\begin{tablenotes}
\footnotesize
\centerline{Note: Cluster-robust standard errors are reported at the submarket level. *** $p$ $<$ 0.01, ** $p$ $<$ 0.05, * $p$ $<$ 0.1.}
\end{tablenotes}
\end{table}

\newpage
\section{Robustness Checks - Alternative Measures of Gen AI Exposure}\label{appendix:new_treat}

To further examine the robustness of our results, we adopt an alternative peer-reviewed measure of occupational AI exposure from \cite{eloundou2024gpts}, who assess the labor market impact potential of large language models across occupations. We reconstruct the treatment classification and re-estimate the models. Tables \ref{tab:Alternative Measures of Generative AI Exposure} and \ref{tab: Alternative Measures of Generative AI Exposure (HTE)} present the corresponding estimation results. The findings remain largely consistent with the main analysis, further supporting the robustness of our conclusions.

\begin{table}[H]
\centering
\caption{The Impact of Generative AI at the Market Level (Alternative Measures of Gen AI Exposure)}
\label{tab:Alternative Measures of Generative AI Exposure}
\newcolumntype{L}[1]{>{\raggedright\arraybackslash}p{#1}}
\newcolumntype{C}[1]{>{\centering\arraybackslash}p{#1}}
\newcolumntype{R}[1]{>{\raggedleft\arraybackslash}p{#1}}
\renewcommand\arraystretch{0.55}
\begin{tabular}{L{2.7cm}C{2.5cm}C{2.5cm}C{2.5cm}C{2.5cm}}
\hline 
 \hline
 & \multicolumn{4}{c}{\(ln(y+1)\)} \\
\cline{2-5}
 & $Jobs$ & $Bids$ & $Transaction$ & $Avg\_Bids\_job$ \\
 \cline{2-5}
 & (1) & (2) & (3) & (4) \\
 \hline
 $Treat * Post$ & -0.233*** & -0.117** & -0.339*** & 0.219*** \\
 & (0.0317) & (0.0508) & (0.0890) & (0.0354) \\
 Constant & 3.0200*** & 5.388*** & 5.629*** & 2.0650*** \\
 & (0.0098) & (0.0157) & (0.0274) & (0.0109) \\
 \hline
 Submarket FE & Yes & Yes & Yes & Yes  \\
 Year-Month FE & Yes & Yes & Yes & Yes \\
 \hline
 Observations & 25,833 & 25,833 & 25,833 & 25,833 \\
 $R^2$ & 0.891 & 0.841 & 0.612 & 0.542 \\
\hline \hline
\end{tabular}
\begin{tablenotes}
\footnotesize
\centerline{Note: Cluster-robust standard errors are reported at the submarket level. *** $p$ $<$ 0.01, ** $p$ $<$ 0.05, * $p$ $<$ 0.1.}
\end{tablenotes}
\end{table}

\begin{table}[H]
\centering
\caption{The Heterogeneous Impact of Generative AI at the Market Level (Alternative Measures of Gen AI Exposure)}
\label{tab: Alternative Measures of Generative AI Exposure (HTE)}
\newcolumntype{L}[1]{>{\raggedright\arraybackslash}p{#1}}
\newcolumntype{C}[1]{>{\centering\arraybackslash}p{#1}}
\newcolumntype{R}[1]{>{\raggedleft\arraybackslash}p{#1}}
\renewcommand\arraystretch{0.55}
\begin{tabular}{L{4.9cm}C{2.5cm}C{2.5cm}C{2.5cm}C{2.5cm}}
\hline 
 \hline
 & \multicolumn{4}{c}{\(ln(y+1)\)} \\
 \cline{2-5}
 & $Jobs$ & $Bids$ & $Transaction$ & $Avg\_Bids\_job$ \\
 \cline{2-5}
 & (1) & (2) & (3) & (4) \\
 \hline
 $Treat * Post$ & -0.246*** & -0.171*** & -0.367*** & 0.183*** \\
 & (0.0344) & (0.0548) & (0.0958) & (0.0382) \\
 $Treat * Post * Programming$ & 0.0297 & 0.118*** & 0.0607 & 0.0765*** \\
 & (0.0221) & (0.0338) & (0.0670) & (0.0266) \\
 Constant & 3.0200*** & 5.388*** & 5.629*** & 2.0650*** \\
 & (0.0098) & (0.0156) & (0.0274) & (0.0109) \\
 \hline
 Submarket FE & Yes & Yes & Yes & Yes  \\
 Year-Month FE & Yes & Yes & Yes & Yes \\
 \hline
 Observations & 25,833 & 25,833 & 25,833 & 25,833 \\
 $R^2$ & 0.891 & 0.841 & 0.612 & 0.542 \\
\hline \hline
\end{tabular}
\begin{tablenotes}
\footnotesize
\centerline{Note: Cluster-robust standard errors are reported at the submarket level. *** $p$ $<$ 0.01, ** $p$ $<$ 0.05, * $p$ $<$ 0.1.}
\end{tablenotes}
\end{table}

\newpage

\begin{landscape}
\section{Recent Literature on AI in Labor Markets}
\label{appendix:recent literature}
\scriptsize
\begin{longtable}{
  >{\raggedright\arraybackslash}p{4.5cm}
  >{\raggedright\arraybackslash}p{2.8cm}
  >{\raggedright\arraybackslash}p{2.2cm}
  >{\raggedright\arraybackslash}p{2.1cm}
  >{\raggedright\arraybackslash}p{2.2cm}
  >{\raggedright\arraybackslash}p{7.5cm}
}
\caption{Recent Literature on AI in Labor Markets} \label{tab:recent literature} \\
\toprule
\textbf{Title (Authors)} & \textbf{Research Context and Focus} & \textbf{Type of Technology} & \textbf{Level of Analysis} & \textbf{Demand-side or Supply-side} & \textbf{Relevant Takeaways} \\
\midrule
\endfirsthead
\multicolumn{6}{c}{Table \ref{tab:recent literature} continued} \\
\toprule
\textbf{Title (Authors)} & \textbf{Research Context and Focus} & \textbf{Type of Technology} & \textbf{Level of Analysis} & \textbf{Demand-side or Supply-side} & \textbf{Relevant Takeaways} \\
\midrule
\endhead
\midrule
\multicolumn{6}{r}{Continued on next page} \\
\endfoot
\bottomrule
\endlastfoot

Threatened by AI: Analyzing users' responses to the introduction of AI in a crowd-sourcing platform \citep{lysyakov2023threatened} 
& Impact of AI Logo System on a Crowd-sourcing Platform
& Specialized AI
& Designer-level 
& Both Sides 
& $\bullet$~Demand displacement effect: The AI launch reduces the number of lower-tier simple logo contests. \par
$\bullet$~Workers' heterogeneous responses: Low-ability designers exit the platform, while higher-ability designers migrate to higher-tier logo contests or non-logo contests. \par \\

\midrule
The short-term effects of generative artificial intelligence on employment: Evidence from an online labor market \citep{hui2024short} & Impact of Generative AI on Online Labor Markets & General-Purpose AI & Freelancer-level &
Supply Side & 
$\bullet$~Demand displacement effect: AI-exposed freelancers experienced fewer jobs and lower earnings after ChatGPT's release. \par
$\bullet$~Heterogeneous displacement effect: Top-rated AI-exposed freelancers are hurt more. \par
\\

\midrule
Skill-biased technical change, again? Online gig platforms and local employment \citep{guo2025skill} & Impact of Gig Platform Entry (TaskRabbit) on Local Worker Participation & Digital Platform & Market-level & Supply Side &
$\bullet$~(Heterogeneous) demand displacement effect: TaskRabbit entry reduces the employment of middle-skilled workers whose tasks overlap with platform matching algorithms. \par
$\bullet$~Redistribution effect: Displaced middle-skilled workers shift into self-employment as small business owners.
\\

\midrule
AI, Skill, and Productivity: The Case of Taxi Drivers \citep{kanazawa2026ai} & Impact of Routing AI on Taxi Drivers' Productivity & Specialized AI & Employee-level & Supply Side &
$\bullet$~Productivity effect: Routing AI reduces cruising search time among taxi drivers. \par
$\bullet$~Deskilling effect (heterogeneity): The productivity improvement mainly happens among low-skilled drivers, narrowing the productivity gap between high- and low-skilled workers and suggesting that AI substitutes for worker skills.
\\

\midrule
Who Is AI Replacing? The Impact of Generative AI on Online Freelancing Platforms  \citep{DemirciOzge2025WIAR} & Impact of Generative AI on Online Labor Markets & General-Purpose AI & Market-level & Demand Side &
$\bullet$~Demand displacement effect: ChatGPT's release reduces the number of automation-prone job postings relative to manual-intensive jobs. \par
\\

\midrule
Reskilling the Workforce for AI: Domain Expertise and Algorithmic Literacy \citep{tambe2026reskilling} & Firms' Hiring for Algorithmic Skill & AI Technology at Large & Firm-level & Demand Side &
$\bullet$~Firms increasingly hire algorithmic skills for non-technical occupations, particularly in roles with high decision-making responsibilities. \par
$\bullet$~Skill-transition effect of tools: The diffusion of algorithmic skills is accelerated by reductions in learning costs (e.g., no-code tools). \par \\

\midrule
Generative AI at Work \citep{brynjolfsson2025generative} 
& Impact of Generative AI Assistant on Customer Service Productivity
& Specialized AI (Fine-Tuned for Customer Service)
& Employee-level
& Supply Side
& $\bullet$~Productivity effect: AI assistance increases worker productivity. \par
  $\bullet$~Deskilling effect (heterogeneity): Low-skill and less experienced workers have larger improvements, compressing the productivity distribution. \par \\

 \midrule
\textbf{This Paper} 
& \textbf{Impact of Generative AI on Online Labor Markets} 
& \textbf{General-Purpose AI}
& \textbf{Market-level \& Freelancer-level} 
& \textbf{Both Sides} 
& \textbf{$\bullet$~Demand displacement effect: Generative AI reduces job postings in exposed submarkets.} \par
  \textbf{$\bullet$~Competition intensification effect: Labor supply declines less than demand, raising the average number of bids per job.} \par
  \textbf{$\bullet$~Skill-transition effect: Freelancers transition toward programming-intensive submarkets, even though those submarkets experience a similar demand displacement.} \par
  \textbf{$\bullet$~Heterogeneous skill-transition effect: High-skilled freelancers are more likely to transition, consistent with differences in absorptive capacity.}\\

\end{longtable}
\end{landscape}

\newpage

\end{APPENDICES}

\end{document}